\documentclass{article}

     \PassOptionsToPackage{numbers}{natbib}

\usepackage[preprint]{neurips_2021}

\usepackage[utf8]{inputenc} 
\usepackage[T1]{fontenc}    
\usepackage{hyperref}       
\usepackage{url}            
\usepackage{booktabs}       
\usepackage{amsfonts}       
\usepackage{nicefrac}       
\usepackage{microtype}      
\usepackage{xcolor}         
\usepackage{graphicx}
\usepackage{microtype}
\usepackage{graphicx}
\usepackage{subfigure}
\usepackage{booktabs} 
\usepackage{makecell}

\usepackage{amsmath}
\usepackage{amsfonts} 
\usepackage{amsthm}
\usepackage{algorithm}
\usepackage{algorithmic}
\usepackage{url} 
\usepackage{amsfonts}       
\usepackage{nicefrac}       
\usepackage{microtype}      
\usepackage{xspace}
\usepackage{amsmath}
\usepackage{interval}
\usepackage{enumitem}
\usepackage{tabularx}
\usepackage{graphicx}
\usepackage{multirow}
\usepackage{color}
\usepackage{longtable}
\usepackage{wrapfig}
\usepackage{hyperref}

\usepackage{geometry}
\usepackage{color,float,amssymb,amsmath}
\usepackage{pdfpages}
\usepackage{wrapfig}
\usepackage{caption}
\usepackage{amsmath}
\usepackage{amsfonts}

\newcommand{\x}{{\mathbf x}}

\newcommand{\G}{{\mathbf G}}
\newcommand{\Y}{{\mathbf Y}}
\newcommand{\R}{\mathbb{R}}

\newcommand{\s}{{\mathbf s}}

\newcommand{\h}{{\mathbf h}}
\newcommand{\y}{{\mathbf y}}
\newcommand{\floor}[1]{\lfloor #1 \rfloor}
\newcommand{\E}{\mathbb{E}}

\title{NRTSI: Non-Recurrent Time Series Imputation}

\author{%
  
  Siyuan Shan \\
  Department of Computer Science\\
  University of North Carolina at Chapel Hill\\
  \texttt{siyuanshan@cs.unc.edu} 
  
  \And
  Yang Li \\
  Department of Computer Science\\
  University of North Carolina at Chapel Hill\\
  \texttt{yangli95@cs.unc.edu} 
  
  \And
  Junier B. Oliva \\
  Department of Computer Science\\
  University of North Carolina at Chapel Hill\\
  \texttt{joliva@cs.unc.edu} 
}

\begin{document}

\maketitle

\begin{abstract}
Time series imputation is a fundamental task for understanding time series with missing data. Existing methods either do not directly handle irregularly-sampled data or degrade severely with sparsely observed data. 
In this work, we reformulate time series as permutation-equivariant sets and propose a novel imputation model NRTSI that does not impose any recurrent structures.
Taking advantage of the permutation equivariant formulation, we design a principled and efficient hierarchical imputation procedure. In addition, NRTSI can directly handle irregularly-sampled time series, perform multiple-mode stochastic imputation, and handle data with partially observed dimensions. Empirically, we show that NRTSI achieves state-of-the-art performance across a wide range of time series imputation benchmarks. 
\end{abstract}

\section{Introduction}
Missing values are common in real-world time series, e.g. trajectories often contain missing data due to unreliable sensors or object occlusion. Recovering those missing values is useful for the downstream analysis of time series. 
Modern approaches impute missing data in a data-driven fashion. For example, recurrent neural networks (RNNs) are applied in \cite{NIPS2018_7432, che2018recurrent, cao2018brits}, methods that are built on Neural Ordinary Differential Equations (NODE) \cite{chen2018neural} are proposed in \cite{rubanova2019latent,de2019gru}, and a family of models called Neural Process \cite{garnelo2018neural, kim2018attentive} that learns a distribution over functions based on the observed data could also be leveraged. 

However, these existing works all have their own deficiencies. Models that are built on RNNs usually employ a sequential imputation order, meaning that the imputed data $x_t$ at timestep $t$ is predicted based on the already imputed data $x_{t-1}$ at the previous timestamp $t-1$. Since $x_{t-1}$ inevitably contains errors, $x_t$ is even more inaccurate and the errors will accumulate through time, resulting in poor long-horizon imputations for time series that are sparsely observed. This problem is known as \textit{error compounding} in the fields of time series analysis \cite{venkatraman2015improving} and reinforcement learning \cite{asadi2018lipschitz}. 
Previous work \cite{liu2019naomi} alleviates this problem with an RNN-based model, NAOMI, that imputes from coarse to fine-grained resolutions in a hierarchical imputation order. However, this approach is only directly applicable on regularly-sample time series since RNNs assume a constant time interval between observations. The family of NODE models \cite{rubanova2019latent, de2019gru, kidger2020neural} can directly process irregularly-sampled data, but they need to integrate complicated differential equations and also have a similar error accumulating problem due to their recurrent nature. Neural Process models \cite{garnelo2018conditional, kim2018attentive, garnelo2018neural} directly handle irregular data by taking timestamps as scalar inputs to the model. However, they impute all the missing data at once which we find deteriorate their performance.


In this work, we propose NRTSI, a \textbf{N}on-\textbf{R}ecurrent \textbf{T}ime \textbf{S}eries \textbf{I}mputation model.
One of our key insights is that when imputing missing values in time series, the valuable information from the observed data is \emph{what happened and when}. 
This information is most naturally represented as a set of (time, data) tuples. We propose to encode the set of tuples using a permutation equivariant model, which respects the unordered nature of the set. This is in stark contrast to previous works (e.g. NAOMI \cite{liu2019naomi}) where observed data are embedded recurrently so that the temporal information (when things happened) is unnecessarily coupled with the order of points being processed. Our natural set representation provides two immediate advantages to recurrent embeddings. Firstly, it can handle irregularly-sampled time series (i.e. non-grid timesteps) since it represents arbitrary times in the form of (time, data) tuples.
Secondly, it enables a more effective hierarchical imputation strategy without the sequential constraint to ameliorate the impact of error compounding. Despite its simplicity, we find NRTSI effectively alleviates the problems of the existing methods in a single framework while achieving state-of-the-art performance across multiple benchmarks. 

Our contributions are as follows: (1) We reinterpret time series as a set of (time, data) tuples and propose a time series imputation approach, NRTSI, using permutation equivariant models. 
(2) We propose an effective hierarchical imputation strategy that takes advantage of the non-recurrent nature of NRTSI and imputes data in a multiresolution fashion. (3) NRTSI can flexibly handle irregularly-sampled data, data with partially observed time dimensions, and perform stochastic imputations for non-deterministic time series. (4) Experiments on a wide range of datasets demonstrate state-of-the-art imputation performance of NRTSI compared to several strong baselines. Codes are available at \url{https://github.com/lupalab/NRTSI}.

\section{Related Work}
\textbf{Time Series Imputation}\ \ \  Existing time series imputation approaches roughly fall into two categories: statistical methods \cite{acuna2004treatment,ansley1984estimation,azur2011multiple,friedman2001elements} and deep generative models \cite{cao2018brits, li2020acflow, luo2018multivariate}. Statistical methods, such as mean-median averaging \cite{acuna2004treatment}, linear regression \cite{ansley1984estimation}, MICE \cite{azur2011multiple} and k-nearest neighbours \cite{friedman2001elements}, often impose strong assumptions on the missing patterns and design hand-crafted rules. Though effective on simple tasks, these methods fail on more complicated scenarios where the missing patterns vary from task to task. Deep generative models offer a more flexible framework for imputation. For example, \cite{che2018recurrent,cao2018brits,choi2020rdis} propose variants of RNNs to impute time series. Generative adversarial networks are leveraged in \cite{fedus2018maskgan, yoon2018gain, luo2018multivariate}. However, all of these works are recurrent. 

NAOMI \cite{liu2019naomi} performs time series imputation via a non-recurrent imputation procedure that imputes from coarse to fine-grained resolutions using a divide-and-conquer strategy. However, NAOMI relies on bidirectional RNNs to process observed time points, which limits its application for irregularly-sampled time data and loses the opportunity to efficiently impute multiple time points in parallel. Moreover, the hierarchical imputation procedure of NAOMI assumes that the multivariate data at a timestep is either completely observed or completely missing along all the dimensions. In contrast, the imputation procedure of NRTSI also works well when dimensions are partially observed. CDSA \cite{ma2019cdsa} performs time series imputation via self-attention without recurrent modules. However, CDSA is specifically designed for geo-tagged data, unable to tackle irregularly-sampled data, and does not exploit multiresolution information.

\textbf{Irregularly-sampled Time Series}\ \ \  RNN is the dominant model for high-dimensional, regularly-sampled time series. However, it is not suitable for irregularly-sampled time series. A commonly used trick is to divide the timeline into equally-spaced intervals and aggregate observations inside one interval by averaging \cite{lipton2016directly}. However, such preprocessing inevitably loses information, especially about the timing of measurements for accurate predictions.
Several recent works \cite{rajkomar2018scalable, che2018recurrent} propose variants of RNNs with continuous dynamics given by a simple exponential decay between observations. These models are further improved by NODE \cite{chen2018neural} based methods that learn the continuous dynamics rather than using hand-crafted exponential decay rules. For example, ODE-RNN \cite{rubanova2019latent, de2019gru} models the latent dynamics using ODE and updates the latent state at observations. This line of works is improved by NeuralCDE \cite{kidger2020neural} that enables the latent state to have a continuous dependency on the observed data via controlled differential equations \cite{lyons2007differential}. Despite the success of these continuous variants of RNNs for irregular time series analysis, these models inherit the recurrent nature of RNNs. Even though the exploding/vanishing gradient problem of ODE-RNN is mitigated by introducing forgetting gates to NODE \cite{lechner2020learning}, 
these models still suffer from error compounding problems.


Similar to NRTSI, SeFT \cite{horn2020set} and attentive neural process (ANP) \cite{kim2018attentive} view a temporal sequence as an unordered set, where each set element is a tuple that contains the observed data $\x_t \in \R^d$ at timestep $t\in\R$ and an encoding \cite{xu2019self} of the time $\phi(t)$.
However, only time series classification is considered in \cite{xu2019self,horn2020set}, while we target at time series imputation. 
Although ANP is applicable for the imputation task, 
the information of what timesteps to impute (target input) is not utilized when ANP uses self-attention to compute the representations of observed data (context input/output pairs).
Therefore, the representation might be suboptimal.
Moreover, ANP does not exploit the multiresolution information of sequences. These two shortcomings impact the performance of ANP as shown below in the experiments.

\section{Methods}
\label{sec:method}

\textbf{Motivation}\ \ \ 
Models such as RNNs that scan sequentially, at first glance, are natural fits for time series data due to their temporal nature. For imputation, however, these models typically suffer from a problem called \textit{error compounding}. To impute $n$ missing data sequentially, they impute the current data $x_t$ based on the previously imputed data $x_{t-1}$ at every step for $t=2,\ldots,n$. Since the first imputed data $x_1$ inevitably contains errors, the errors will accumulate for $n-1$ steps through time to $x_n$ and degrade the imputation performance significantly especially when $n$ is large. Moreover, some sequential models, such as RNNs, can only deal with regularly spaced time points.
To remedy these deficiencies, we reinterpret time series as a set of (time, data) tuples. The set formulation allows us to conveniently develop a hierarchical scheme that reduces the number of imputation steps required compared to the sequential scheme and thus effectively alleviates the error compounding problem. 
It also directly enables imputing irregularly sampled time points, since the set can contain tuples for arbitrary time points. \textit{Note that since the time information is provided in the (time, data) tuples, the sequential order of the time series is not lost, and we can easily transform the set back to a sequence.}

\textbf{Formulation}\ \ \  Throughout the paper, we denote a set as $\mathbf X=\{\x_i\}_{i=1}^N$ with the set element $\x_i \in \mathcal X$, where $\mathcal X$ represents the domain of each set element.
We denote a time series with $N$ observations as a set $\mathbf S=\{\s_i\}_{i=1}^N$, each observation $\s_i$ is a tuple $(t_i, \x_i)$, where $t_i \in \R^+$ denotes the observation time and $\x_i \in \R^d$ represents the observed data. 
Given an observed time series $\mathbf S$, we aim to impute the missing data based on $\mathbf S$. We also organize data to impute as a set $\hat{\mathbf S}=\{\hat{\s}_j\}_{j=1}^M$, where $M$ is the number of missing time points. 
Each set element $\hat{\s}_j$ is a tuple $(\hat{t}_j, \Delta \hat{t}_j)$, where $\hat{t}_j\in\R^+$ is a timestep to impute and $\Delta \hat{t}_j\in\R^+$ denotes the missing gap (i.e. the time interval length between $\hat{t}_j$ and its closest observation time in $\mathbf S$). Formally, $\Delta \hat{t}_j$ is defined as 
$\Delta \hat{t}_j = \min_{(t_i, \x_i)\in \mathbf S}  |t_i - \hat{t}_j|.$
Note that both $\hat{t}_j$ and $t_i$ are real-valued scalars as our model is designed to handle arbitrary irregularly-sampled timesteps rather than fixed grid points.
The missing gap $\Delta \hat{t}_j$ is essential for our hierarchical imputation procedure. As will be discussed in Sec \ref{sec:imputation_procedure}, we select a subset $\mathbf G \subseteq \hat{\mathbf S}$ to impute at each hierarchy level based on the missing gap of the target time points. The imputation results are denoted as $\mathbf H=\{\h_j\}_{j=1}^{|\mathbf G|}$ with $\h_j \in \R^d$, where $\mathbf H$ is predicted using an imputation model $f$ as $\mathbf H=f(\mathbf G; \mathbf S)$. 

\textbf{Background}\ \ \ A set is a collection that does not impose order among its elements. \textit{Permutation Equivariant} and \textit{Permutation Invariant} are two commonly used terminologies for set modeling.

\textit{(Permutation Equivariant)} Let $f: \mathcal X^N \rightarrow \mathcal Y^N$ be a function, then $f$ is permutation equivariant iff for any permutation $\pi(\cdot)$,  $f(\pi(\mathbf{X}))=\pi(f(\mathbf{X}))$.

\textit{(Permutation Invariant)} Let $f: \mathcal X^{N} \rightarrow \mathcal Y$ be a function, then $f$ is permutation invariant iff for any permutation $\pi(\cdot)$, $f(\pi(\mathbf X)) = f(\mathbf X)$.

\begin{figure}[t!]
\centering
\includegraphics[width=0.9\columnwidth]{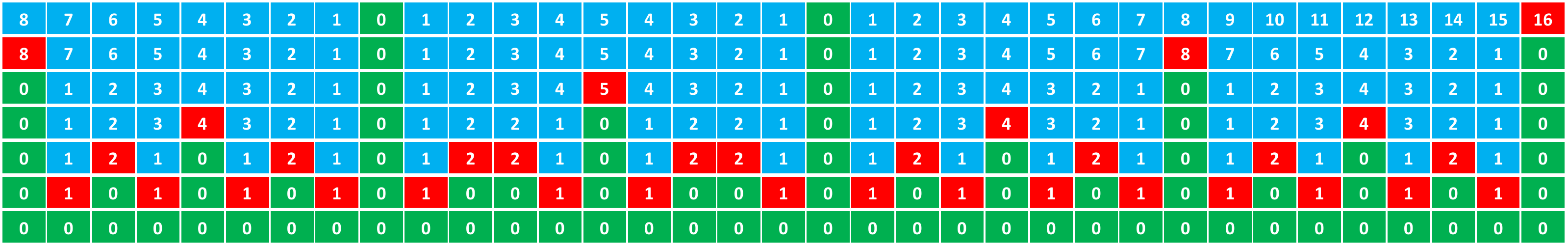} 
\caption{Illustration of the imputation procedure. Blue, green and red boxes respectively represent missing data, observed data, and data to impute next. Numbers inside each box represent the missing gap to the closest observed data and we assume the missing gap of observed data to be 0. Given a time series with 2 observed values and 32 missing values, the imputation procedure starts at the first row and ends at the bottom row where all data are imputed.}
\vspace{-10pt}
\label{fig1}
\end{figure}

\begin{wrapfigure}{R}{0.57\textwidth}
    \vspace{-10pt}
    \begin{minipage}{0.57\textwidth}
      \begin{algorithm}[H]
      \scriptsize
        \caption{\textsc{Imputation Procedure}}
        \begin{algorithmic}[1]
        \scriptsize
        \label{algo_concise}
          \REQUIRE $f_\theta^l$: imputation model at resolution level $l$; $L$: the maximum resolution level; $\mathbf S$: observed data; $\hat{\mathbf S}$: data to impute
\STATE Initialize $\G \leftarrow \emptyset$, $l \leftarrow 0$
\WHILE{$l \leq L$}
\STATE $\hat{\mathbf S}^l \leftarrow \{(\hat{t}_j, \Delta \hat{t}_j) \mid (\hat{t}_j,\Delta \hat{t}_j) \in \hat{\mathbf S}$, $\floor{2^{L-l-1}}<\Delta \hat{t}_j\leq 2^{L-l}\}$
\WHILE{$\hat{\mathbf S}^l$ is not empty}
\STATE $\G \leftarrow \{(\hat{t}_j,\Delta \hat{t}_j)\ |\ (\hat{t}_j,\Delta \hat{t}_j) \in \hat{\mathbf S}^l, \Delta \hat{t}_j=\max_{j}\Delta \hat{t}_j \}$
\STATE $\mathbf H \leftarrow f_{\theta}^{l}(\G;\ \mathbf S)$
\STATE $\mathbf S \leftarrow \mathbf S \cup \mathbf H$, $\hat{\mathbf S} \leftarrow \hat{\mathbf S} \setminus \G$, $\hat{\mathbf S}^l \leftarrow \hat{\mathbf S}^l \setminus \G$
\STATE Update the missing gap information in $\hat{\mathbf S}$ and $\hat{\mathbf S}^l$
\ENDWHILE
\STATE $l \leftarrow l + 1 $
\ENDWHILE
\STATE \textbf{return} Imputation result $\mathbf S$
\end{algorithmic}
\end{algorithm}
\end{minipage}
\vspace{-10pt}
\end{wrapfigure}

\subsection{Hierarchical Imputation}
\label{sec:imputation_procedure} 
In this section, we introduce our proposed hierarchical imputation procedure given the imputation models $f$ at each hierarchy level. We defer the details about the imputation models to Sec.~\ref{sec:model}.

Generative models have been shown to benefit from exploiting the hierarchical structure of data \cite{karras2017progressive, liu2018generating}. Here, we propose to leverage a multi-resolution procedure for time series imputation. Specifically, we divide the missing time points into several hierarchies using their missing gaps (i.e. the closest distance to an observed time point). Intuitively, missing data that are far from the observed data are more difficult to impute. According to their missing gaps, we can either impute from small gap time points to large gap ones or vice versa. Empirically, we find starting from large missing gaps works better (as also indicated by \cite{liu2019naomi}). Given the imputed values at the current hierarchy level, the imputation at the higher hierarchy level will depend on those values. Note that the hierarchical imputation inevitably introduces some recurrent dependencies among missing time points, but since the number of hierarchy levels is typically much smaller than the number of missing time points, the error compounding problem of NRTSI is not as severe as the sequential models. To further reduce the imputation error at each level, we utilize a separate imputation model $f_\theta^l$ for each level $l$. The imputation model takes in all the observed data and the already imputed data at lower hierarchy levels as a set to impute the missing data at the current level. At each hierarchy level, the missing time points are imputed in descending order of their missing gaps. Please refer to Algorithm \ref{algo_concise} for the proposed imputation procedure (for a detailed version please see Appendix \ref{append_sec_train}). We illustrate the imputation procedure in Figure \ref{fig1} where at each hierarchy level NRTSI can impute multiple missing points in parallel thanks to the set representation of time series. Note that NRTSI can also handle irregularly-sampled time series, though we only show the imputation procedure on regularly sampled ones in Figure \ref{fig1} for convenience.

Similar to our proposed hierarchical imputation procedure, NAOMI \cite{liu2019naomi} also explore the multiresolution structure of time series. NAOMI first encodes the observed data using a bidirectional RNN. 
Then the missing data are imputed by several multiresolution decoders followed by updating the RNN hidden state to take the newly imputed data into account. To make the hidden state updates efficient, NAOMI chooses to first impute the data with the largest missing gap located between the two earliest observed time points. Therefore, the first imputed data might not have the largest missing map throughout the whole time series. In contrast, NRTSI always imputes the data with the global largest missing gap first thanks to our set formulation. Also, NAOMI only imputes a single time point at each step, while NRTSI can impute multiple time points together at each hierarchy level.

\subsection{Imputation Model $f_\theta$} 
\label{sec:model}
In this section, we describe the imputation model $f_\theta$ used at each hierarchy level. The model takes in a set of known time points $\mathbf{S}$ (either observed or previously imputed at lower hierarchy levels) and imputes the values for a set of target time points $\mathbf{G}$. To respect the unordered nature of sets, the imputation model $f_\theta$ is designed to be permutation equivariant w.r.t. $\mathbf G$ and permutation invariant w.r.t. $\mathbf S$, i.e.
$
    \rho(\mathbf H) = f_\theta(\rho(\mathbf{G}); \pi(\mathbf{S})),
$
where $\pi$ and $\rho$ represent two arbitrary permutations. 
Theoretically, any permutation equivariant architecture can be seamlessly plugged in as long as we use a permutation invariant embedding on $\mathbf S$. Representative architectures include DeepSets \cite{zaheer2017deep}, ExNODE \cite{li2020exchangeable} and Transformers \cite{vaswani2017attention,lee2019set}. In this work, we adopt the multi-head self-attention mechanism in Transformers for its established strong capability of modeling high-order interactions. 

\textbf{Time Encoding Function}\ \ \  Transformers do not require the sequences to be processed in sequential order. Originally, the input to Transformers is a set of word embeddings and positional embeddings. Although the input is an unordered set, the positions of word embeddings are informed by their corresponding positional embeddings. Similar to the positional embedding in Transformers, we embed the time using a function $\phi$ to transform a time stamp $t\in\R^+$ to a higher dimensional vector $z=\phi(t)\in\R^{\tau}$, where
\begin{equation}
z_{2k}(t) := \text{sin}\left(\frac{t}{\nu^{2k/\tau}}\right) \qquad z_{2k+1}(t) := \text{cos}\left(\frac{t}{\nu^{2k/\tau}}\right)
\label{eq:time}
\end{equation}
with $k\in\{0, \ldots, \tau/2\}$, $\tau$ denoting the dimensionality of the time embedding, and $\nu$ representing the expected maximum time scale for a given data set. This time embedding function is first proposed in \cite{horn2020set}. Though the absolute time is used in \eqref{eq:time}, the relative time that is conceptually similar to the relative positional embedding \cite{shaw2018self} for Transformers can also be used, and we leave it for future work.

\textbf{Implementation}\ \ \ At each hierarchy level $l$, a subset of missing time points $\mathbf{G}$ are first selected based on their missing gaps $\Delta \hat{t}_j$, then the imputation model $f_\theta^l$ imputes the missing values by $\mathbf{H} = f^l_\theta(\mathbf{G};\mathbf{S})$, where $\mathbf{S}=\{(\phi(t_i), \x_i)\}$ and $\mathbf{G}=\{\phi(\hat{t}_j)\}$. Note that $\Delta \hat{t}_j$ are ignored here since they are 
only used to define the hierarchy levels (see Algorithm \ref{algo_concise}). The elements in $\mathbf S$ and $\mathbf G$ are transformed to tensors by concatenating the data $\x \in \R^d$ and the time encoding vector $\phi(t) \in \R^{\tau}$ computed via \eqref{eq:time}. Since the elements in $\mathbf G$ do not contain $\x$, we use $d$-dimensional zero vectors $\mathbf{0} \in \R^d$ as placeholders. We also add a binary scalar indicator to distinguish missing values and observed values. That is, 
\begin{equation}
\begin{aligned}
    s_i = (\phi(t_i), \x_i) \in \mathbf{S} \quad &\rightarrow \quad s_i = [\phi(t_i), \x_i, 1] \in \R^{\tau+d+1},\\
    g_j = \phi(\hat{t}_j) \in \mathbf{G} \quad &\rightarrow \quad g_j = [\phi(\hat{t}_j), \mathbf{0}, 0] \in \R^{\tau+d+1},
\end{aligned}
\end{equation}
where $[\cdot]$ represents the concatenation operation. Now that the elements in $\mathbf{S}$ and $\mathbf{G}$ are all transformed to vectors with same dimensionality, we can combine them into one set and pass it through the permutation equivariant imputation model $f_\theta$, i.e. $\mathbf{H}=f_\theta(\mathbf{S} \cup \mathbf{G})$. Specifically, we implement $f_\theta$ by the following steps: 
\begin{equation}
\label{eq:flow}
    \mathbf{S}^{(1)} \cup \mathbf{G}^{(1)} = f_{\text{in}}(\mathbf{S} \cup \mathbf{G}) \qquad
    \mathbf{S}^{(2)} \cup \mathbf{G}^{(2)} = f_{\text{enc}}(\mathbf{S}^{(1)} \cup \mathbf{G}^{(1)}) \qquad
    \mathbf{H} = f_{\text{out}}(\mathbf{G}^{(2)}).
\end{equation}

\begin{wrapfigure}{R}{0.35\textwidth}
\centering
\vspace{-15pt}
\includegraphics[width=0.35\columnwidth]{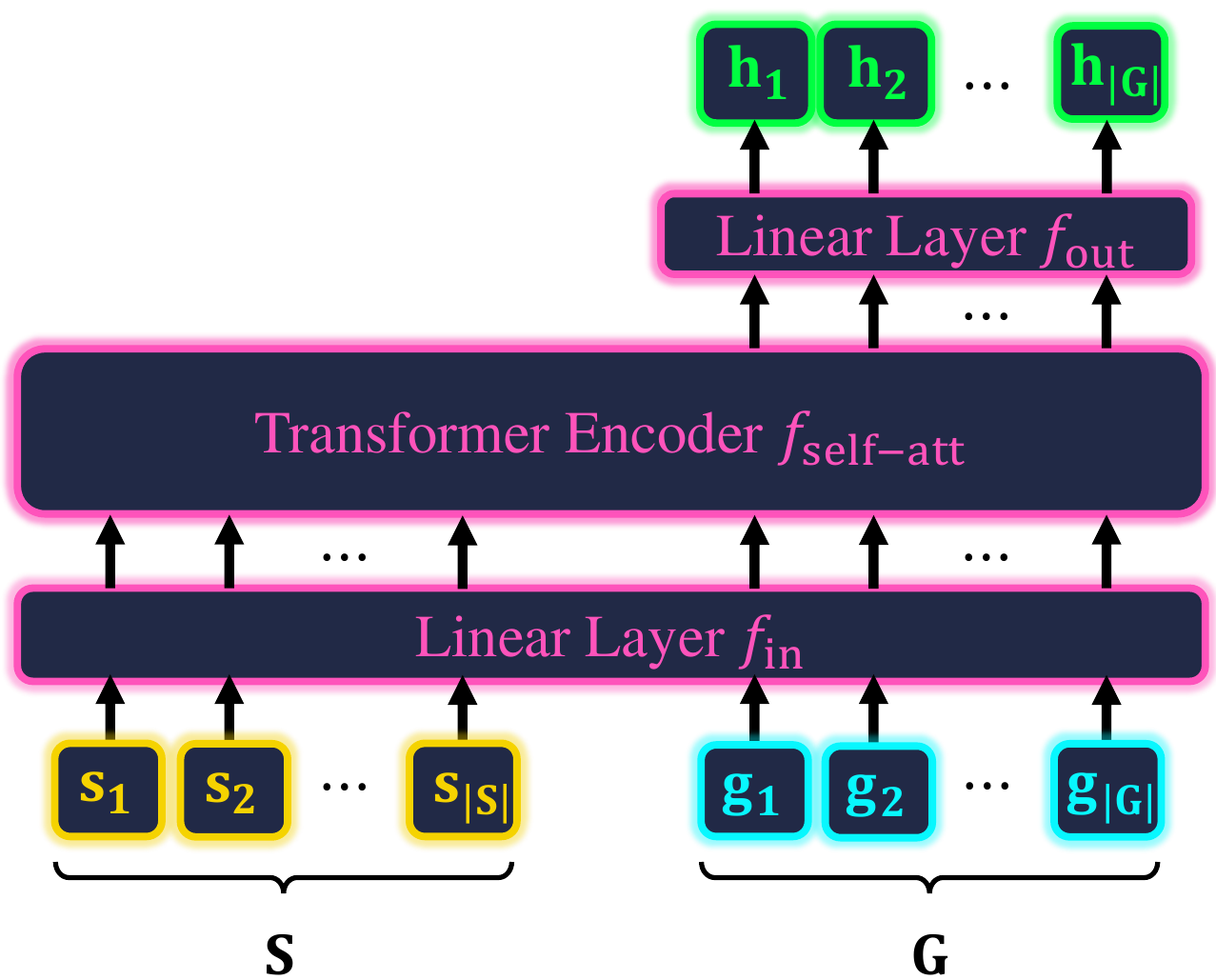}
\vspace{-17pt}
\caption{Imputation model.}
\vspace{-15pt}
\label{fig2}
\end{wrapfigure}
At the first step, a linear layer $f_{\text{in}}: \R^{\tau+d+1} \rightarrow \R^{d_{h}}$ maps the input data to a high-dimensional space. Then, a permutation equivariant Transformer encoder $f_{\text{enc}} : \R^{d_h} \rightarrow \R^{d_h}$ is used to model the interactions between $\mathbf{S}^{(1)}$ and $\mathbf{G}^{(1)}$. The Transformer encoder is composed of multiple alternating multi-head self-attention layers and feedforward layers, allowing the elements in $\mathbf{S}^{(1)}$ and $\mathbf{G}^{(1)}$ to effectively exchange information, i.e. $\mathbf{G}^{(1)}$ can attend to $\mathbf{S}^{(1)}$ to gather the observed information and $\mathbf{S}^{(1)}$ can be informed about what timestamps to impute by attending to $\mathbf{G}^{(1)}$.  Finally, the imputation results $\mathbf{H}$ are obtained via another linear layer $f_{\text{out}}: \R^{d_h} \rightarrow \R^{d}$ on $\mathbf{G}^{(2)}$. The architecture of the imputation model is illustrated in Figure \ref{fig2}. Since the set elements are processed individually by the linear layers and the attention mechanism \cite{vaswani2017attention,luong2015effective,NEURIPS2020_35464c84} in $f_{\text{enc}}$ is permutation equivariant, the overall model $f_\theta$ is permutation equivariant w.r.t. $\mathbf{G}$ and permutation invariant w.r.t. $\mathbf{S}$.

\textbf{Training Objective}\ \ \  We denote our imputation model with learnable parameters $\theta$ at level $l$ as $f^l_{\theta}$, which include the two linear layers and the Transformer encoder. The optimization objective is
 \begin{align}
 \label{loss}
     \min_{\theta}\E_{\mathbf{G}\sim p(\mathbf{G}), \mathbf{S}\sim p(\mathbf{S}),\Y\sim p(\Y)} \bigg[\frac{1}{|\mathbf{G}|}\sum_{j=1}^{|\mathbf{G}|}\mathcal{L}(\h_j, \y_j)\bigg],
 \end{align}
 where $\h_j \in \mathbf{H} = f^l_{\theta}(\mathbf{G}\cup\mathbf{S})$ is an imputed data and $\y_j \in \Y$ denotes the corresponding ground truth imputation target. For deterministic datasets, we use Mean Square Error (MSE), i.e. $\mathcal{L}(\h_j, \y_j)=||\h_j - \y_j||^{2}_{2}$. For stochastic datasets, we minimize the negative log-likelihood of a Gaussian distribution with diagonal covariance, i.e. 
 \begin{align}
 \label{eq:stochastic}
     \mathcal{L}(\h_j, \y_j)=-\text{log}\,\mathcal{N}(\y_j|\mu(\h_j),\text{diag}(\sigma(\h_j))),
 \end{align}
 where $\mu:\R^d \rightarrow \R^d$ and $\sigma: \R^d \rightarrow \R^d$ are two linear mappings.

When training our model, we follow a similar procedure in Algorithm \ref{algo_concise}. The only difference is at line 7 of Algorithm \ref{algo_concise}, where we regard the ground truth imputation target $
\Y$ as observed data for the subsequent training steps rather than the imputed $\mathbf{H}$. This resembles the Teacher Forcing algorithm \cite{williams1989learning} used to stabilize the training of RNNs. The detailed training procedure is given in Appendix \ref{append_sec_train}.

\subsection{Model Variants}
\label{sec:variant}
Above, we describe the general formulation of NRTSI. Next, we show it naturally handles irregularly sampled time series, stochastic time series, and partially observed time series in a unified framework.

\textbf{Irregularly-sampled Time Series}\ \ \ For regularly-sampled time series, there are typically multiple time points with the same missing gaps. For irregularly-sampled time series, however, missing time points tend to have unique missing gaps. Therefore, imputing from large missing gaps to small missing gaps will reduce to an autoregressive model, which could incur a high computation demand. Instead, we modify line 5 in Algorithm \ref{algo_concise} by imputing time points with similar missing gaps together, i.e. $\G \leftarrow \{\ (\hat{t_j},\hat{\Delta t_j})\ |\ (\hat{t_j},\hat{\Delta t_j}) \in \hat{\mathbf S}^l, \hat{\Delta t_j} \in \left(a,b\right]\}$ where $a = \max_{j}\hat{\Delta t_j} - \Delta$, $b = a + \Delta$, and $\Delta\in \R^+$ is a hyperparameter.


\textbf{Stochastic Time Series}\ \ \ Some time series are stochastic, e.g. given sparsely observed football player locations there exists multiple possible trajectories (see Figure \ref{fig:nfl}). For deterministic datasets, we can impute all the data in $\mathbf G$ simultaneously as no sampling is required. For stochastic datasets, however, we need to sample from the distribution \eqref{eq:stochastic} and the samples may be incongruous if we sample for all the elements in $\mathbf G$ simultaneously. This is because in \eqref{eq:stochastic} we learn an individual distribution for each element rather than a joint distribution for all the elements. To solve this problem, we propose to impute data with large missing gaps one by one. Based on the observation that missing data with small missing gaps are almost deterministic, they can be imputed simultaneously in parallel to avoid the high complexity of sampling sequentially. In practice, we find this modification does not significantly slow down the imputation process as there are much fewer data with large missing gaps compared to the ones with small missing gaps.

\textbf{Partially Observed Time Series}\ \ \  The discussion so far assumes that the multivariate data at a timestep is either completely observed or completely missing along all the dimensions. In practice, a timestep may be only partially observed (with a subset of features missing at that time). Our hierarchical imputation procedure can be easily extended to this partially observed scenario.
However, missing gaps may no longer be the driving factor for an effective imputation order, as the number of dimensions observed may affect the effectiveness of imputations.
Therefore, we modify Algorithm \ref{algo_concise} to impute the timesteps with the most missing dimensions first rather than the timesteps with the largest missing gap. 
We also modify the data representation to $s_i=[\phi(t_i),\x_i, \mathbf{m}_i] \in \R^{d+\tau+d}$ where $\mathbf{m}_i \in \{0,1\}^d$ is a binary mask indicating which dimensions are observed.
\begin{table*}[t]
\caption{Quantitative comparison on Billiards dataset. Statistics closer to the expert indicate better performance.}
\vspace{-10pt}
\label{table:billiards_stats}
\begin{center}
\begin{scriptsize}
\begin{tabular}{c|cccccccc}
\toprule
\textbf{Models}  & \texttt{Linear}  & \texttt{KNN}  & \texttt{GRUI}  & \texttt{MaskGAN} & \texttt{SingleRes}  & \texttt{NAOMI} & NRTSI & Expert    \\
\midrule
\textbf{Sinuosity} & 1.121 & 1.469 & 1.859 & 1.095 & 1.019 & 1.006 & \textbf{1.003} & 1.000 \\
\textbf{step change ($\times 10^{-3}$)} & $\textbf{0.961}$ & 24.59 & 28.19 & 15.35 & 9.290 & 7.239 & 5.621 & 1.588 \\
\textbf{reflection to wall} & 0.247 & 0.189 & 0.225 & 0.100 & 0.038 & 0.023 & \textbf{0.021} & 0.018  \\
\textbf{L2 loss ($\times 10^{-2}$)} & 19.00 & 5.381 & 20.57 & 1.830 & 0.233 & 0.067 & \textbf{0.024} & 0.000 \\
\bottomrule
\end{tabular}
\end{scriptsize}
\end{center}
\vspace{-0.1in}
\end{table*}
\begin{figure}[t!]
\vspace{-5pt}
\center
\subfigure[Observed]{\includegraphics[width=0.15\linewidth]{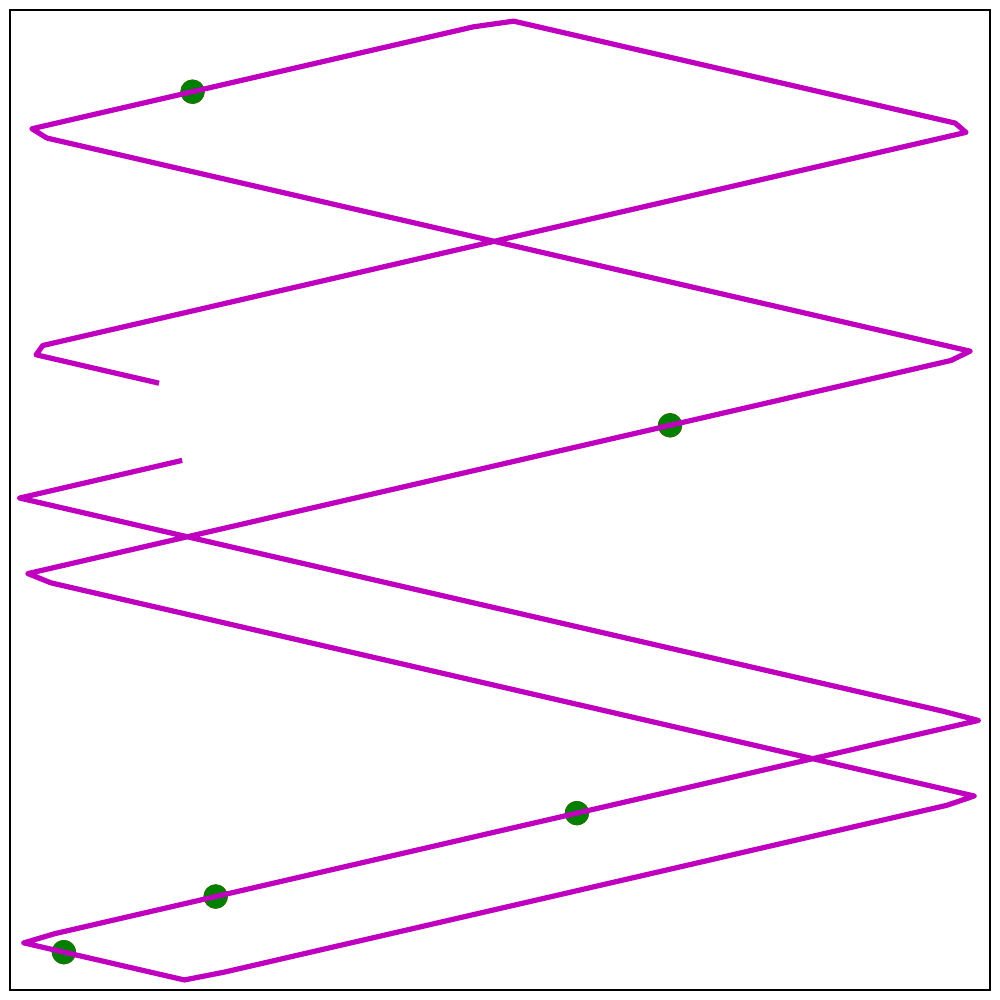}}
\subfigure[Gap 16]{\includegraphics[width=0.15\linewidth]{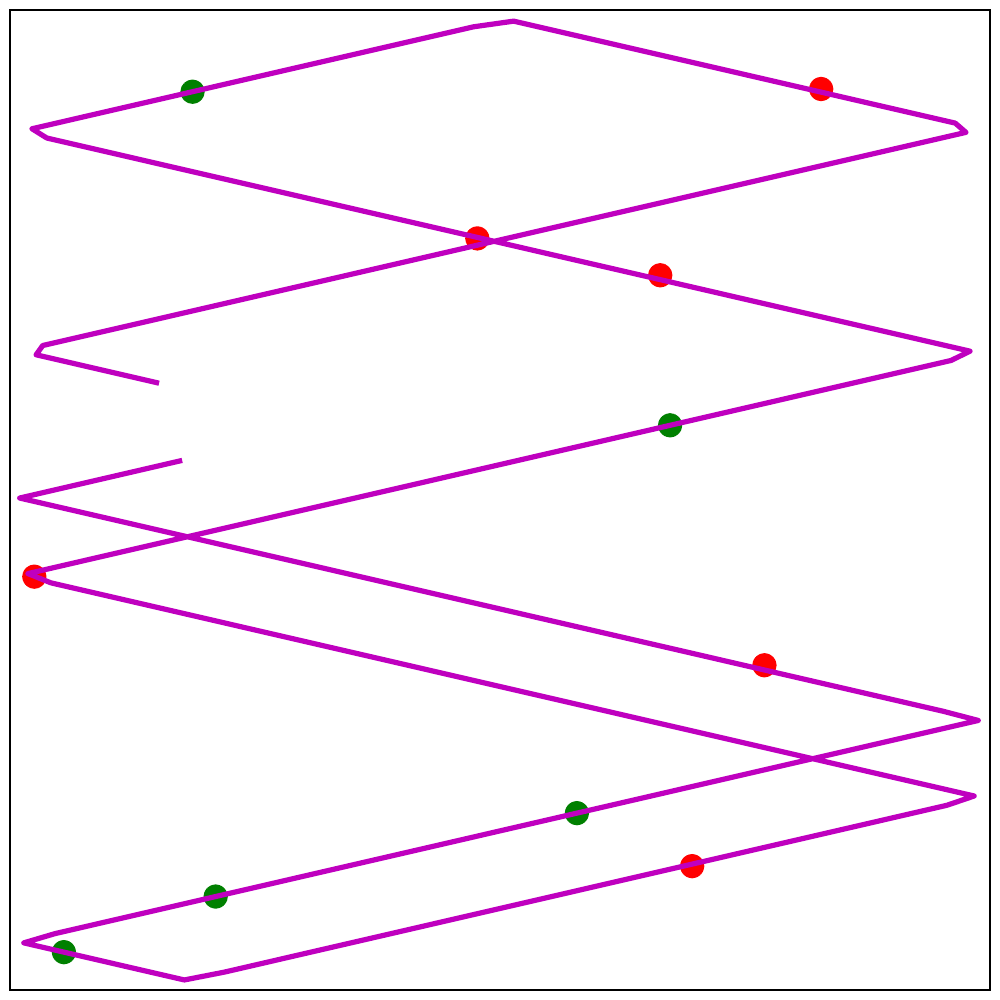}}
\subfigure[Gap 8]{\includegraphics[width=0.15\linewidth]{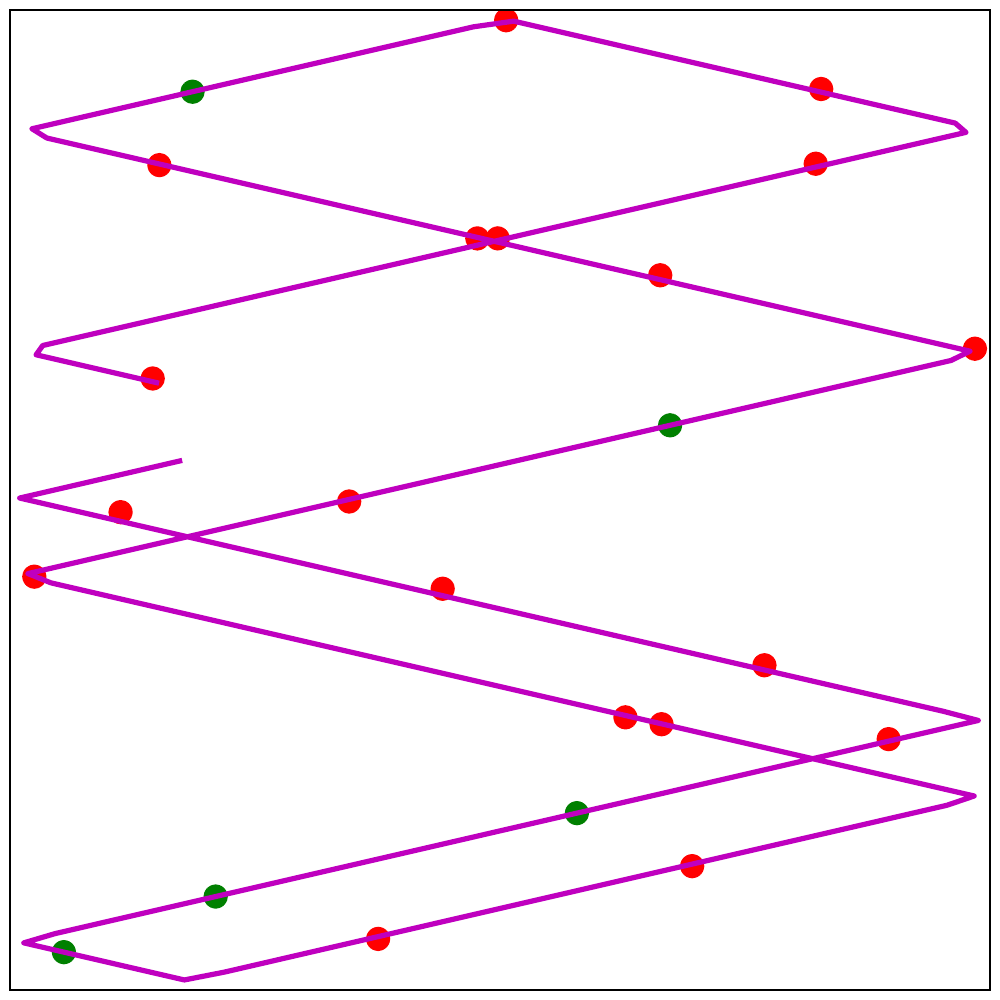}}
\subfigure[Gap 4]{\includegraphics[width=0.15\linewidth]{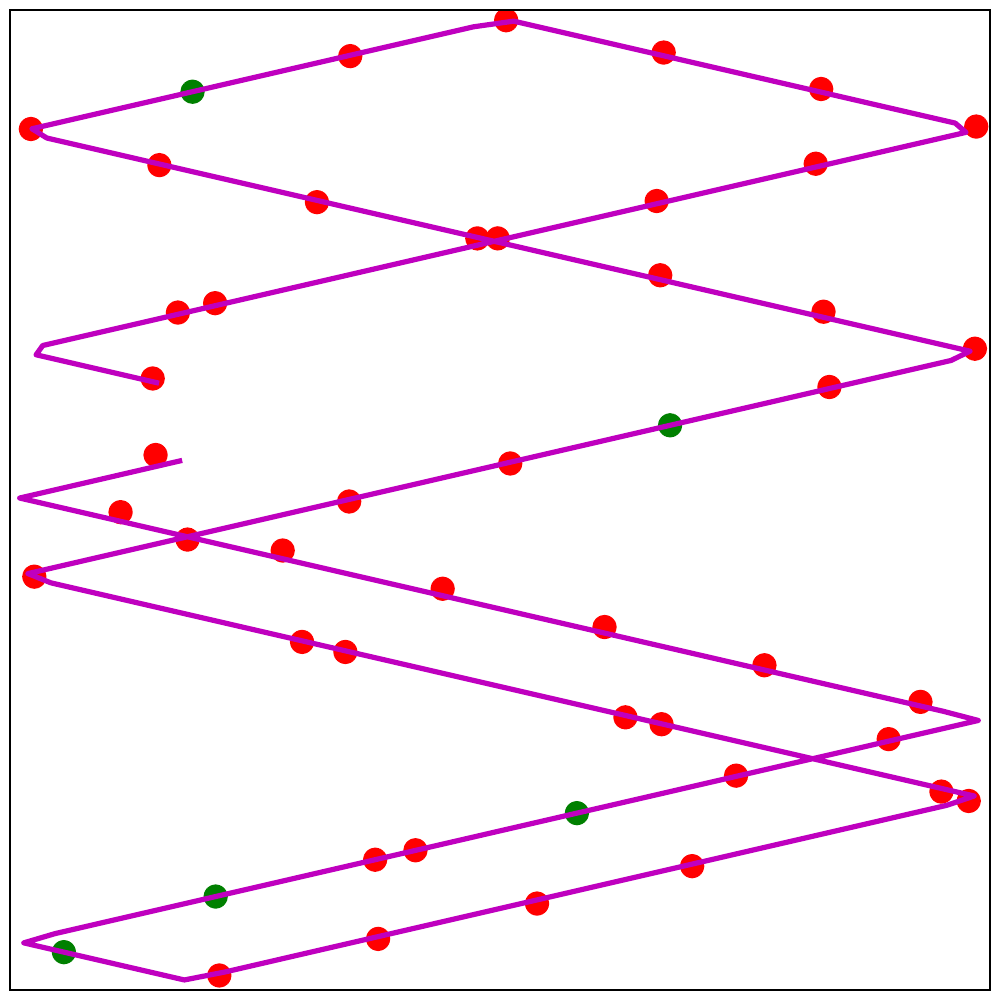}}
\subfigure[Gap 2]{\includegraphics[width=0.15\linewidth]{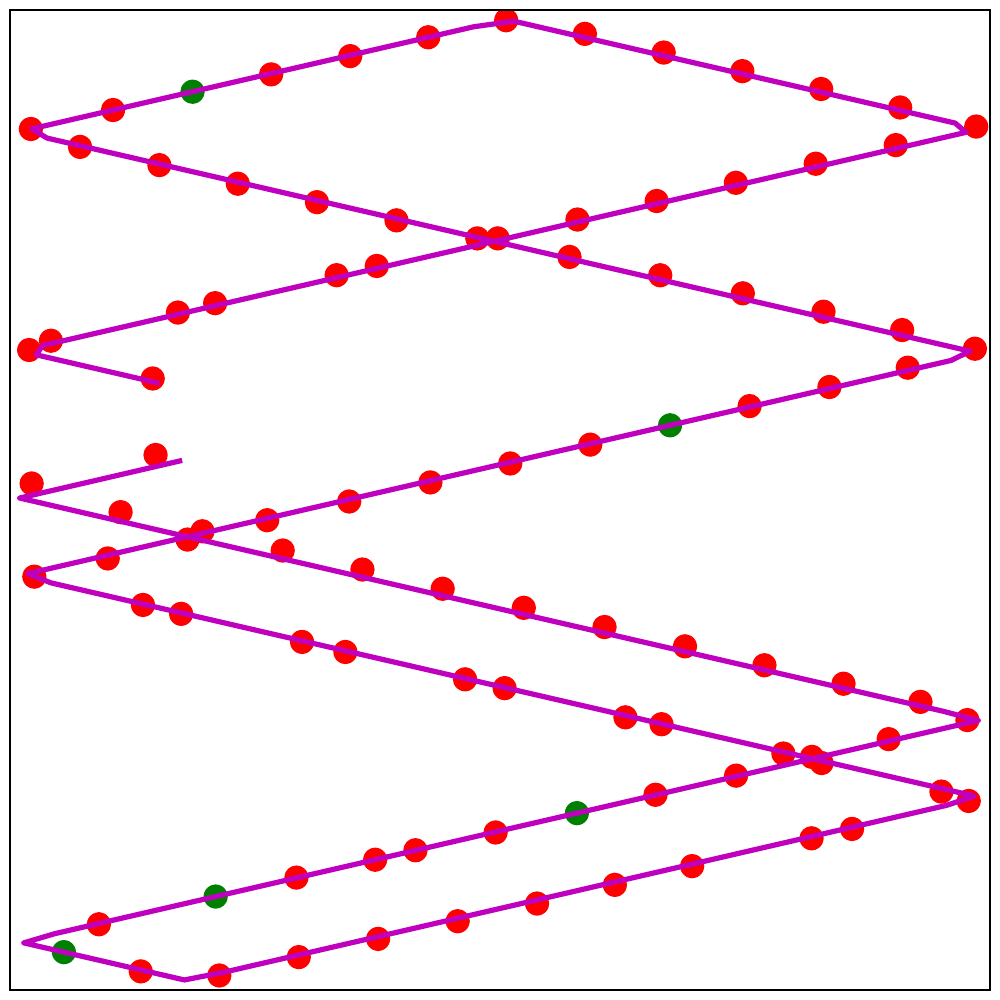}}
\subfigure[Gap 1]{\includegraphics[width=0.15\linewidth]{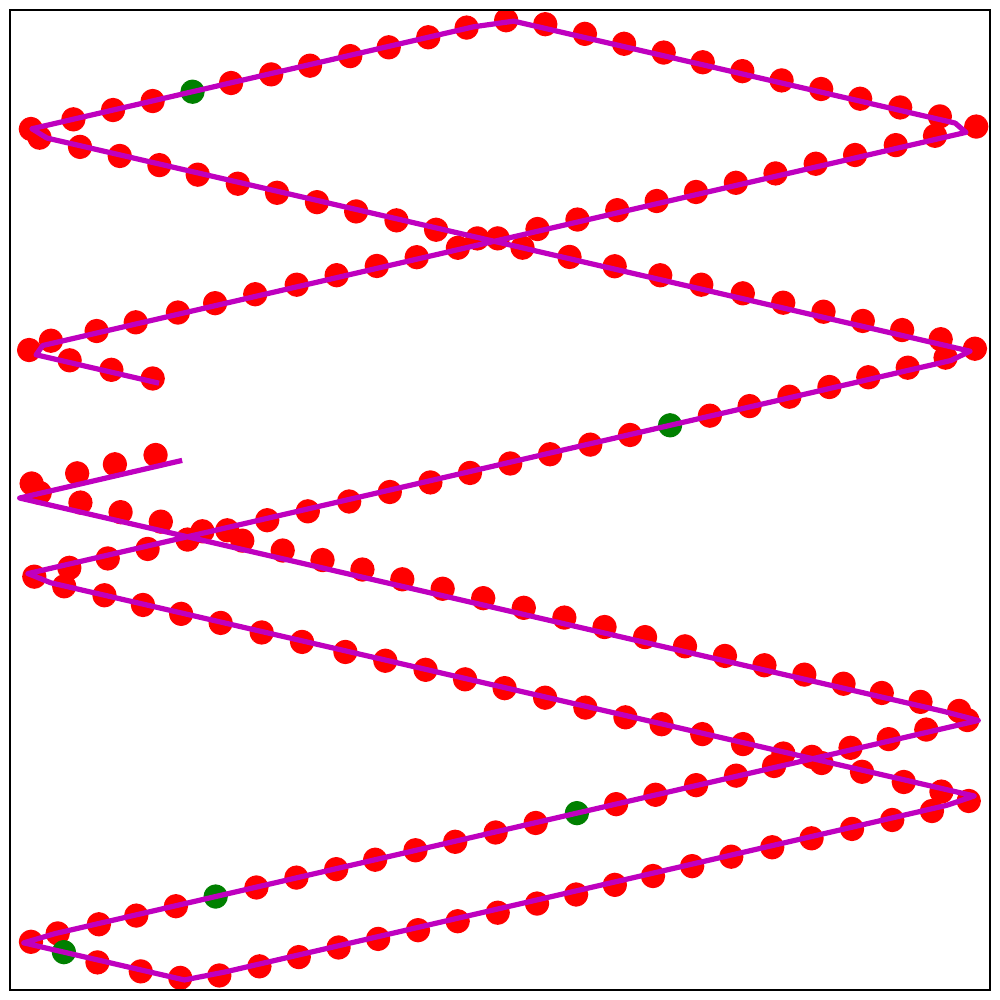}}
  \caption{Imputation procedure on the Billiards dataset. The red points denote imputed data while the green points denote observed data. The purple solid line is the ground-truth trajectory. The initial observed data is shown in (a), the imputed data with missing gaps 16 to 1 are shown in (b)-(f). We omit the intermediate results at missing gaps 15, 7, 6, and 3 due to the limitation of space.}
  \label{fig:billiards}
 \vspace{-10pt}
\end{figure}
\section{Experiments}
In this section, we evaluate NRTSI on popular time series imputation benchmarks against strong baselines. Extensive hyperparameter searching is performed for all the baselines and their number of parameters are reported in Appendix \ref{sec:baseline_hyper}. For fair comparisons, we use the same training/validation/testing splits for all the methods. To closely match the setting of the baselines, some datasets might not have a validation set. We also 
use the same method to randomly mask out data for all the methods. Due to the space limit, we defer the details about datasets, architectures, and training procedures to Appendix.
\subsection{Billiards Ball Trajectory}
\label{sec:billiard}
\textbf{Dataset}\ \ \  Billiards dataset \cite{salimans2016weight} contains 4,000 training and 1,000 testing sequences, which are regularly-sampled trajectories of Billiards balls in a rectangular world. Each ball is initialized with a random position and a random velocity and the trajectory is rolled out for 200 timesteps. All balls have a fixed size and uniform density, and friction is ignored.


\begin{wrapfigure}{R}{0.5\textwidth}
\center
\vspace{-20pt}
\subfigure{\includegraphics[width=0.24\linewidth]{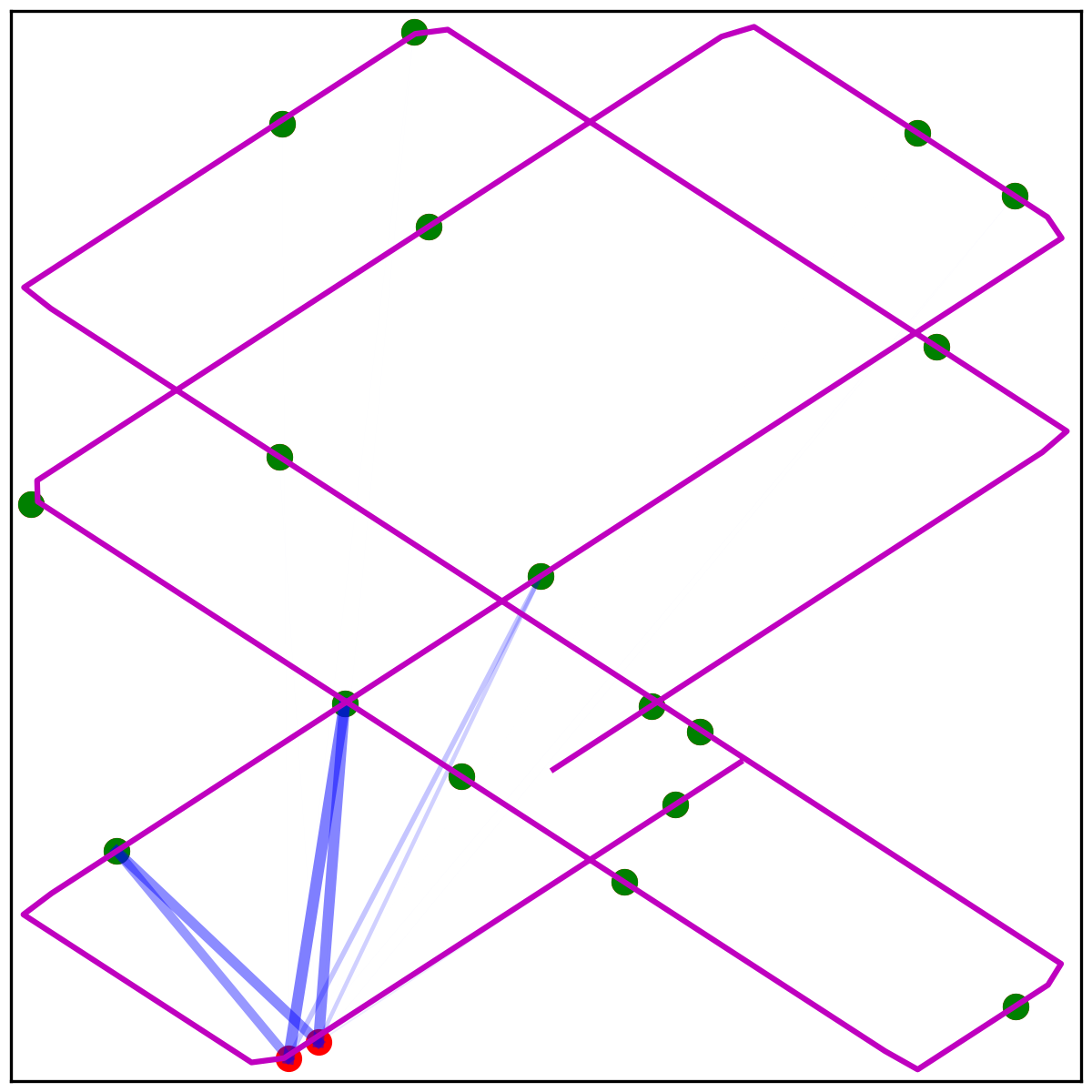}}
\subfigure{\includegraphics[width=0.24\linewidth]{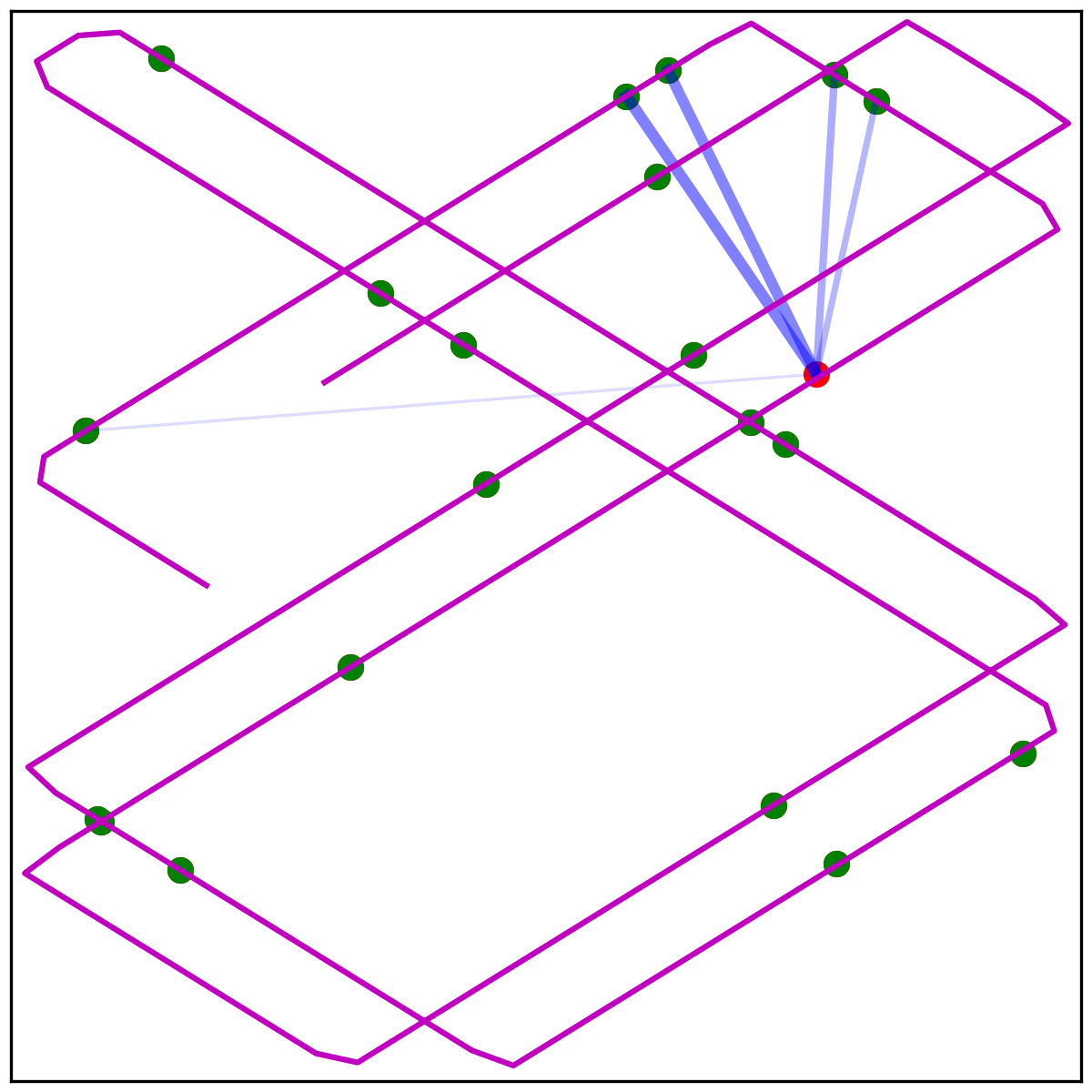}}
\subfigure{\includegraphics[width=0.24\linewidth]{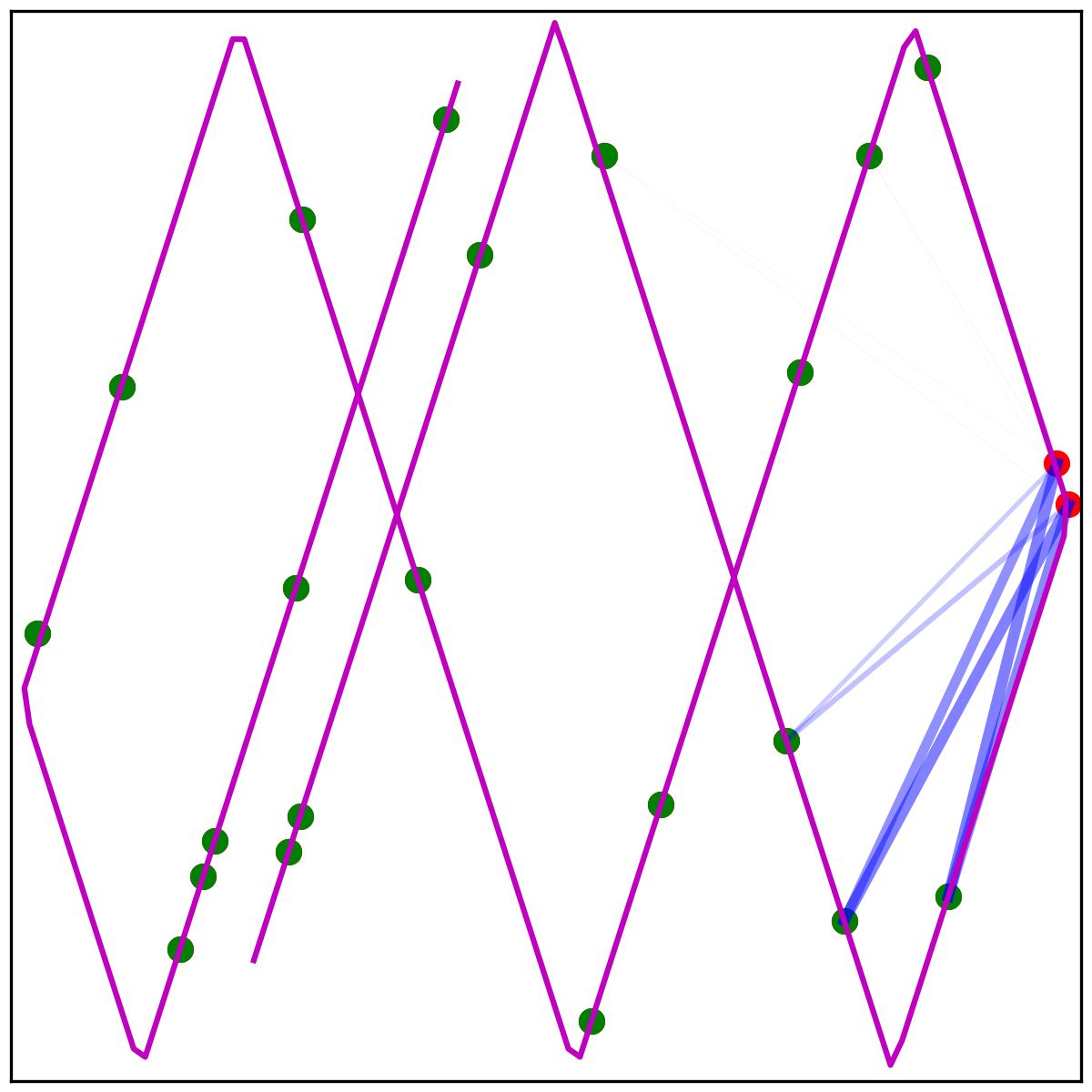}}
\subfigure{\includegraphics[width=0.24\linewidth]{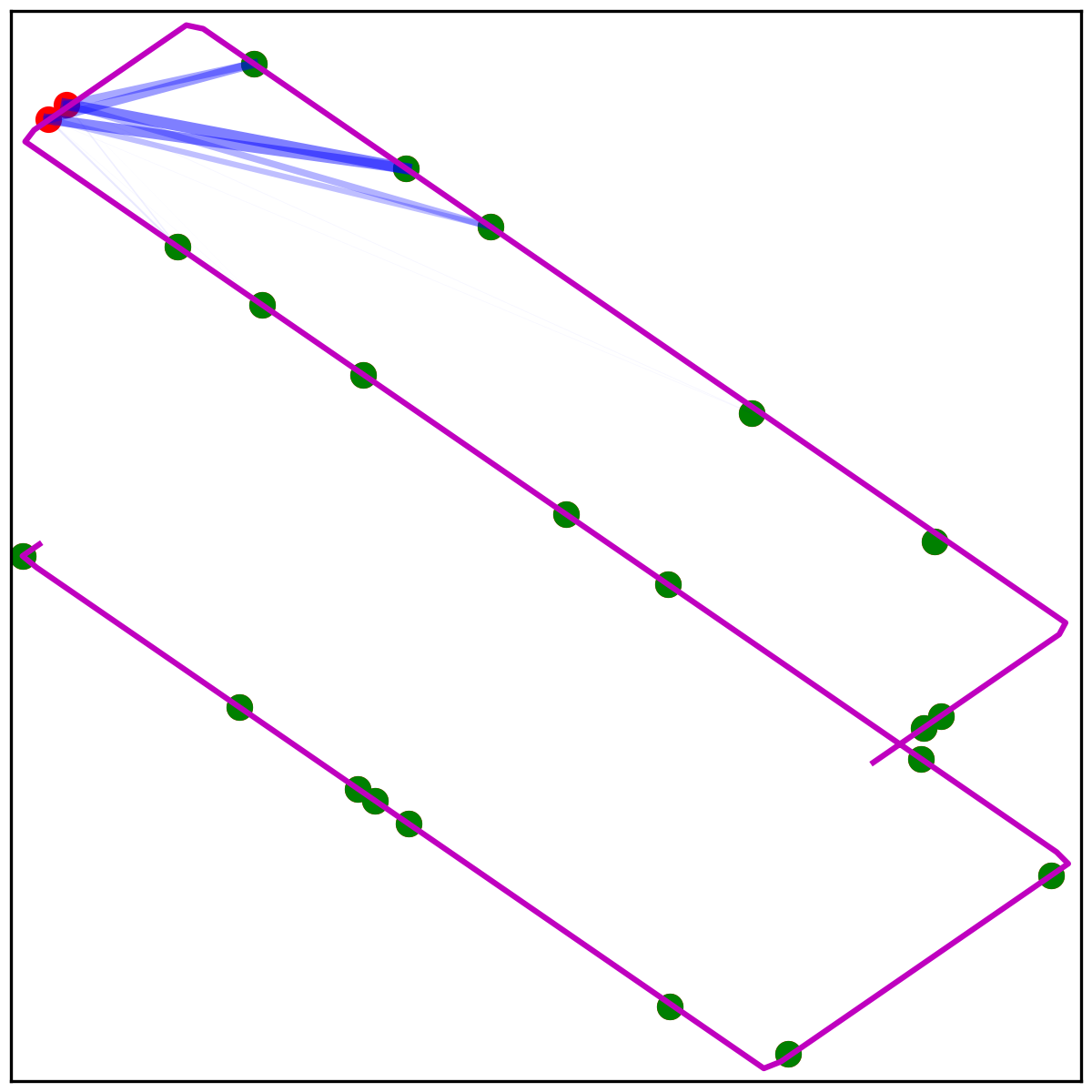}}
  \caption{Visualization of one attention head. Red points are the imputed data, green points are the observed data. The softmax attention weights are visualized via the blue lines. The wider and less transparent the blue lines are, the larger the attention weights.}
  \label{fig:att}
\vspace{-8pt}
\end{wrapfigure}

\textbf{Results}\ \ \  On this dataset, we train and evaluate using MSE loss between imputed values and ground truth. We also report three additional metrics, \textit{Sinuosity}, \textit{step change} and \textit{reflection to wall}, as reported in \cite{liu2019naomi} to assess the realism of the imputed trajectories. We closely follow the setting in \cite{liu2019naomi} and compare to all baselines mentioned there. Please refer to \cite{liu2019naomi} for details about the baseline models. Quantitative results are reported in Table \ref{table:billiards_stats}, where \texttt{Expert} denotes the ground truth trajectories. Following \cite{liu2019naomi}, we randomly select 180 to 195 timesteps as missing for each trajectory and repeat the test set 100 times to make sure a variety of missing patterns are covered. Statistics that are closer to the expert ones are better. From Table \ref{table:billiards_stats}, we can see NRTSI reduces the $L_2$ loss by 64\% compared to NAOMI and compares favorably to baselines on all metrics except \textit{step change}, as linear interpolation maintains a constant step size change by design, which matches with the underlying dynamics of this dataset. In Figure \ref{fig:billiards}, we visualize the hierarchical imputation procedure on one example. Initially, only five points are observed (shown in green). Then, NRTSI imputes the time points with the largest missing gap (16) based on those observed points. It then hierarchically imputes the remaining missing time points with smaller missing gaps. Note that NRTSI imputes multiple time points together at each hierarchy level. The final imputed trajectory not only aligns well with the ground truth but also maintains a constant speed and straight lines between collisions. To better understand how NRTSI imputes missing data, in Figure \ref{fig:att}, we visualize the attention maps of one attention head. We can see this attention head models short-range interactions in a single direction. Additional attention patterns are presented in Appendix \ref{sec:vis_att}.


\begin{wraptable}{R}{0.45\textwidth}
\vspace{-20pt}
    \caption{Traffic data L2 loss ($\times 10^{-4}$).}
    \vspace{-10pt}
    \begin{center}
    \setlength{\tabcolsep}{2.5pt}
    \begin{scriptsize}
    \begin{tabular}{ccccccc}
        \toprule
        \texttt{Linear}  & \texttt{GRUI} & \texttt{KNN} & \texttt{MaskGAN} & \texttt{SingleRes} & \texttt{NAOMI} & NRTSI  \\
        \midrule
         15.59 & 15.24 & 4.58 & 6.02 & 4.51 & 3.54 & \textbf{3.22} \\
        \bottomrule
    \end{tabular}
    \end{scriptsize}
    \end{center}
    \label{table:traffic_l2}
\vspace{-10pt}
\end{wraptable}
\subsection{Traffic Time Series}
\textbf{Dataset}\ \ \  The PEMS-SF traffic \cite{Dua:2019} dataset contains 267 training and 173 testing sequences of length 144 (regularly-sampled every 10 mins throughout the day). It is a multivariate dataset with 963 dimensions at each time point, which represents the freeway occupancy rate from 963 sensors.

\textbf{Results}\ \ \  Similar to the Billiards experiment above, we train and evaluate using MSE loss. We also compare to the same set of baselines as the Billiards experiment. Following \cite{liu2019naomi}, we generate masked sequences with 122 to 140 missing values at random and repeat the testing set 100 times. The $L2$ losses are reported in Table \ref{table:traffic_l2}. We can see NRTSI outperforms all the baselines.
\begin{wraptable}{R}{0.35\textwidth}
\vspace{-30pt}
    \caption{MuJoCo Simulation Data MSE loss ($10^{-3}$) comparison.}
    \vspace{-10pt}
    \begin{center}
    \setlength{\tabcolsep}{4pt}
    \begin{scriptsize}
    \begin{tabular}{c|cccc}
        \toprule
        Method  & 10\% & 20\% & 30\% & 50\%  \\
        \midrule
        \texttt{RNN GRU-D} & 19.68 & 14.21 & 11.34 & 7.48 \\
        \texttt{ODE-RNN} & 16.47 & 12.09 & 9.86 & 6.65 \\
        \texttt{NeuralCDE} & 13.52 & 10.71 & 8.35 & 6.09 \\
        \texttt{Latent-ODE} & \textbf{3.60} & 2.95 & 3.00 & 2.85 \\
        \texttt{NAOMI} & 4.42 & 2.32 & 1.46 & 0.93 \\
        NRTSI & 4.06 & \textbf{1.22} & \textbf{0.63} & \textbf{0.26} \\
        \bottomrule
    \end{tabular}
    \end{scriptsize}
    \end{center}
    \vspace{-20pt}
    \label{table:mujoco}
\end{wraptable}

\subsection{MuJoCo Physics Simulation} 
\textbf{Dataset}\ \ \  MuJoCo is a physical simulation dataset created by \cite{rubanova2019latent} using the ``Hopper'' model from the Deepmind Control Suite \cite{tassa2018deepmind}. Initial positions and velocities of the hopper are randomly sampled such that the hopper rotates in the air and falls on the ground. The dataset is 14-dimensional and contains 10,000 sequences of 100 regularly-sampled time points for each trajectory.

\textbf{Results}\ \ \ MSE loss is used to train and evaluate NRTSI. Baseline models include Latent-ODE \cite{rubanova2019latent}, ODE-RNN \cite{rubanova2019latent}, GRU-D \cite{che2018recurrent}, NeuralCDE \cite{kidger2020neural} and NAOMI \cite{liu2019naomi}. We report the MSEs with different observation rates in Table \ref{table:mujoco}. NRTSI compares favorably to all baselines with 20\%, 30\% and 50\% observed data. When only 10\% data are observed, NRTSI is comparable to Latent-ODE and NAOMI.

\subsection{Football Player Trajectory}
\textbf{Dataset}\ \ \ This dataset is from the NFL Big Data Bowl 2021 \cite{nfl}, which contains 9,543 time series with 50 regularly-sampled time points for each trajectory. The sampling rate is fixed at 10 Hz. Each time series contains the 2D trajectories of 6 offensive players and is therefore 12-dimensional. During training and testing, we treat all players in a trajectory equally and randomly permute their orders. 8,000 trajectories are randomly selected for training and the remaining ones are used for testing.

\begin{figure}[ttt!]
\begin{tabular}{cc}
\begin{minipage}{.5\textwidth}
    \centering
    \subfigure[5 timesteps observed]{\includegraphics[width=0.49\linewidth]{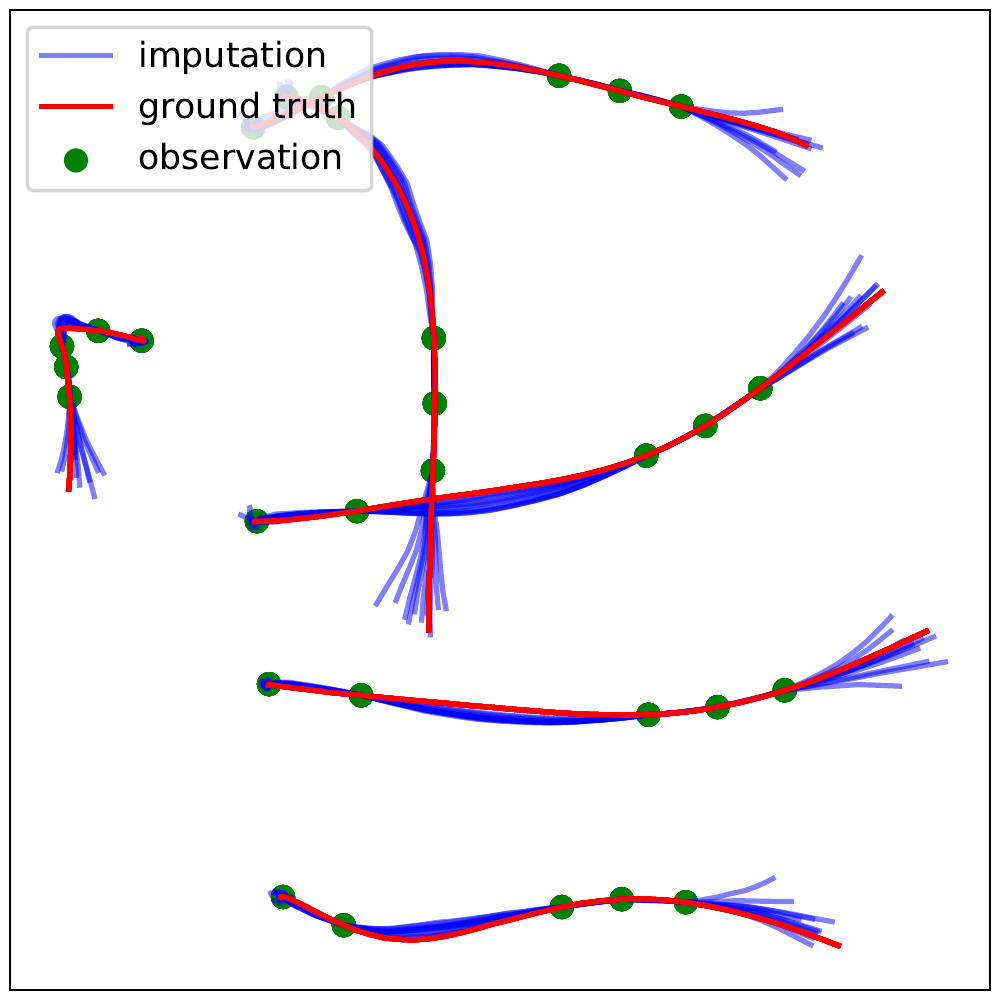}}
    \subfigure[2 timesteps observed]{\includegraphics[width=0.49\linewidth]{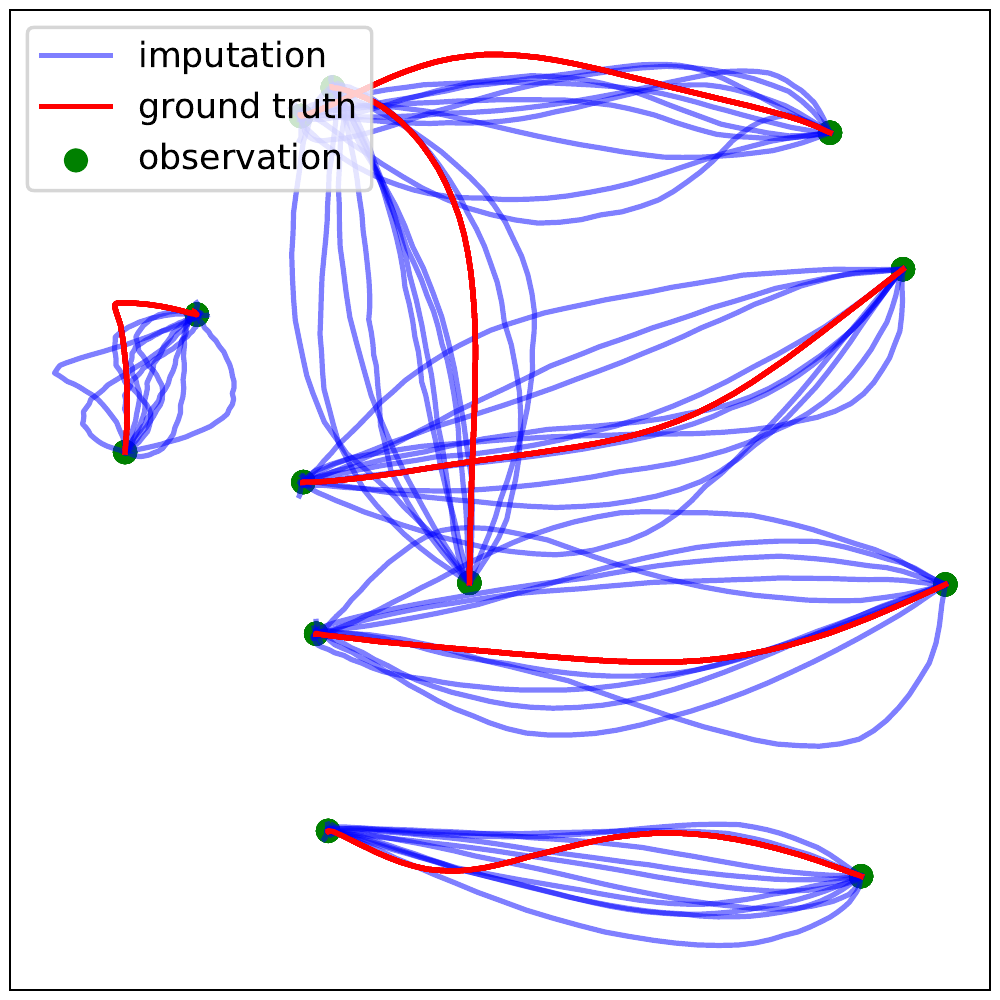}}
    \vspace{-10pt}
    \caption{Imputed trajectories of football players.} 
    \vspace{-10pt}
    \label{fig:nfl}
\end{minipage} &
\hfill
    \begin{minipage}{.44\textwidth}
    \vspace{0pt}
\begin{center}
\setlength{\tabcolsep}{2pt}
\begin{scriptsize}
\begin{tabular}{c|cccc}
\toprule
\textbf{Models} & \texttt{Latent-ODE} & \texttt{NAOMI}  & NRTSI & Expert\\
\midrule
\textbf{step change ($\times 10^{-3}$)} & 1.473 & 3.227 & \textbf{2.401} & 2.482 \\
\textbf{avg length} & 0.136 & 0.236 & \textbf{0.175} & 0.173 \\
\textit{min}\textbf{MSE} ($\times 10^{-3}$) & 19.53 & 4.079 & \textbf{1.908} & 0.000\\
\textit{avg}\textbf{MSE} / \textit{min}\textbf{MSE} & 1.16 & 1.12 & \textbf{2.13} & --- \\
\textbf{Imputation Time (s)} & \textbf{0.13} & 0.64 & 0.51 & ---\\
\bottomrule
\end{tabular}
\end{scriptsize}
\caption{Quantitative comparison on Football Player Trajectory. A larger \textit{avg}MSE / \textit{min}MSE indicate better diversity. Other statistics closer to the expert indicate better performance.}
\label{table:nfl}
\end{center}
\vspace{-0.1in}
    \end{minipage}

\end{tabular}
\end{figure}

\textbf{Evaluation Metrics}\ \ \  This dataset is stochastic and contains multiple modes since there could be many possible trajectories based on the sparsely observed data. Therefore, we follow \cite{park2020diverse} to use \textit{min}MSE to evaluate the precision and the ratio between \textit{avg}MSE and \textit{min}MSE to evaluate the diversity of the imputed trajectories. \textit{min}MSE and \textit{avg}MSE respectively denote the minimum and the average MSEs between the ground truth and $n$ independently sampled imputation trajectories. A small \textit{min}MSE indicates that at least one of $n$ samples has high precision, while a large \textit{avg}MSE implies that these $n$ samples are spread out. We further define \textit{avg}MSE / \textit{min}MSE to measure the diversity of sampled trajectories. Similar to \cite{liu2019naomi}, we also use \textit{average trajectory length} and \textit{step change} to assess the quality of sampled trajectories.

\textbf{Results}\ \ \  For this dataset, we train NRTSI to minimize the negative log-likelihood as in \eqref{eq:stochastic}. For each trajectory, we randomly select 40 to 49 timesteps as missing. According to the discussion in Sec \ref{sec:variant}, data with missing gaps larger than 4 are imputed one by one, while data with smaller missing gaps are imputed in parallel. As shown in Figure \ref{table:nfl}, NRTSI compares favorably to NAOMI, which is the state-of-the-art method on this dataset. NRTSI also outperforms Latent-ODE by a large margin. The speed does not degrade too much by the one-by-one imputation on the large missing gap time points. It only takes 0.51 seconds to impute a batch of 64 trajectories on GTX 1080Ti, which is faster than NAOMI. The \textit{min}MSE and \textit{avg}MSE scores are calculated using 10 independently imputed trajectories. We also plot several imputed trajectories in Figure \ref{fig:nfl}. The imputation is accurate around observed time points and captures the uncertainty for points far from the observations. When the observation is sparse, NRTSI can impute multiple realistic trajectories.

\begin{wraptable}{R}{0.40\textwidth}
\vspace{-20pt}
    \caption{$L_2$ loss on irregularly-sampled Billiards data ($\times 10^{-2}$).}
    \vspace{-10pt}
    \begin{center}
    \setlength{\tabcolsep}{2pt}
    \begin{scriptsize}
    \begin{tabular}{ccccc}
        \toprule
        \texttt{Latent-ODE} & \texttt{NeuralCDE} & \texttt{ANP} & \texttt{NAOMI-$\Delta_t$} & NRTSI  \\
        \midrule
          19.48 & 34.01 & 29.31 & 1.121 & \textbf{0.042} \\
        \bottomrule
    \end{tabular}
    \end{scriptsize}
    \end{center}
    \label{table:irr_billiard}
    \vspace{-13pt}
\end{wraptable}

\subsection{Irregularly-sampled Time Series} We first test NRTSI on a toy synthetic dataset. Following \cite{rubanova2019latent}, we create an irregularly-sampled dataset containing 1,000 sinusoidal functions with 100 time points for each function, from which 90 timesteps are randomly selected as missing data. As shown in Appendix  Figure \ref{fig:irr_sin}, both Latent-ODE \cite{rubanova2019latent} and NRTSI capture the periodic pattern of this dataset, but NRTSI is more accurate and obtains a much lower MSE ($7.17  \times 10^{-4}$) compared to Latent-ODE ($5.05 \times 10 ^{-2}$).

Next, we evaluate NRTSI on an irregularly-sampled Billiards dataset. This dataset is created with the same parameters (e.g. initial ball speed range, travel time) as in Sec \ref{sec:billiard}. The only difference is that this dataset is irregularly-sampled. We compare NRTSI with two representative ODE-based approaches that can deal with irregularly-sampled data, namely Latent-ODE \cite{rubanova2019latent} and NeuralCDE \cite{kidger2020neural}. We also modify NAOMI \cite{liu2019naomi} to handle irregularly-sampled data, which we call NAOMI-$\Delta_t$. The time gap information between observations is provided to the RNN update function of NAOMI-$\Delta_t$. 
We also compare to Attentive Neural Process (ANP) \cite{kim2018attentive} which can handle irregularly-sampled data as ANP simply encodes times as scalars.
In Table \ref{table:irr_billiard}, we report the MSE losses between imputed and ground truth values. We randomly select 180 to 195 timesteps out of the total 200 timesteps as missing for each trajectory. According to Table~\ref{table:irr_billiard}, NRTSI outperforms the baselines by a large margin despite extensive hyperparameter search for these baselines. We show several trajectories imputed by NRTSI in Appendix \ref{sec:nrtsi_add_vis}. To further investigate the poor performance of Latent-ODE, NeuralCDE, and ANP, we visualize their imputed trajectories with different numbers of observed data in Appendix \ref{sec:viz_latentode_neuralcde}. It can be seen that when the observation is dense (150 points observed), they all perform well. However, they have difficulty in predicting the correct trajectories when the observation becomes sparse (e.g. with only 5 points observed). The excellent performance of NRTSI and NAOMI-$\Delta_t$ indicates the benefits of the multiresolution imputation procedure. Furthermore, the superiority of NRTSI over NAOMI-$\Delta_t$ demonstrates the advantage of the proposed set modeling approach.

\begin{wraptable}{R}{0.55\textwidth}
\centering
\vspace{-25pt}
\caption{The MSE comparison under different missing rates on the air quality dataset and the gas sensor dataset.}
\label{table:pod}
\begin{tiny}
\setlength{\tabcolsep}{2.4pt}
\begin{tabular}{@{}c|c|cccccccc@{}}
\toprule
\multirow{2}{*}{Dataset}      & \multirow{2}{*}{Method} & \multicolumn{8}{c}{missing rate}                                                                                                              \\
                              &                           & 10\%            & 20\%            & 30\%            & 40\%            & 50\%            & 60\%            & 70\%            & 80\%            \\ \midrule
\multirow{5}{*}{Air} 
                          & \texttt{Latent-ODE}                     & .2820          & .2954          & .3161          & .3291          & .3462          & .3569          & .3614          & .3762          \\
                          
                          & \texttt{NeuralCDE}                     & .2951          & .3129          & .3337          & .3524          & .3898          & .4074          & .4192          & .4865          \\
                          
                              & \texttt{BRITS}                     & .1659          & .2076          & .2212          & .2088          & .2141          & .2660          & .2885          & .3421          \\

                              & \texttt{RDIS}                   & .1409 & .1807 & .2008 & .1977 & .2041 & .2528 & .2668 & .3178 \\ \cmidrule(l){2-10}  
& NRTSI               & \textbf{.1230} & \textbf{.1155} & \textbf{.1189} & \textbf{.1250} & \textbf{.1297} & \textbf{.1378} & \textbf{.1542} & \textbf{.1790} \\
\midrule
\multirow{5}{*}{Gas} 
                         & \texttt{Latent-ODE}                     & .1251          & .1282          & .1278          & .1299          & .1332          & .1387          & .1487          & .1979          \\
                         
                          & \texttt{NeuralCDE}                     & .0685          & .0773          & .0821          & .1044          & .1251          & .1538          & .1754          & .3011          \\
                          
                              & \texttt{BRITS}                     & .0210          & .0226          & .0233          & .0279          & .0338          & .0406          & .0518          & .1595          \\
                          
                              & \texttt{RDIS}                   & .0287 & .0226 &  .0241 & .0251 & \textbf{.0277} & .0321 & \textbf{.0350} & .0837 \\
                              \cmidrule(l){2-10}
                      
& NRTSI              & \textbf{.0165} & \textbf{.0195} &  \textbf{.0196} & \textbf{.0229} & .0286 & \textbf{.0311} & .0362 & \textbf{.0445} \\
                              \bottomrule
\end{tabular}
\vspace{-8pt}
\end{tiny}
\end{wraptable}


\subsection{Partially Observed Time Series}
The air quality dataset \cite{zhang2017cautionary} and the gas sensor dataset \cite{burgues2018estimation} are two popular datasets to evaluate the scenario where dimensions at each time point are partially observed. Data in these datasets are 11-dimensional and 19-dimensional respectively. For both datasets, we follow RDIS \cite{choi2020rdis} to select 48 consecutive timesteps to construct one regularly-sampled time series. We compare NRTSI to the state-of-the-art method RDIS \cite{choi2020rdis}, as well as BRITS \cite{cao2018brits}, Latent-ODE and NeuralCDE. In Table \ref{table:pod}, we report the MSE scores by randomly masking out some dimensions for all timesteps with several different missing rates. For the air quality dataset, NRTSI outperforms the baselines for all missing rates, and for the gas sensor dataset, NRTSI wins on most of the missing rates. 

\section{Discussion}
\label{sec:discussion}
Throughout the experiments, we conduct an extensive hyperparameter searching for the baselines to make sure they perform as well as they can be. As shown in Appendix \ref{sec:baseline_hyper}, we find that the best configurations of these baselines have fewer parameters than NRTSI does, and increasing their model capacity does not lead to better results. Thus, the superiority of NRTSI is due to the novel architecture rather than naively using more parameters or tuning hyperparameters only for NRTSI. In fact, NRTSI shares hyperparameters across all the datasets (see Appendix \ref{sec:NRTSI_arch}). 
In terms of the imputation speed, though NRTSI contains more parameters than NAOMI, the increase of the number of parameters does not slow down the imputation speed of NRTSI. According to Figure \ref{table:nfl}, NRTSI is faster than NAOMI as NRTSI can impute multiple missing data in parallel. This speedup is also observed on other datasets. To evaluate the contribution of each component in NRTSI, we conduct several ablation studies in Appendix \ref{sec:nrtsi_ablation} and find that both the hierarchical imputation procedure and the proposed attention mechanism are required to achieve good performance. 
Despite the $O(L^2)$ complexity of the attention mechanism where $L$ denotes the sequence length, a simple extension may enable NRTSI to handle extremely long sequences: use global attention only for a small number of data with large missing gaps, while for data with small missing gaps only attend to local data with a small horizon $l \ll L$ as imputation at fine-grained scale only depends on local data.

\section{Conclusion} 
In this work, we introduce a novel time-series imputation approach named NRTSI. NRTSI represents time series as permutation equivariant sets and leverages a Transformer-based architecture to impute the missing values. We also propose a hierarchical imputation procedure where missing data are imputed in the order of their missing gaps (i.e. the distance to the closest observations). NRTSI is broadly applicable to numerous applications, such as irregularly-sampled time series, partially observed time series, and stochastic time series. Extensive experiments demonstrate that NRTSI achieves state-of-the-art performance on commonly used time series imputation benchmarks.


\bibliography{main}
\bibliographystyle{plainnat.bst}

\section*{Checklist}

\begin{enumerate}

\item For all authors...
\begin{enumerate}
  \item Do the main claims made in the abstract and introduction accurately reflect the paper's contributions and scope?
    \answerYes{}
  \item Did you describe the limitations of your work?
    \answerYes{Please see Section \ref{sec:discussion}.}
  \item Did you discuss any potential negative societal impacts of your work?
    \answerYes{Please see Appendix \ref{sec:impact}.}
  \item Have you read the ethics review guidelines and ensured that your paper conforms to them?
    \answerYes{}
\end{enumerate}

\item If you are including theoretical results...
\begin{enumerate}
  \item Did you state the full set of assumptions of all theoretical results?
    \answerNA{}
	\item Did you include complete proofs of all theoretical results?
    \answerNA{}
\end{enumerate}

\item If you ran experiments...
\begin{enumerate}
  \item Did you include the code, data, and instructions needed to reproduce the main experimental results (either in the supplemental material or as a URL)?
    \answerYes{All the codes to reproduce the expeirments are in the supplemental materials. All the datasets can be downloaded via the links in the supplemental materials.}
  \item Did you specify all the training details (e.g., data splits, hyperparameters, how they were chosen)?
    \answerYes{Please see Appendix \ref{sec:dataset} for data splits, Appendix \ref{sec:traning_detail} for training details, Appendix \ref{sec:NRTSI_arch}} for the hyperparameters of the proposed method NRTSI, Appendix \ref{sec:baseline_hyper} for the hyperparameters of the baselines and how they are chosen.
	\item Did you report error bars (e.g., with respect to the random seed after running experiments multiple times)?
    \answerYes{All the experimental results in our paper are calculated by repeating the test set 100 times to make sure a variety of missing patterns are covered.}
	\item Did you include the total amount of compute and the type of resources used (e.g., type of GPUs, internal cluster, or cloud provider)?
    \answerYes{Please see Appendix \ref{sec:traning_detail}.}
\end{enumerate}

\item If you are using existing assets (e.g., code, data, models) or curating/releasing new assets...
\begin{enumerate}
  \item If your work uses existing assets, did you cite the creators?
    \answerYes{See Appendix \ref{sec:dataset}.}
  \item Did you mention the license of the assets?
    \answerYes{See Appendix \ref{sec:dataset}.}
  \item Did you include any new assets either in the supplemental material or as a URL?
    \answerYes{We provide the link to download the Football Player Trajectory dataset in the supplemental material.}
  \item Did you discuss whether and how consent was obtained from people whose data you're using/curating?
    \answerYes{See Appendix \ref{sec:dataset}.}
  \item Did you discuss whether the data you are using/curating contains personally identifiable information or offensive content?
    \answerYes{See Appendix \ref{sec:dataset}.}
\end{enumerate}

\item If you used crowdsourcing or conducted research with human subjects...
\begin{enumerate}
  \item Did you include the full text of instructions given to participants and screenshots, if applicable?
    \answerNA{}
  \item Did you describe any potential participant risks, with links to Institutional Review Board (IRB) approvals, if applicable?
    \answerNA{}
  \item Did you include the estimated hourly wage paid to participants and the total amount spent on participant compensation?
    \answerNA{}
\end{enumerate}

\end{enumerate}

\newpage
\appendix

\section{Detailed Imputation and Training Procedures}
\label{append_sec_train}

\begin{algorithm}[H]
        \caption{\textsc{Imputation Procedure}}
        \begin{algorithmic}[1]
        \label{algo}
          \REQUIRE $f_\theta^l$: imputation model at resolution level $l$, $L$: the maximum resolution level, $\mathbf S$: observed data, $\hat{\mathbf S}$: data to impute
\STATE Initialize current imputation set $\G$ empty, i.e. $\G \leftarrow \emptyset$, current resolution level $l$ to 0, i.e. $l \leftarrow 0$
\WHILE{$l \leq L $}
\STATE Find times points to impute at current level $l$, i.e. $\hat{\mathbf S}^l \leftarrow \{\ (\hat{t}_j, \Delta \hat{t}_j)\ |\ (\hat{t}_j,\Delta \hat{t}_j) \in \hat{\mathbf S}$ and  $\floor{2^{L-l-1}}<\Delta \hat{t}_j\leq 2^{L-l}\}$
\WHILE{$\hat{\mathbf S}^l$ is not empty}
\STATE Find the data to impute with the largest missing gap and put them into $\G$, i.e. $\G \leftarrow \{\ (\hat{t}_j,\Delta \hat{t}_j)\ |\ (\hat{t}_j,\Delta \hat{t}_j) \in \hat{\mathbf S}^l, \Delta \hat{t}_j=\max_{j}\Delta \hat{t}_j \}$
\STATE Compute the imputation results $\mathbf H=\{(\hat{t}_j, \h_j)\}_{j=1}^{|\mathbf{G}|}$ based on the observed data $\mathbf S$, i.e. $\mathbf H \leftarrow f_{\theta}^{l}(\G;\ \mathbf S)$
\STATE Regard imputed data as observed data and delete them from $\hat{\mathbf S}$ and $\hat{\mathbf S}^l$, i.e. $\mathbf S \leftarrow \mathbf S \cup \mathbf H$, $\hat{\mathbf S} \leftarrow \hat{\mathbf S} \setminus \G$, $\hat{\mathbf S}^l \leftarrow \hat{\mathbf S}^l \setminus \G$
\STATE Update the missing gap information in $\hat{\mathbf S}$ and $\hat{\mathbf S}^l$ 
\ENDWHILE
\STATE $l \leftarrow l + 1 $
\ENDWHILE
\STATE \textbf{return} Imputation result $\mathbf S$
\end{algorithmic}
\end{algorithm}

The only difference between the training procedure shown below and the imputation procedures shown in the main text is that during training, we regard the ground truth imputation target $
\Y$ as observed data for the subsequent training steps rather than the imputation result $\mathbf{H}$. This idea follows the philosophy of the well-known Teacher Forcing algorithm \cite{williams1989learning} commonly used to stabilize and speed up the training of RNNs.
\begin{algorithm}[H]
    \caption{\textsc{Training Procedure at level }$l$}
        \begin{algorithmic}[1]
        \label{algo:train}
          \REQUIRE $f_\theta^l$: imputation model at resolution level $l$, $\mathbf S$: observed data, $\hat{\mathbf S}$: data to impute, $\Y$: ground-truth imputation target
\STATE Initialize current imputation set $\G$ as an empty set, i.e. $\G \leftarrow \emptyset$
\IF{$l$ < L}
\STATE Initialize $\theta$ from the higher level trained model $f^{l+1}_{\theta}$ 
\ELSE
\STATE Randomly initialize $\theta$
\ENDIF
\WHILE{$f^l_{\theta}$ does not converge}
\STATE Sample a batch of training data $\hat{\mathbf{S}}\sim p(\hat{\mathbf{S}}), \mathbf{S}\sim p(\mathbf{S}),\Y\sim p(\Y)$
\STATE Find times points to impute at current level $\hat{\mathbf S}^l \leftarrow \{\ (\hat{t}_j, \Delta \hat{t}_j)\ |\ (\hat{t}_j,\Delta \hat{t}_j) \in \hat{\mathbf S}$ and  $\floor{2^{L-l-1}}<\Delta \hat{t}_j\leq 2^{L-l}\}$
\WHILE{$\hat{\mathbf S}^l$ is not empty}
\STATE Find the data to impute with the largest missing gap and put them into $\G$, i.e. $\G \leftarrow \{\ (\hat{t}_j,\Delta \hat{t}_j)\ |\ (\hat{t}_j,\Delta \hat{t}_j) \in \hat{\mathbf S}^l, \Delta \hat{t}_j=\max_{j}\Delta \hat{t}_j \}$
\STATE Compute the imputation results $\mathbf H=\{(\hat{t}_j, \h_j)\}_{j=1}^{|\mathbf{G}|}$ based on observed data $\mathbf S$, i.e. $\mathbf H \leftarrow f_{\theta}^{l}(\G;\ \mathbf S)$
\STATE Train $\theta$ via the objective function in \eqref{loss} in the main text 
\STATE Regard the ground-truth imputation target $\Y$ as observed points and delete $\mathbf{G}$ from  $\hat{\mathbf S}^l$, i.e. $\mathbf{S} \leftarrow \mathbf{S} \cup \Y$, $\hat{\mathbf S}^l \leftarrow \hat{\mathbf S}^l \setminus \mathbf{G}$
\ENDWHILE
\ENDWHILE
        \end{algorithmic}
    \end{algorithm}

\section{Model Architectures of NRTSI}
\label{sec:NRTSI_arch}
Our imputation model contains two linear layers $f_{\text{in}}$ and $f_{\text{out}}$ with bias terms as well as a Transformer encoder $f_\text{enc}$ that contains $N$ repeated self-attention blocks. We set $N$ to 8 for all the datasets reported in our paper except the irregularly-sampled sinusoidal dataset where we set $N$ to 2. Each self-attention block consists of a multi-head self-attention layer followed by a feedforward layer. We use 12 attention heads and every head has a key/query/value pair with a dimension of 128. Then, the output vectors of all heads are concatenated to form a 1,536-dimensional vector and projected to a 1,024-dimensional vector using a linear layer. The feedforward layer contains two linear layers where the first one projects the 1,024-dimensional vector to a 2,048-dimensional vector and the second one projects the 2,048-dimensional vector back to a 1,024-dimensional vector. 

Our model with 8 self-attention blocks contains 84.0M learnable parameters. Though this number is larger than the number of parameters of the baselines, we show in Appendix \ref{sec:baseline_hyper} that increasing the number of parameters of these baselines cannot improve their performance. 

For all the datasets, we set $\tau$ and $\nu$ in \eqref{eq:time} in the main text to 8 and 100 respectively. We set $\Delta$ in Section \ref{sec:variant} to 1.

    \begin{figure*}[]
\center
\subfigure[Block 1 Head 3]{\includegraphics[width=0.25\linewidth]{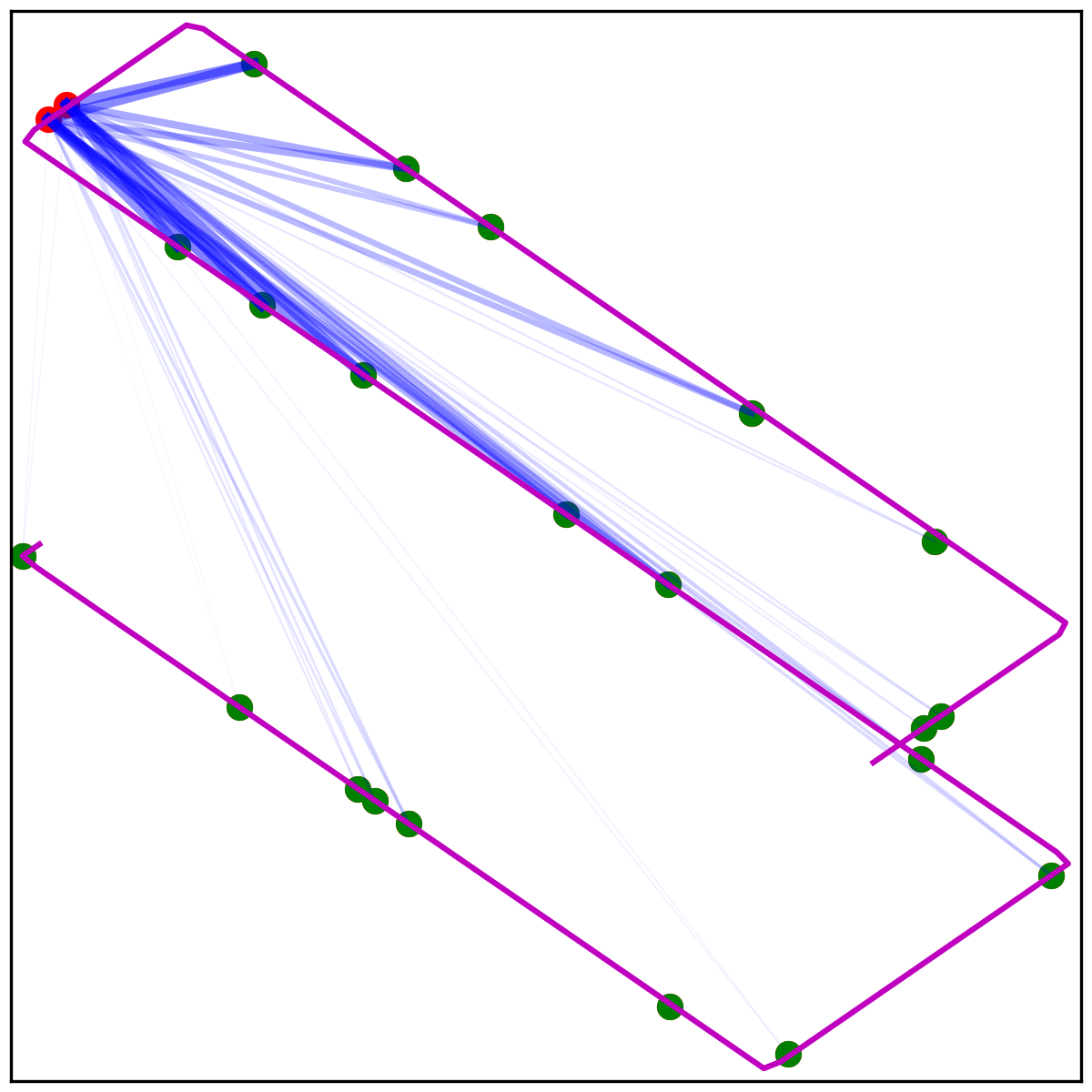}}
\subfigure[Block 1 Head 8]{\includegraphics[width=0.25\linewidth]{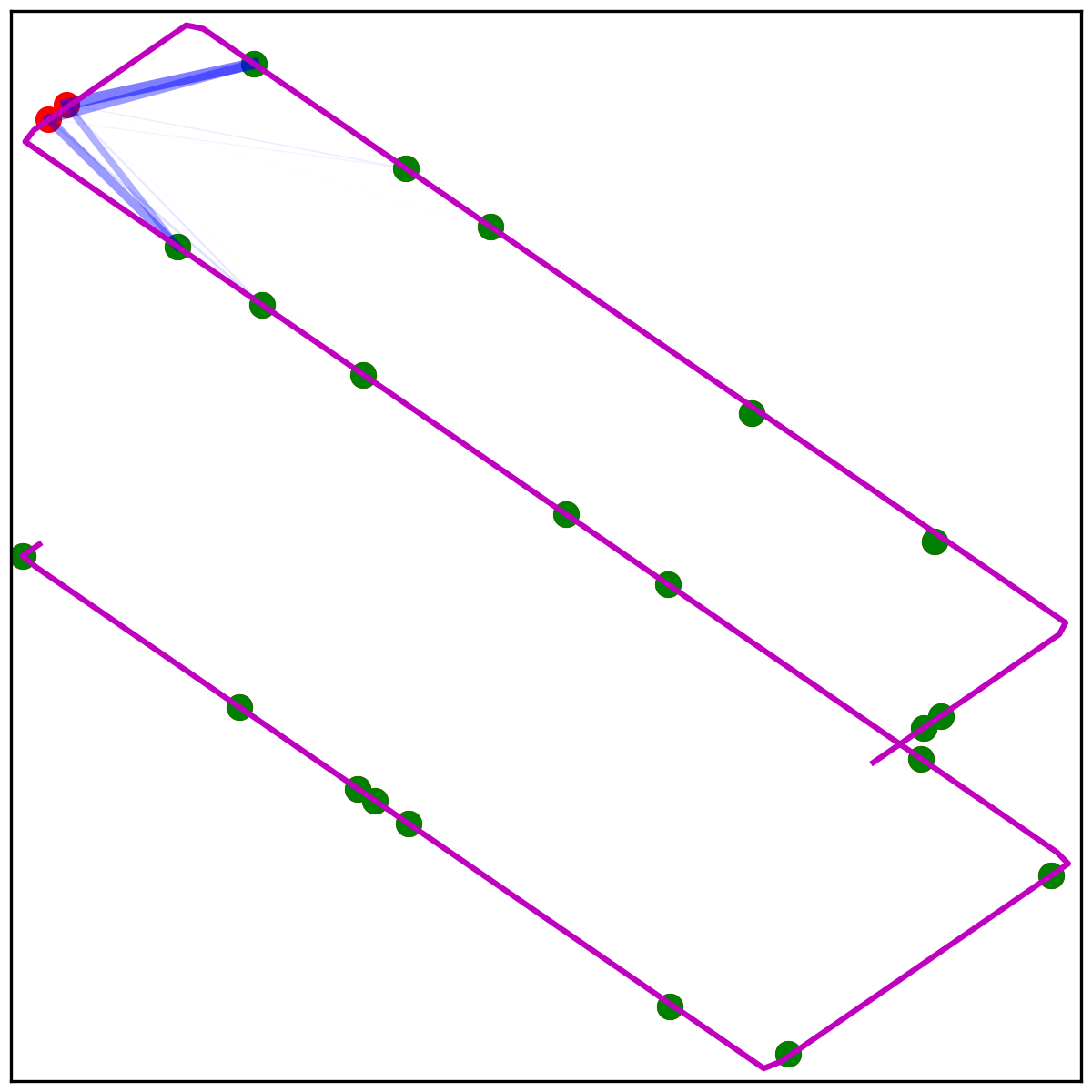}}
\subfigure[Block 1 Head 9]{\includegraphics[width=0.25\linewidth]{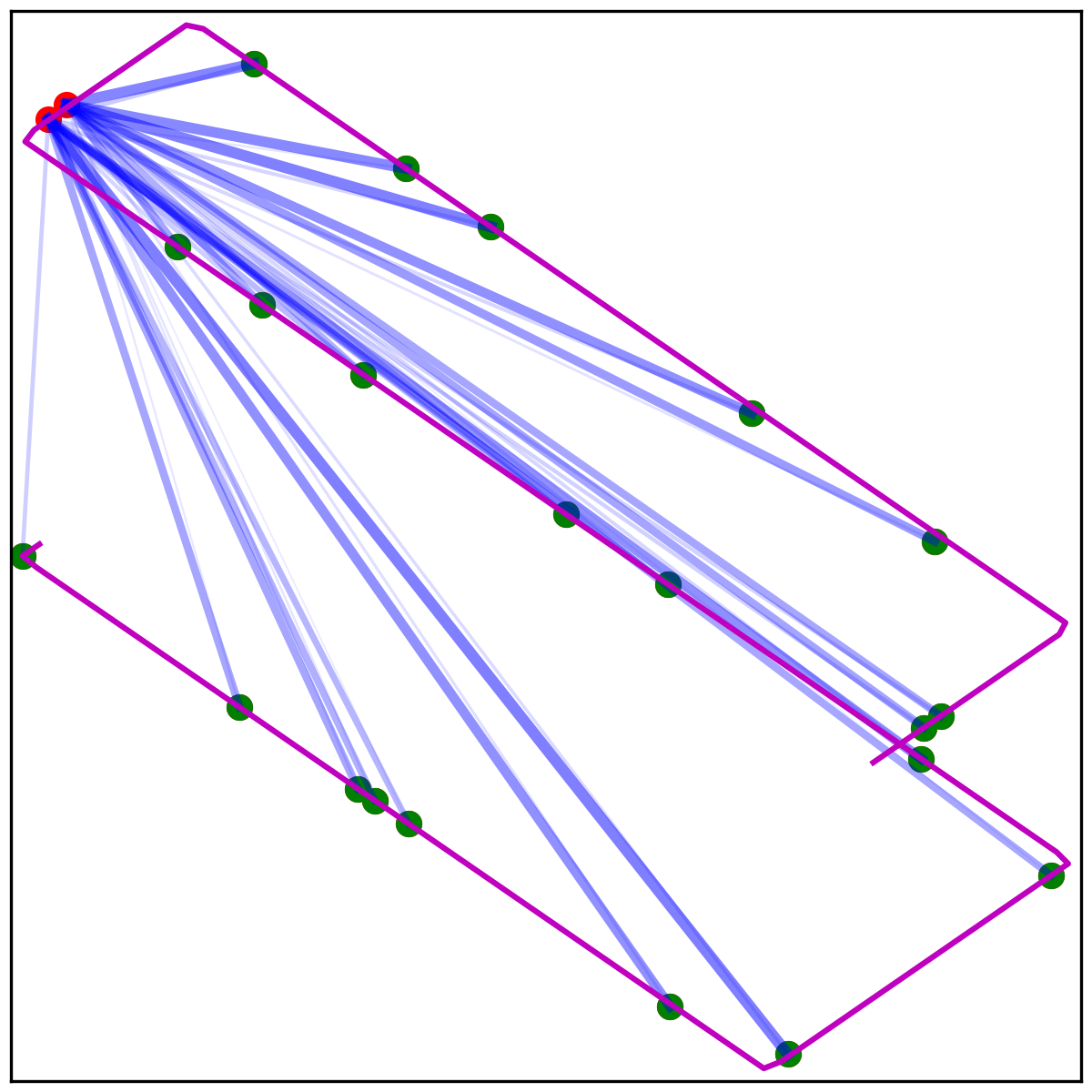}}

\subfigure[Block 5 Head 1]{\includegraphics[width=0.25\linewidth]{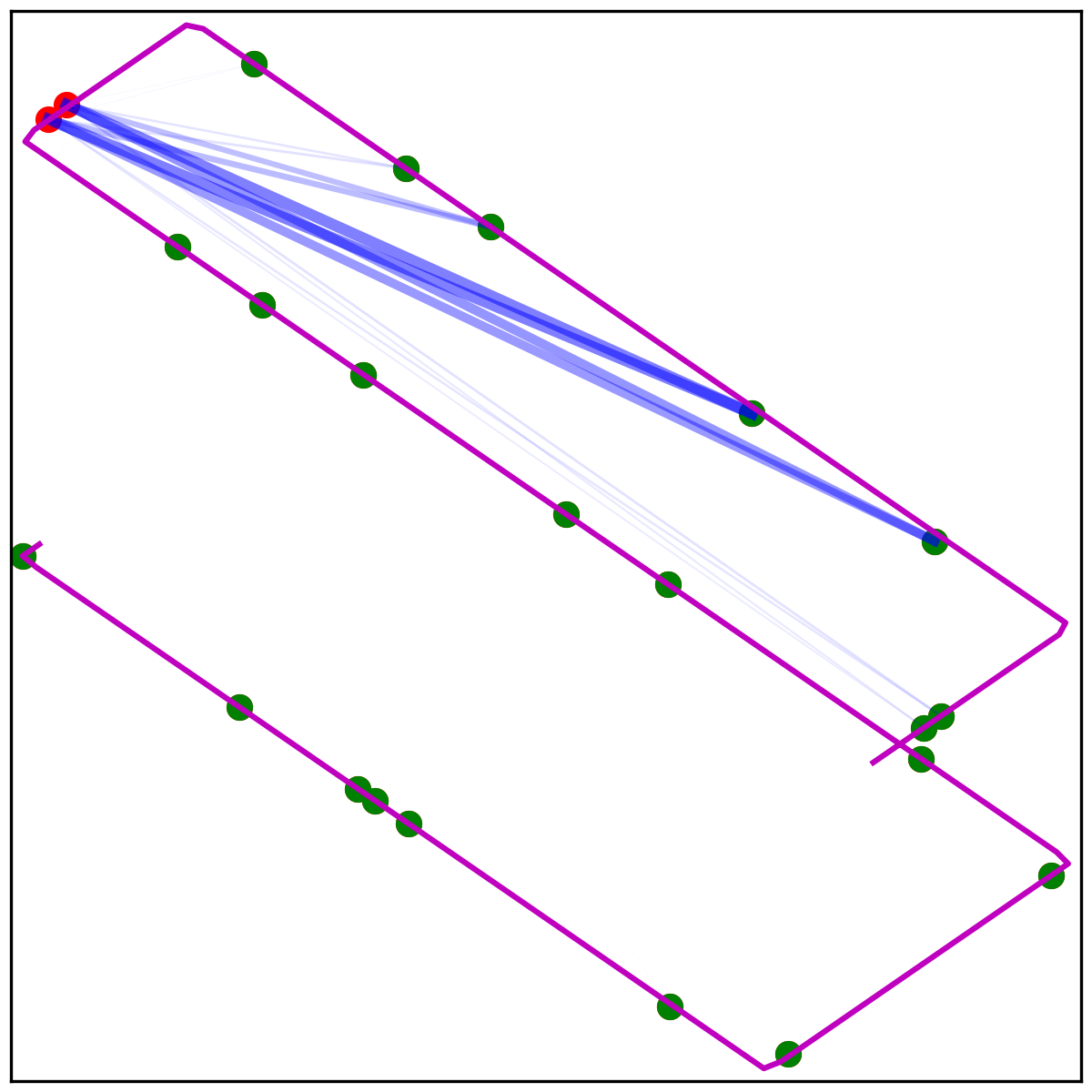}}
\subfigure[Block 5 Head 5]{\includegraphics[width=0.25\linewidth]{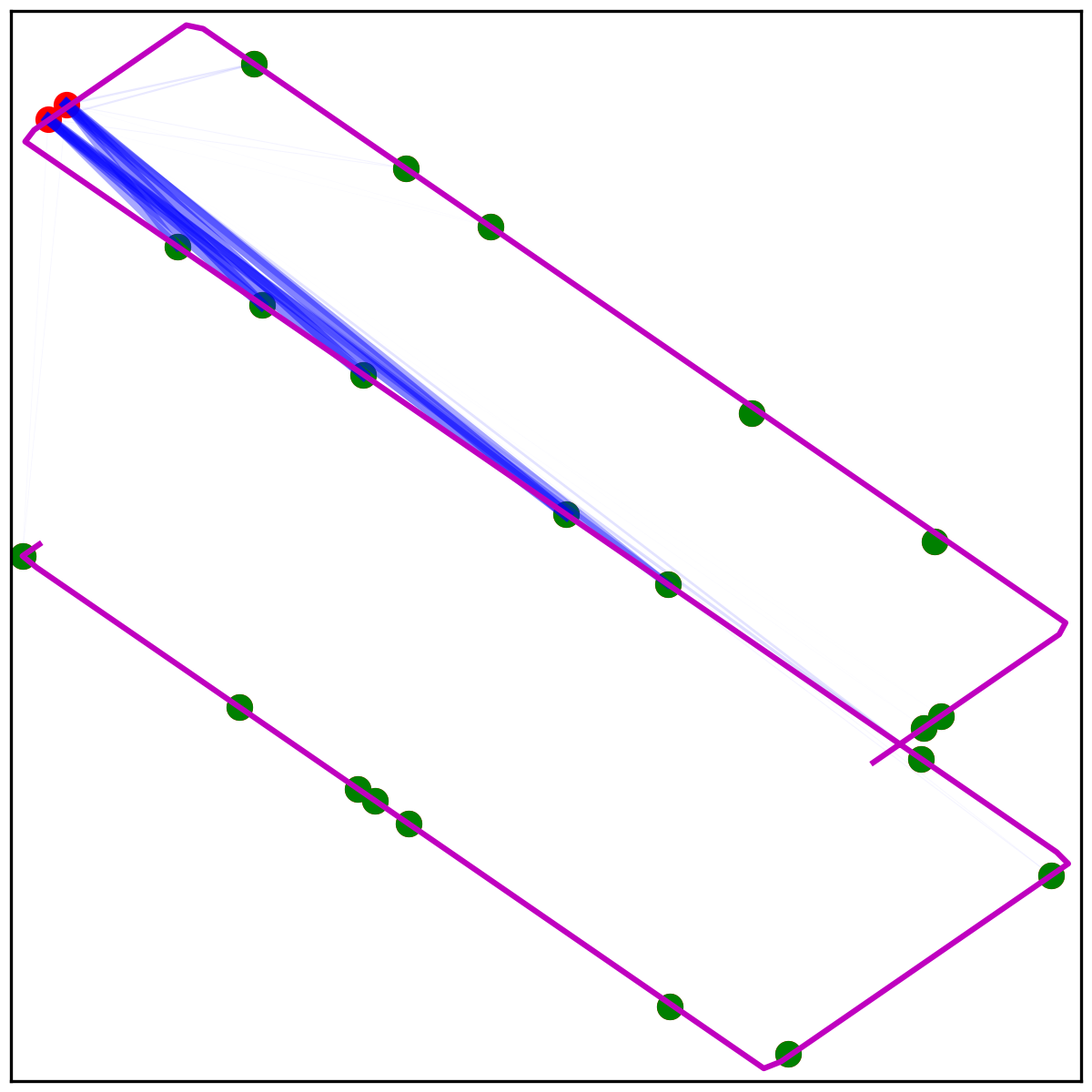}}
\subfigure[Block 5 Head 10]{\includegraphics[width=0.25\linewidth]{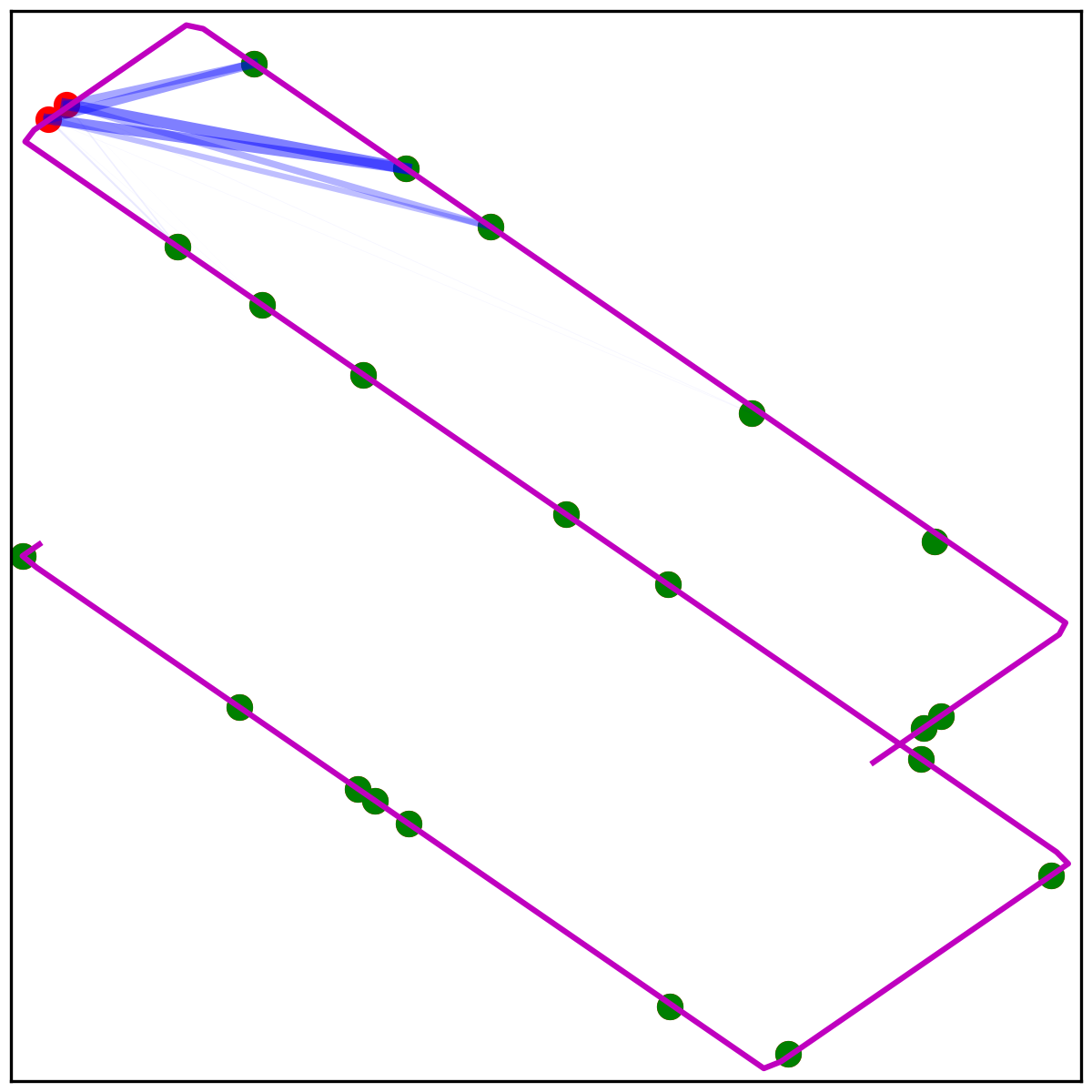}}

\subfigure[Block 7 Head 5]{\includegraphics[width=0.25\linewidth]{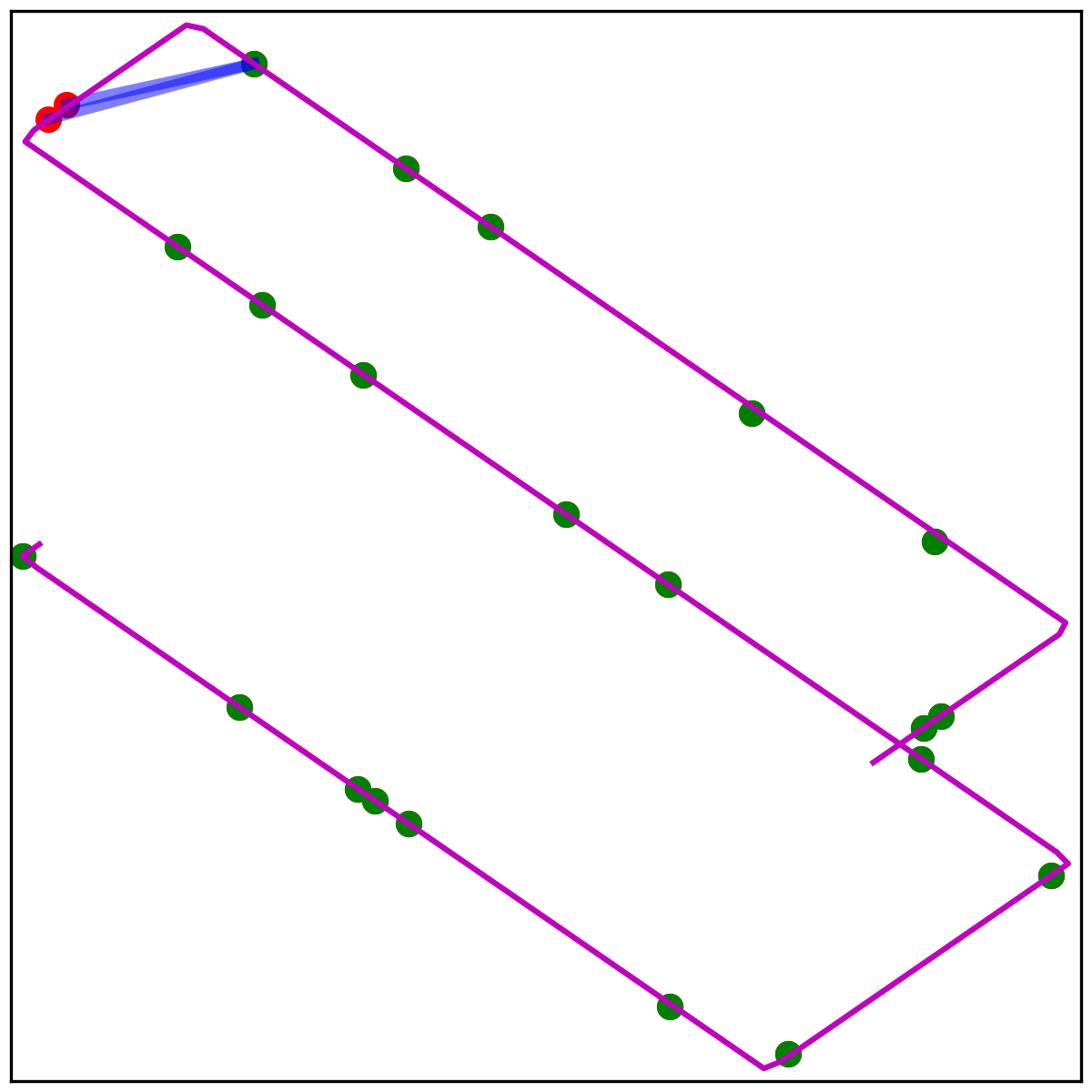}}
\subfigure[Block 7 Head 7]{\includegraphics[width=0.25\linewidth]{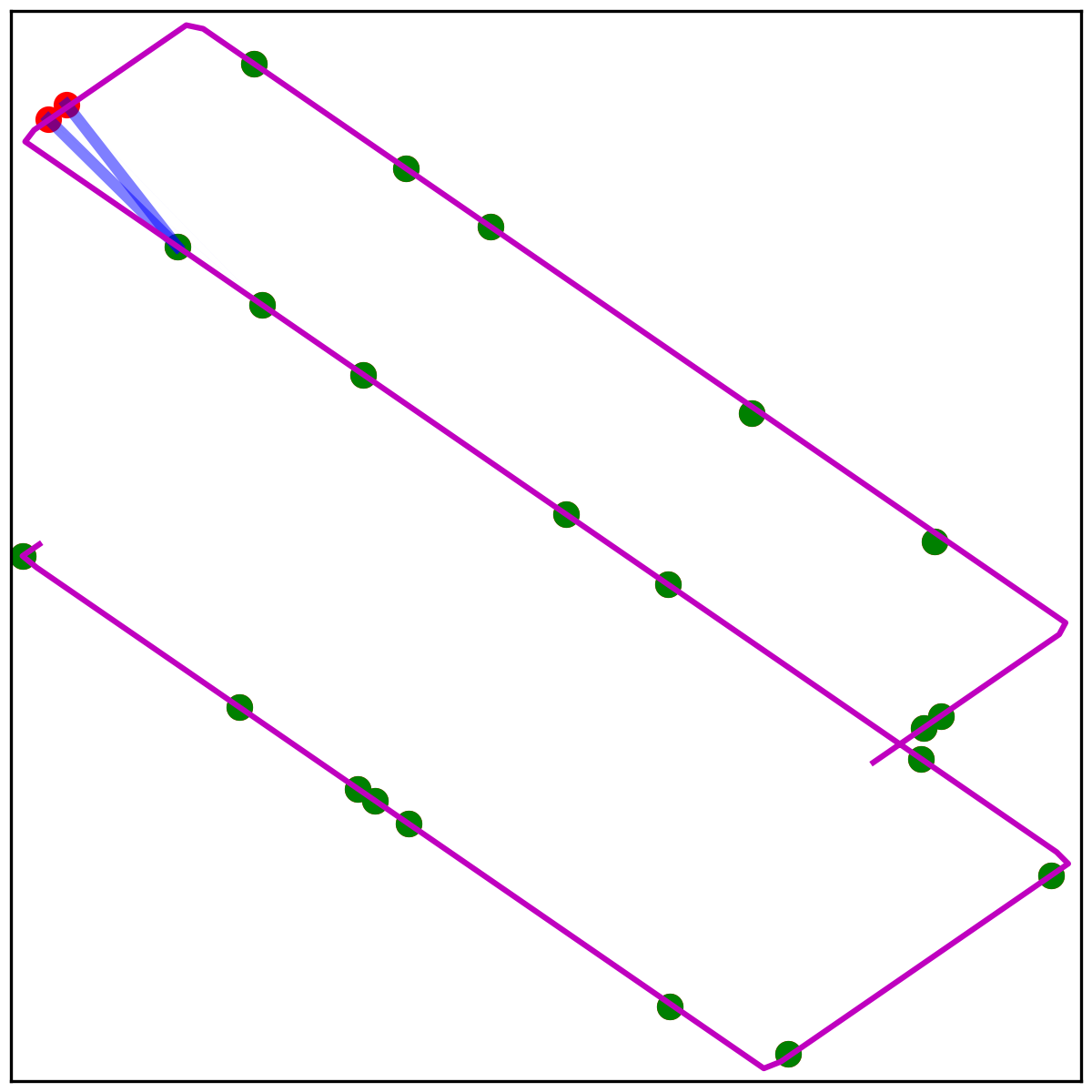}}
\subfigure[Block 7 Head 12]{\includegraphics[width=0.25\linewidth]{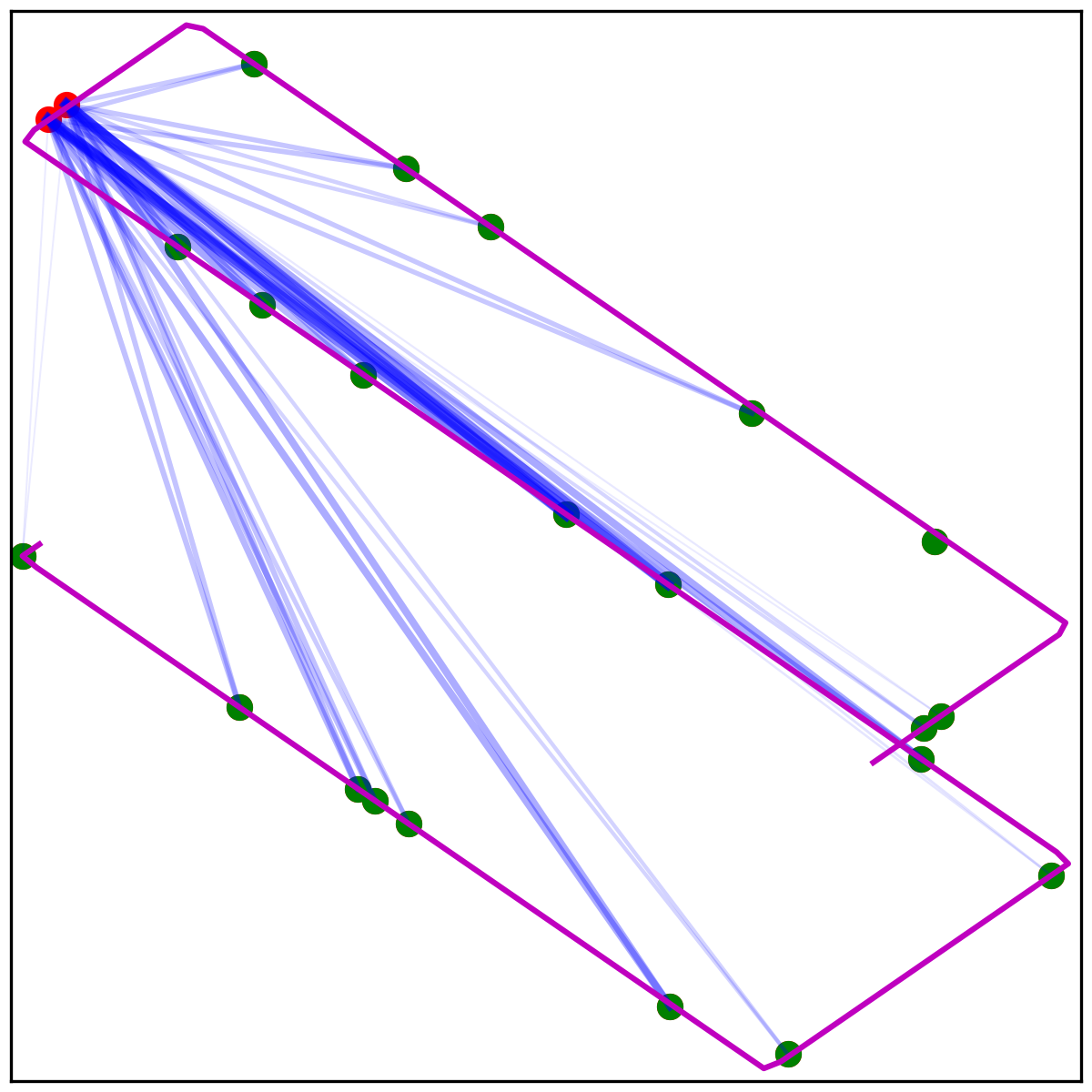}}

  \caption{Visualization of the attention maps at different self-attention blocks and different heads on the regularly-sampled Billiard dataset. There are 8 self-attention blocks with 12 heads each in our model. Red points are the imputed data, green points are the observed data and the purple solid lines are ground-truth trajectories. The softmax attention weights are visualized via the blue lines. The wider and less transparent the blue lines are, the larger the attention weights.}
  \label{fig:billiard_att}
\end{figure*}

\begin{figure*}[h]
\center
\subfigure{\includegraphics[width=0.16\linewidth]{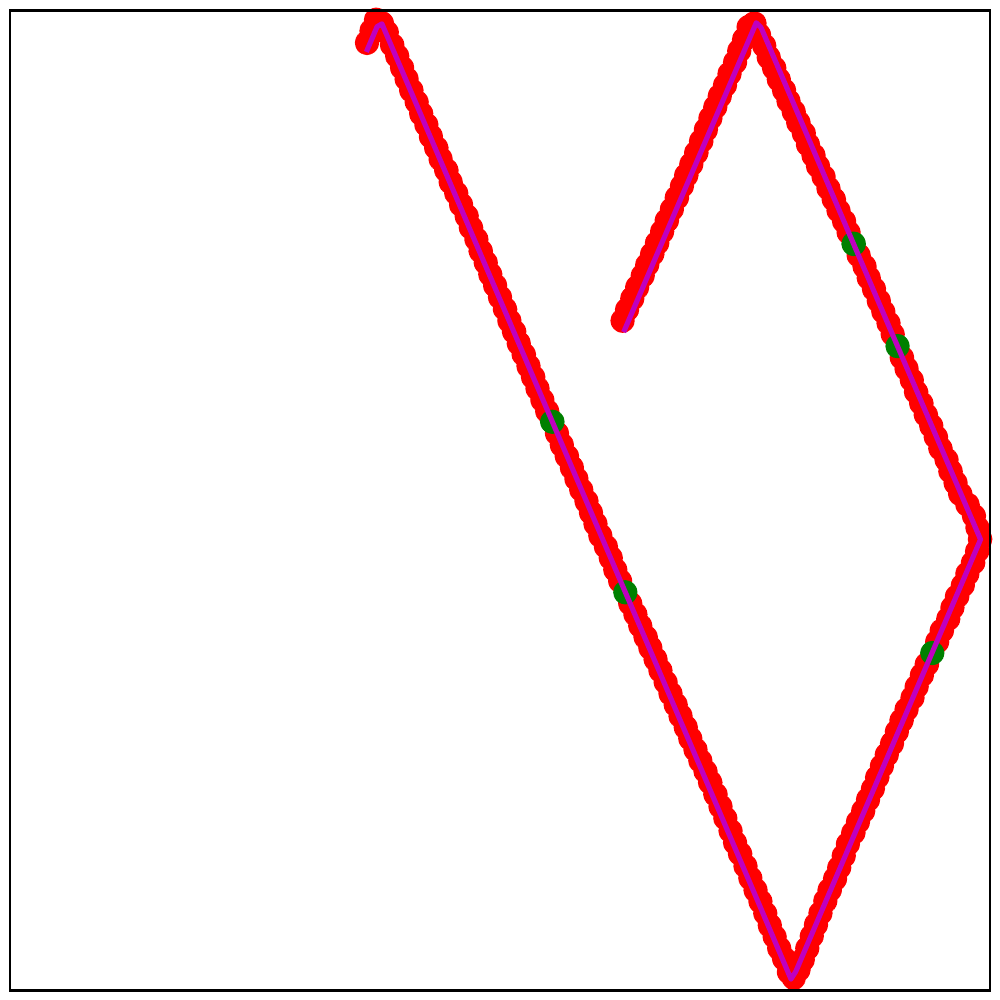}}
\subfigure{\includegraphics[width=0.16\linewidth]{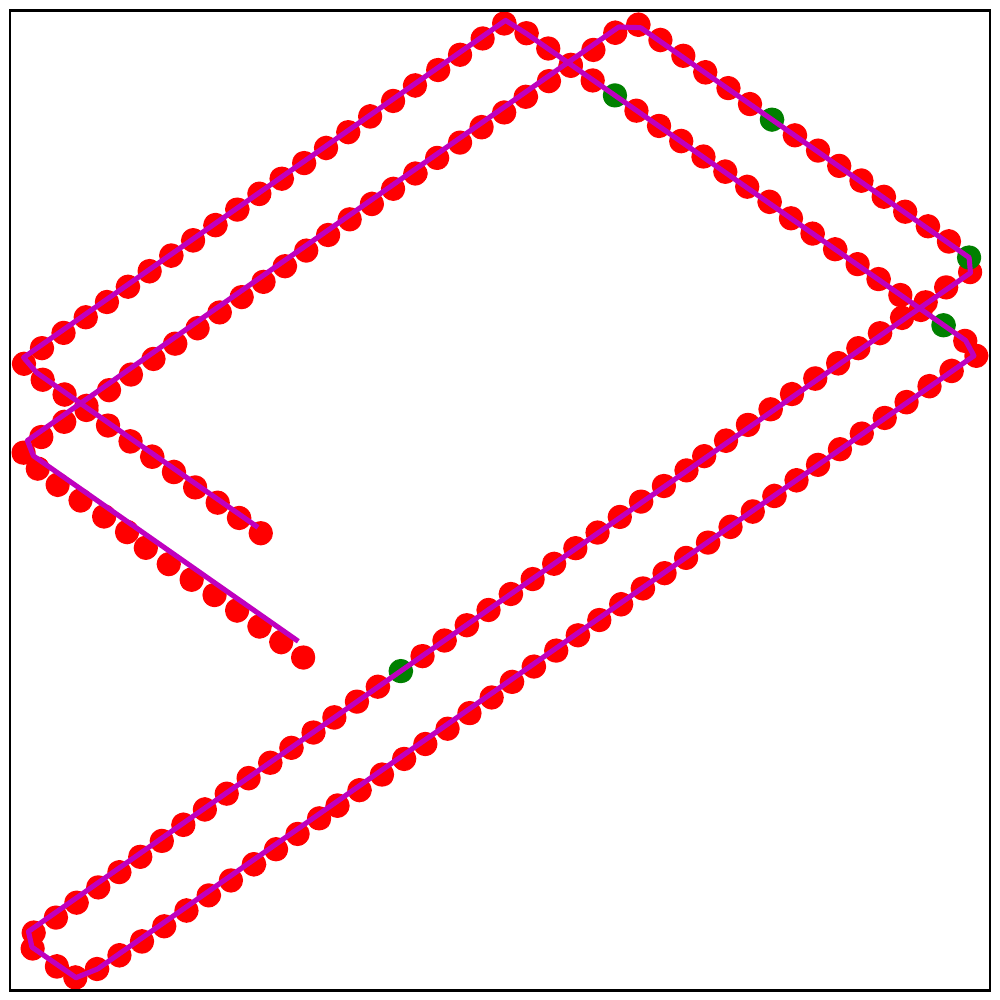}}
\subfigure{\includegraphics[width=0.16\linewidth]{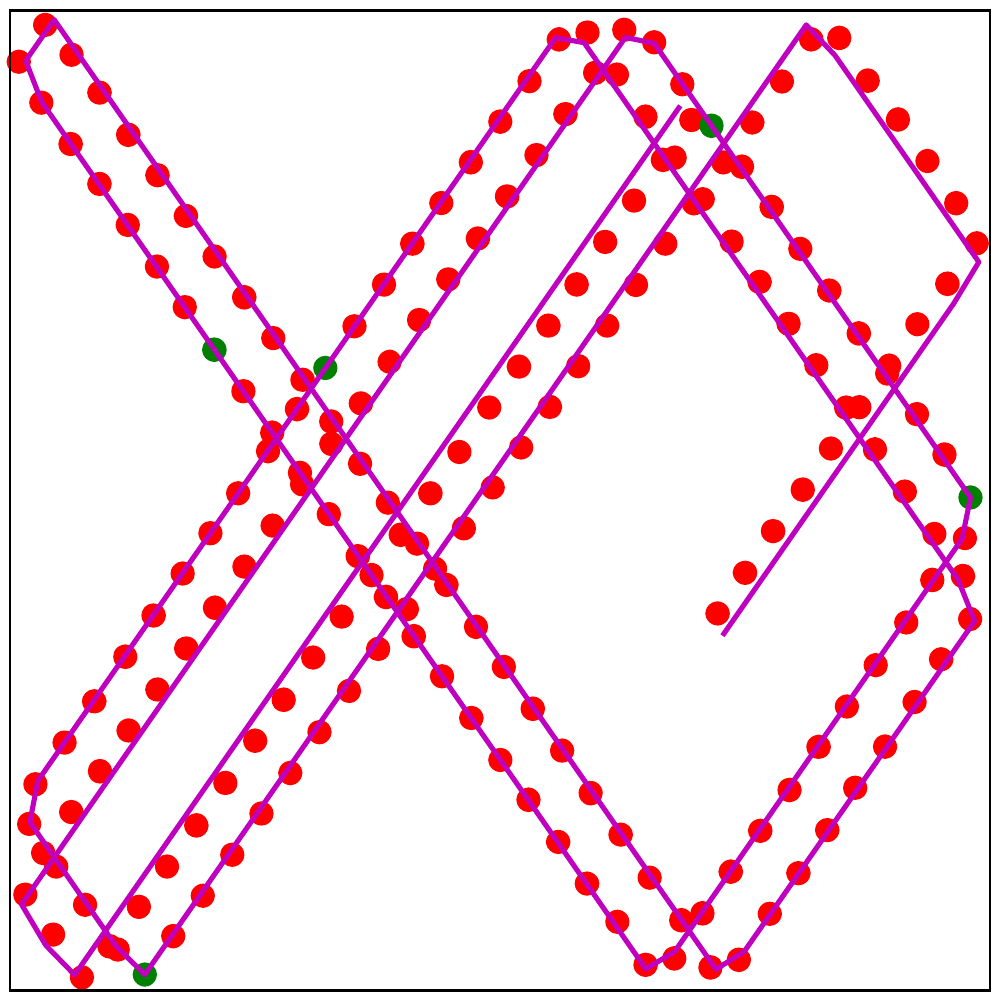}}
\subfigure{\includegraphics[width=0.16\linewidth]{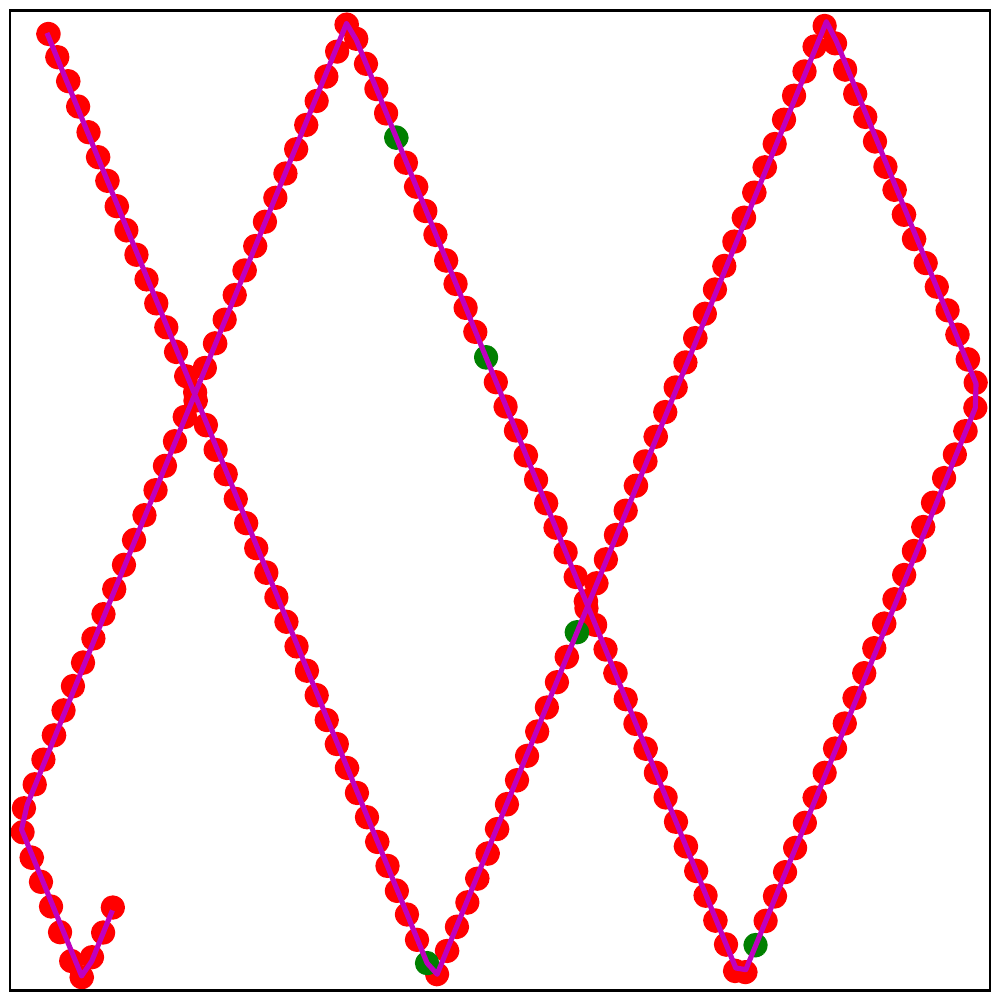}}
\subfigure{\includegraphics[width=0.16\linewidth]{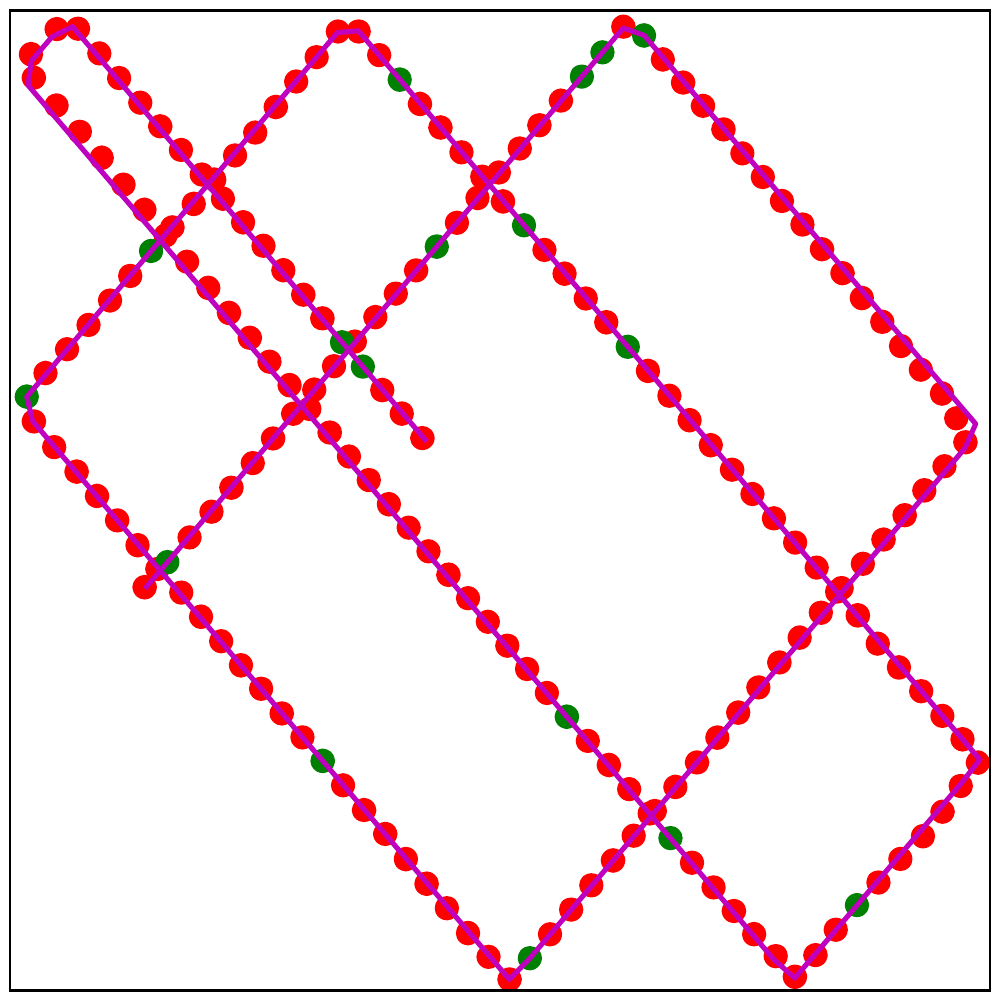}}
\subfigure{\includegraphics[width=0.16\linewidth]{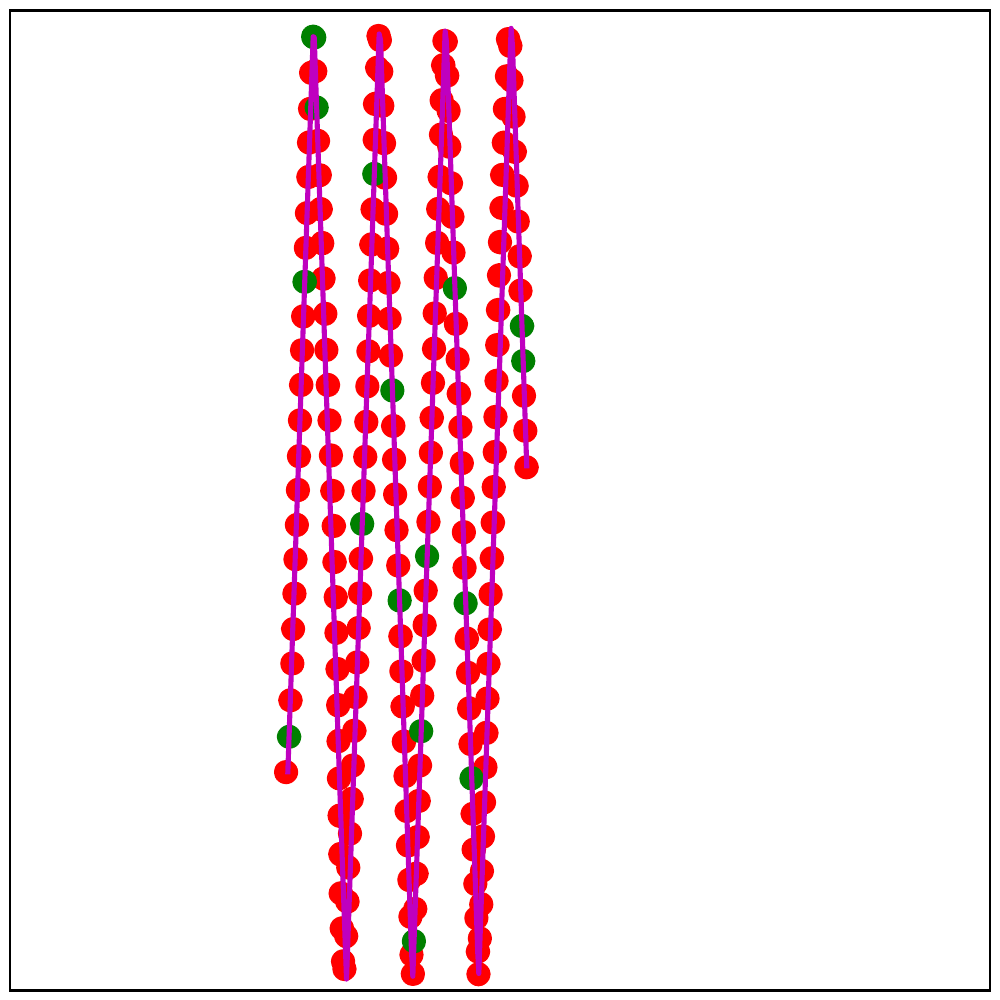}}
  \caption{Trajectories imputed by NRTSI on the regularly-sampled Billiards dataset. Red points are the imputed data, green points are the observed data and the purple solid lines are ground-truth trajectories.}
  \label{fig:billiard}
\end{figure*}

\begin{figure*}[h]
\center
\subfigure{\includegraphics[width=0.25\linewidth, trim={2.03cm 1.3cm 1cm 1.08cm},clip]{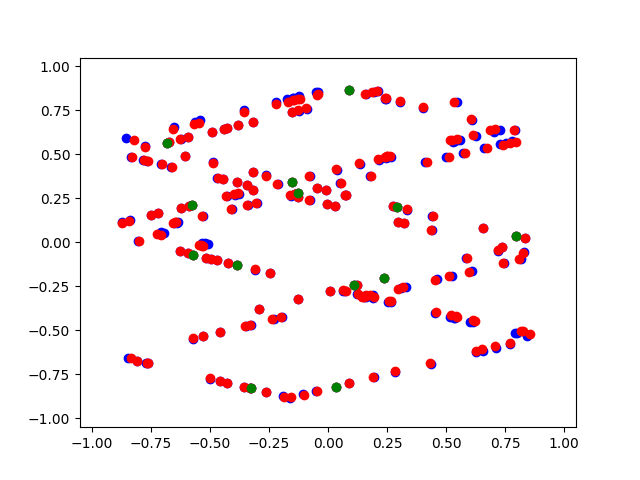}}
\subfigure{\includegraphics[width=0.25\linewidth, trim={2.03cm 1.3cm 1cm 1.08cm},clip]{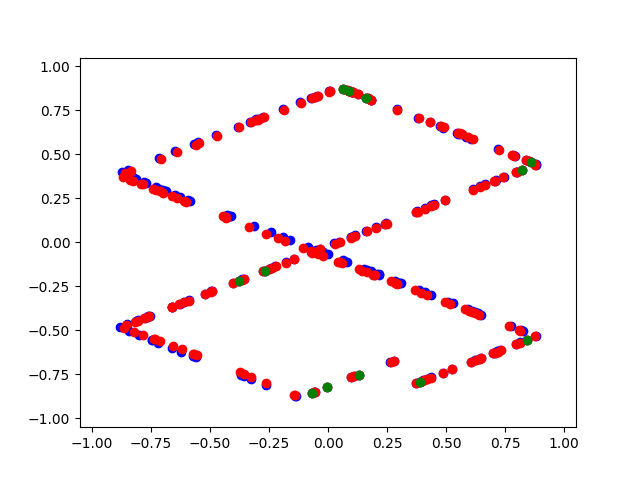}}
\subfigure{\includegraphics[width=0.25\linewidth, trim={2.03cm 1.3cm 1cm 1.08cm},clip]{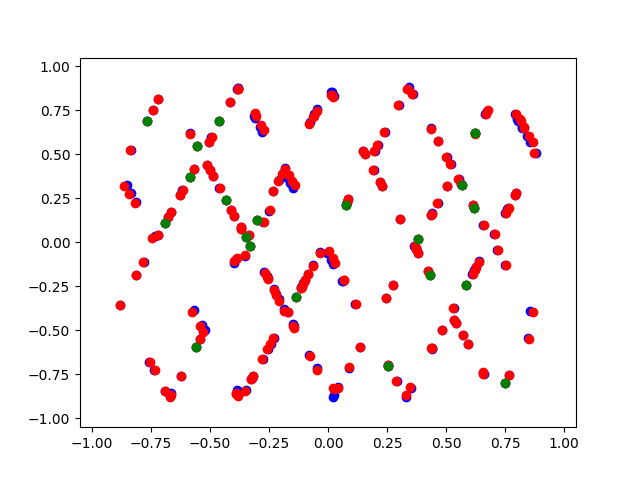}}
\subfigure{\includegraphics[width=0.25\linewidth, trim={2.03cm 1.3cm 1cm 1.08cm},clip]{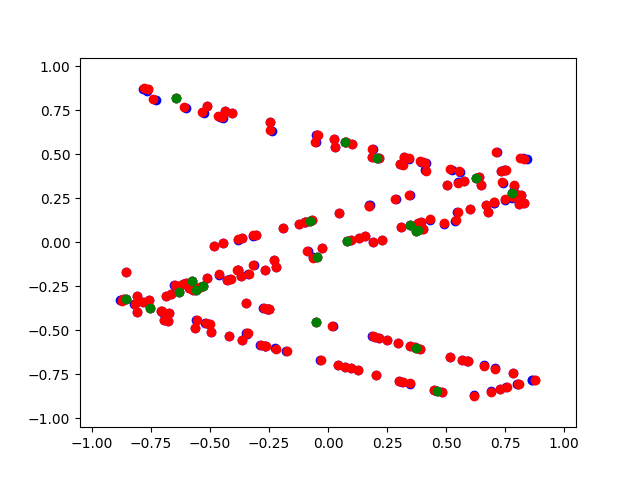}}
\subfigure{\includegraphics[width=0.25\linewidth, trim={2.03cm 1.3cm 1cm 1.08cm},clip]{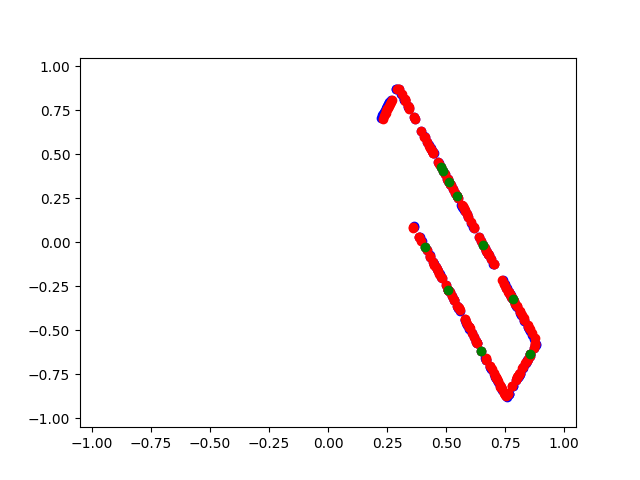}}
\subfigure{\includegraphics[width=0.25\linewidth, trim={2.03cm 1.3cm 1cm 1.08cm},clip]{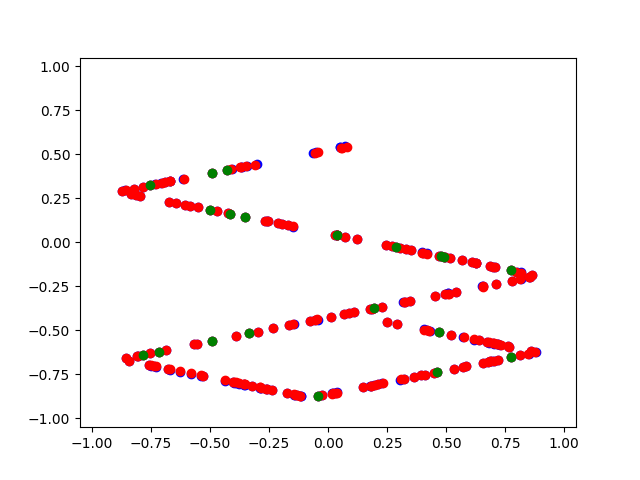}}
  \caption{Trajectories imputed by NRTSI on the irregularly-sampled Billiards dataset. The red points are the imputed data, the green points are the observed data and the blue points are ground-truth data.}
  \label{fig:irrbilliard}
\end{figure*}

\begin{figure*}[h]
\center
\subfigure{\includegraphics[width=0.30\linewidth]{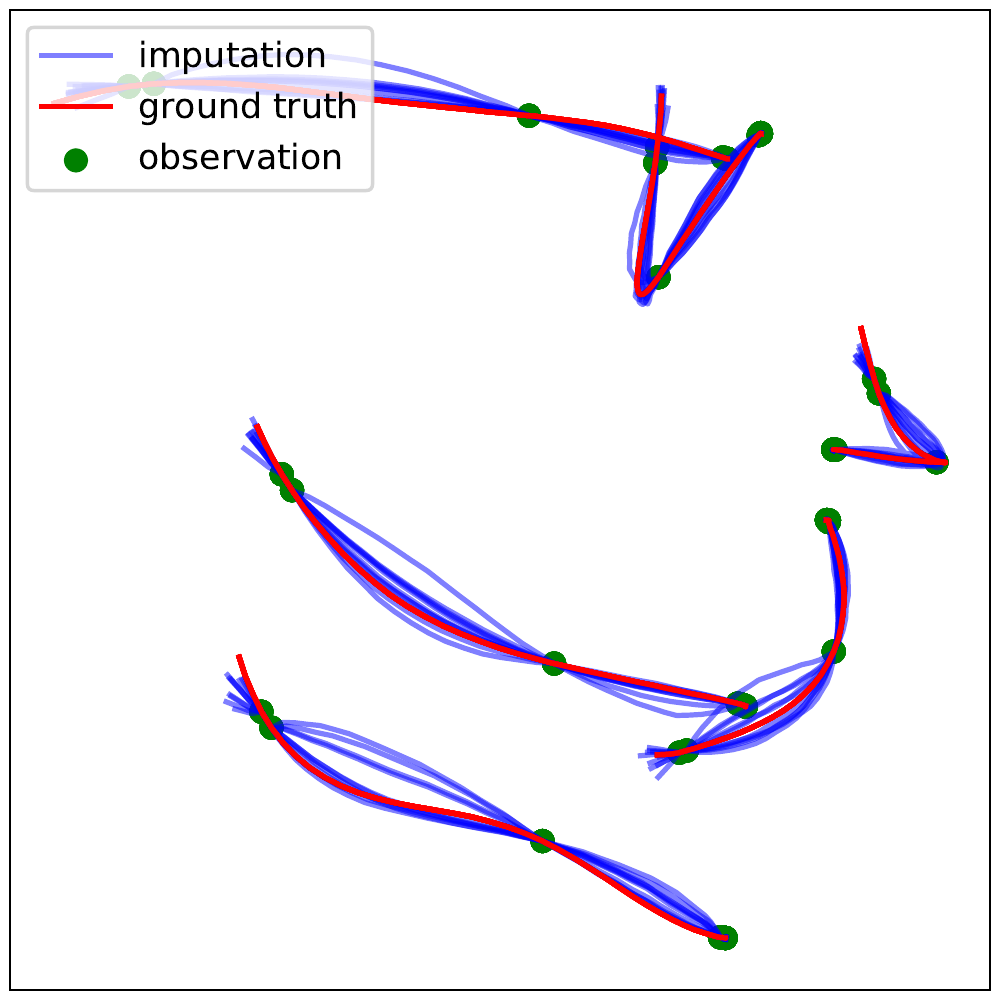}}
\subfigure{\includegraphics[width=0.30\linewidth]{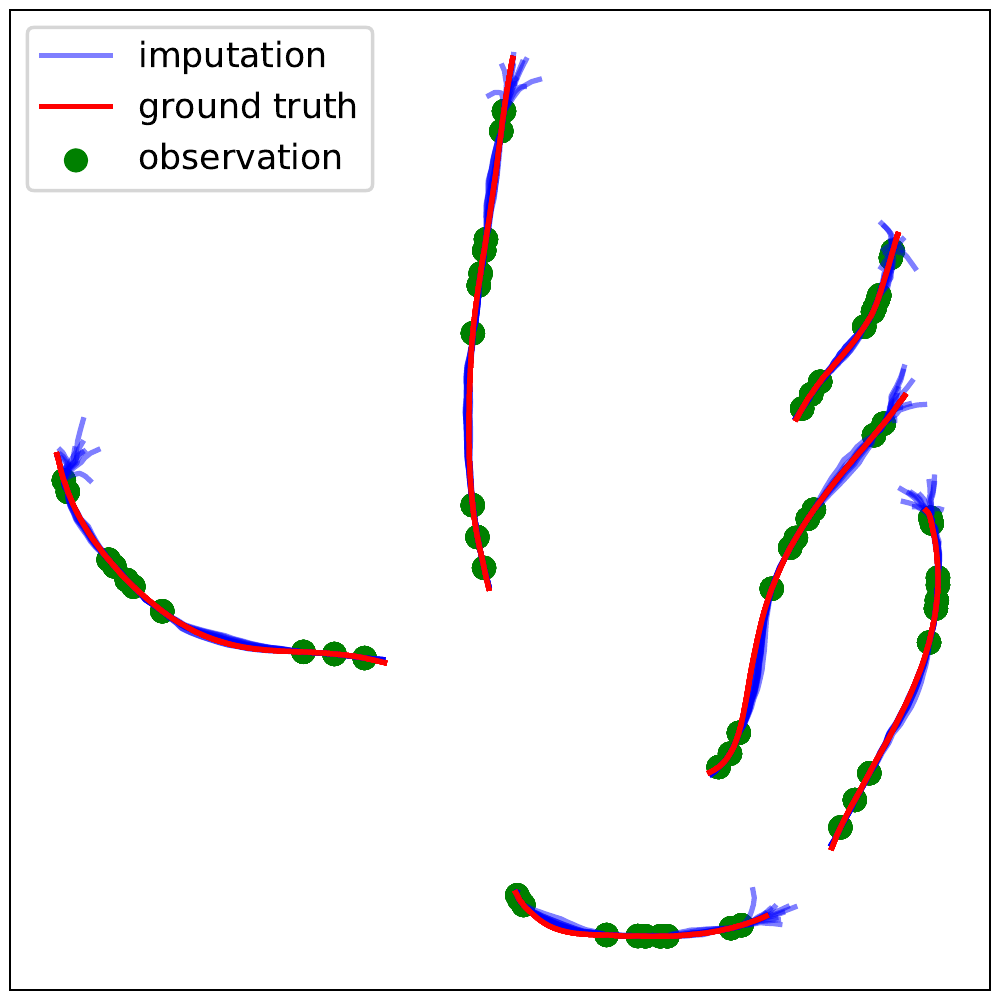}}
\subfigure{\includegraphics[width=0.30\linewidth]{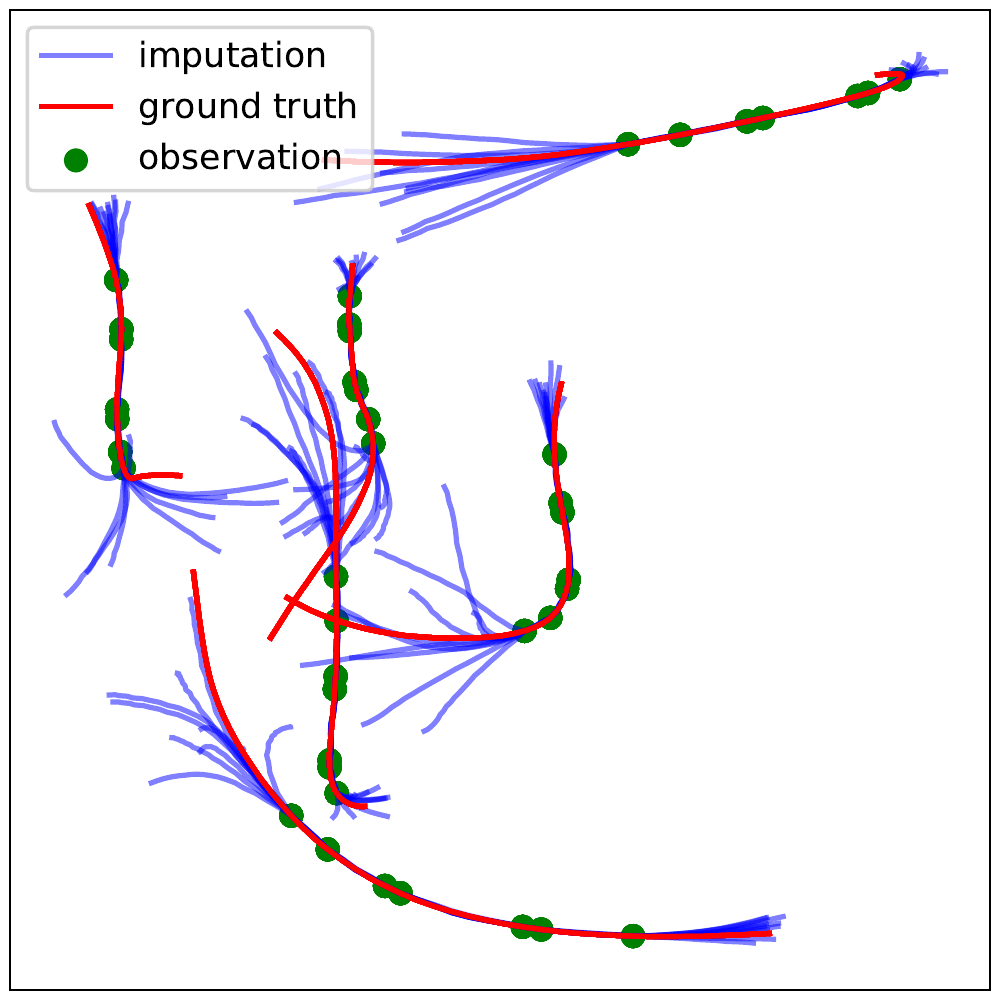}}
\subfigure{\includegraphics[width=0.30\linewidth]{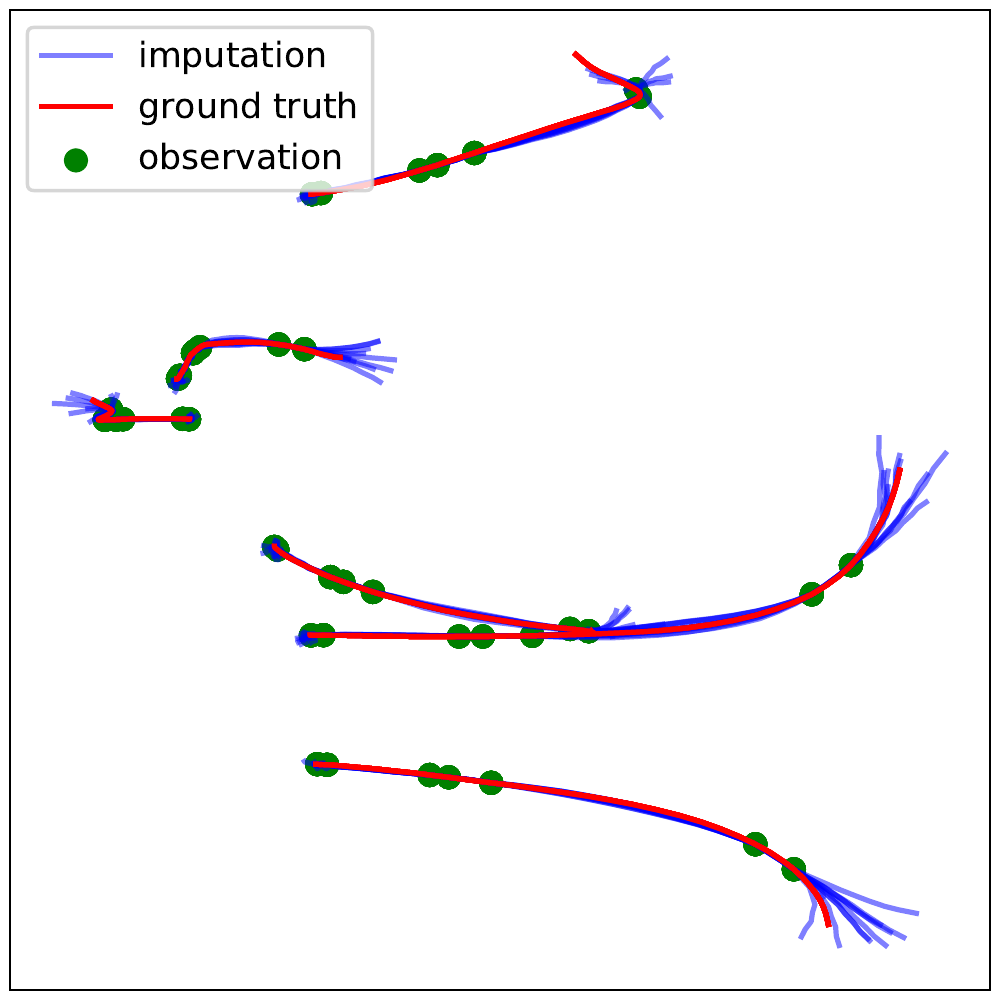}}
\subfigure{\includegraphics[width=0.30\linewidth]{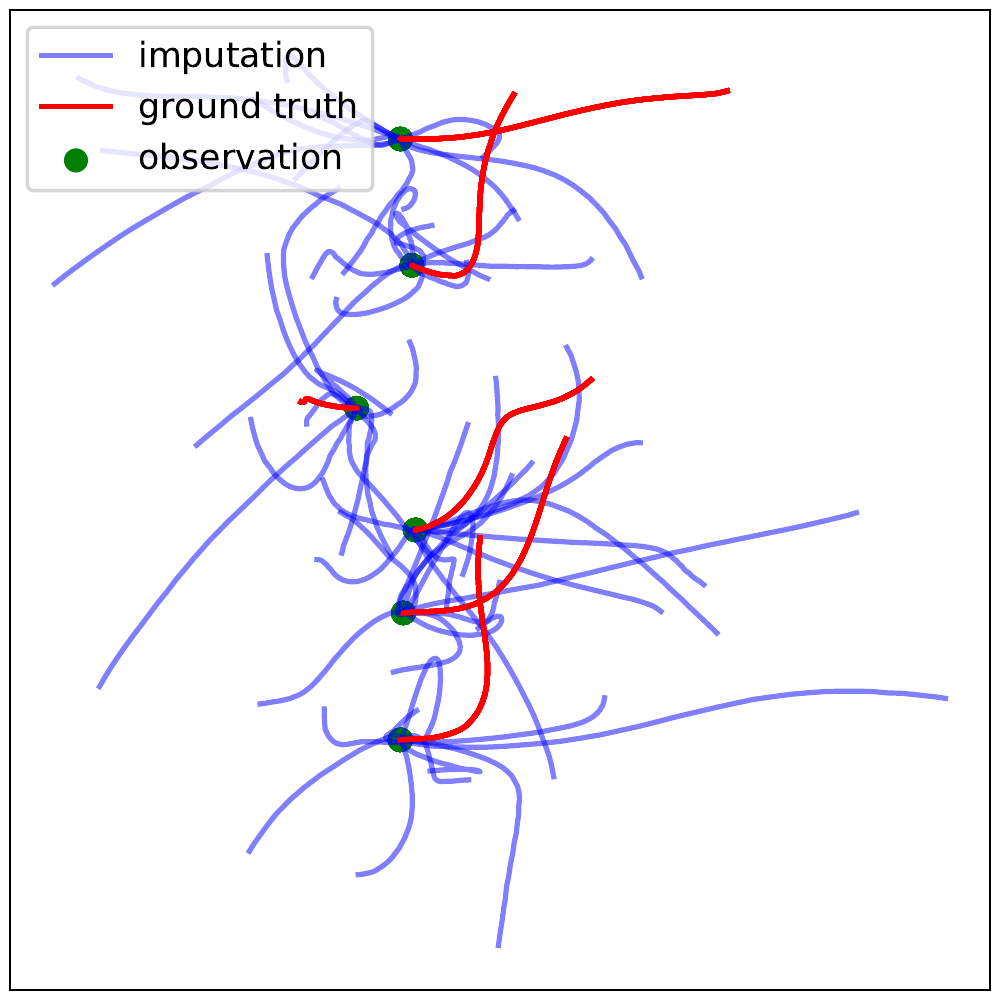}}
\subfigure{\includegraphics[width=0.30\linewidth]{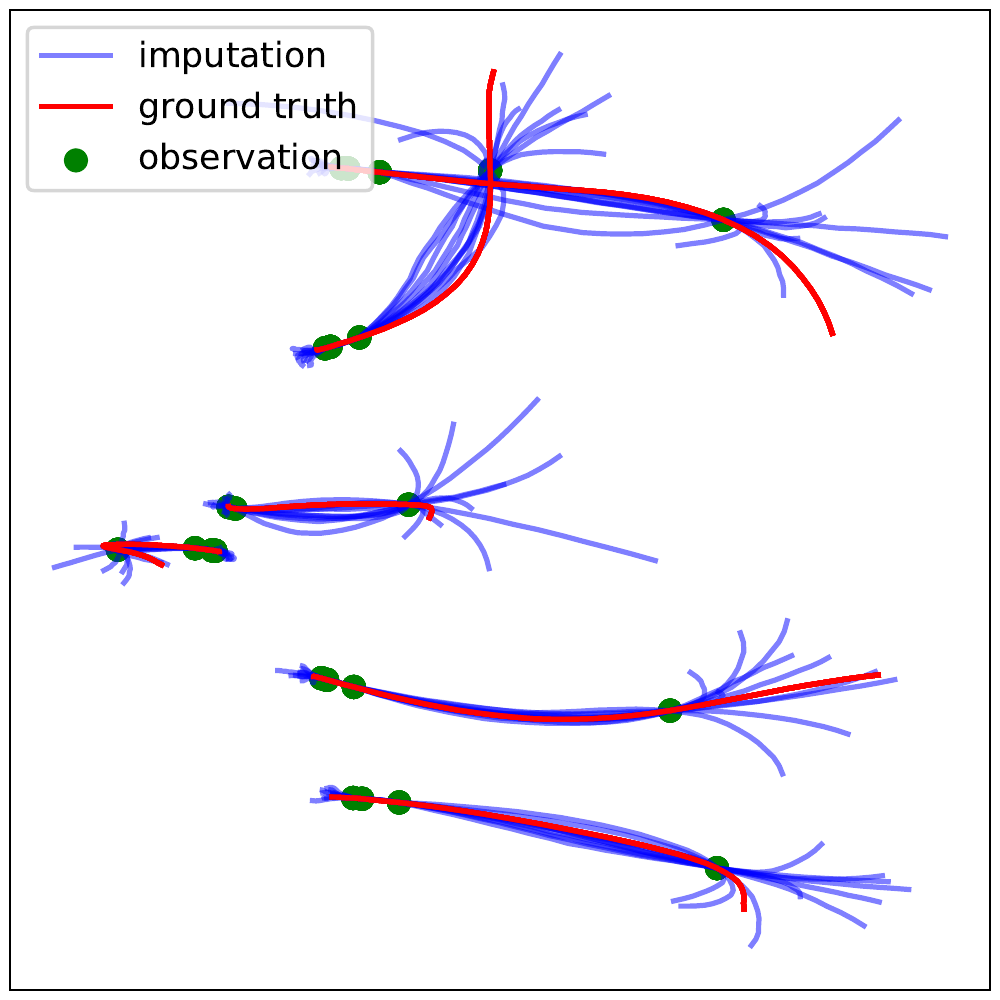}}
  \caption{Stochastic trajectories imputed by NRTSI on the football player trajectory dataset. Green points are observed locations, red lines are ground truth trajectories, and blue lines are imputations sampled independently.}
  \label{fig:nfl_additional}
\end{figure*}

We find that the usually used layer normalization \cite{ba2016layer} and Dropout \cite{srivastava2014dropout} techniques in Transformer are detrimental to imputation performance, and they are removed from our model. 

In practice, we find it beneficial to add a mask to every attention map to prevent elements in $\mathbf{G}^{(1)}$ from attending to each other. This modification encourages elements in $\mathbf G^{(1)}$ to pay more attention to $\mathbf S^{(1)}$ and reduces the mutual influence between elements in $\mathbf G^{(1)}$.

\section{Training Details of NRTSI}
\label{sec:traning_detail}
For all the datasets considered in our paper, we use a maximum resolution level of 4, which corresponds to a maximum missing gap of $2^4 = 16$ that our model can impute. Except for the partially observed dimension scenario, every resolution level is taken care of by a corresponding imputation model. For the partially observed dimension scenario, we find a single model is enough. We train NRTSI with Adam \cite{kingma2014adam} using a starting learning rate of $1\times10^{-4}$ and decrease it to $1\times10^{-5}$ when the validation loss reaches a plateau. 

 When we start training on level $l$, we adopt the transfer learning strategy to initialize the parameters of $f_\theta^{l}$ from the trained weights of model $f_\theta^{l+1}$. The assumption is that the trained higher-level model already captures the dynamics at a fine-grained scale, and it is beneficial to train lower-level models based on the prior knowledge of higher-level dynamics. 

During training, we regard 195 timestamps as missing for both the regularly-sampled and the irregularly-sampled Billiards dataset, 140 timestamps as missing for the traffic dataset, 90 timestamps as missing for the MuJoCo dataset, 40 to 49 timestamps as missing for the football player trajectory dataset, 90 timestamps as missing for the irregularly-sampled sinusoidal dataset. For the gas dataset and the air quality dataset, we use a missing rate of 80\% during training.  

We use an NVIDIA GeForce GTX 1080 Ti GPU to train our model. We implement our model using PyTorch \cite{paszke2019pytorch}.

\section{Visualization of Attention} 
\label{sec:vis_att}
Visualizing the attention maps at different self-attention blocks and different heads is helpful to understand how NRTSI imputes missing data. 

We first visualize the attention maps of an attention head in NRTSI in Figure \ref{fig:att} where this attention head models short-range interactions in a single direction. Then, we show several other representative patterns captured by NRTSI in Figure \ref{fig:billiard_att}. According to Figure \ref{fig:billiard_att}, short-range interactions are captured by Block 1 Head 8, Block 5 Head 10, Block 7 Head 5, and Block 7 Head 7. Middle range interactions are captured by Block 1 Head 3, Block 5 Head 1, and Block 5 Head 5. Long-range interactions are captured by Block 1 Head 9 and Block 7 Head 12. It is interesting to find that some attention heads capture interactions on both the forward and backward directions (e.g. Block 1 Head 3, Block 1 head 8, Block 1 Head 9, Block 7 Head 12), while some attention heads capture interactions in a single direction (e.g. Block 5 Head 1, Block 5 Head 5, Block 5 Head 10, Block 7 Head 5, Block 7 Head 7).

\section{Datasets Details}
\label{sec:dataset}
All the datasets can be downloaded or generated from the codes submitted. In the ``readme.md'' file, links to download these datasets can be found. We also include the python scripts to generate the irregularly-sampled Billiards dataset and the irregularly-sampled sinusoidal function dataset.
\label{sec:dataset_details}
\subsection{Billiards Dataset} We download this dataset from the official GitHub repository \footnote{\url{https://github.com/felixykliu/NAOMI}} of NAOMI \cite{liu2019naomi} to guarantee our results are comparable to the results in the NAOMI paper. Note that this dataset does not have a validation set because NAOMI does not have a validation set. To make our results comparable to NAOMI we have to follow its setup.  
\subsection{{PEMS-SF traffic dataset}} We follow NAOMI to download this dataset form \url{https://archive.ics.uci.edu/ml/datasets/PEMS-SF} and follow NAOMI to use the same default training/testing split. Note that this dataset does not have a validation set because NAOMI does not have a validation set. To make our results comparable to NAOMI we have to follow its setup. 
\subsection{MuJoCo dataset} We download this dataset from the official GitHub repository \footnote{\url{https://github.com/YuliaRubanova/latent_ode}} of Latent ODE \cite{rubanova2019latent} to guarantee our results are comparable to the results in Latent ODE paper. We follow Latent ODE to randomly split the dataset into 80\% train and 20\% test with the random seed. Note that this dataset does not have a validation set because Latent ODE does not have a validation set. To make our results comparable to Latent ODE we have to follow its setup.
\subsection{Irregularly-sampled Billiards Dataset} This dataset is created with the same parameters (e.g. initial ball speed range, travel time) to create its regularly-sampled counterpart in Section 4.1 in the main text. The only difference is that this dataset is irregularly-sampled. The billiard table is square and the side length of the square is 0.8828. The initial ball speed range is 0.0018 to 0.1075. The travel time for each trajectory is 200 seconds. We randomly generate 12,000 trajectories for training and 1,000 trajectories for testing.
\subsection{The Gas Sensor Dataset and the Air Quality Dataset} We strictly follow RDIS \cite{choi2020rdis} to use the same strategy to prepare the gas sensor dataset and the air quality dataset. Please refer to the RDIS paper for details. For a fair comparison, we use the same random seed to perform train/validation/test split as done in RDIS.
\subsection{Football Player Trajectory Dataset} We collect this football player trajectory dataset from the 2021 Big Data Bowl data \cite{nfl}. For American football games, the number of players on each team is 11. However, data of linemen are not provided in \cite{nfl}. As a result, the number of players is less than 11. To collect a dataset with the same number of players, we only use data with exactly 6 offensive players.

This dataset is publicly available and does not contain personally identifiable information or offensive content.

\section{Additional Visualization of NRTSI} 
\label{sec:nrtsi_add_vis}
In this section, we provide more visualization results of the imputed trajectories on the regularly-sampled Billiards dataset in Figure \ref{fig:billiard}, the irregularly-sampled Billiards dataset in Figure \ref{fig:irrbilliard}, as well as the football player trajectory dataset in Figure \ref{fig:nfl_additional}. These figures are randomly selected without cherry-picking.

\section{Visualization of the Trajectories imputed by Latent ODE, Neural CDE and ANP} 
\label{sec:viz_latentode_neuralcde}
\begin{figure*}[h]
\center
\subfigure[150 timestamps observed]{\includegraphics[width=0.24\linewidth, trim={2.03cm 1.3cm 1cm 1.08cm},clip]{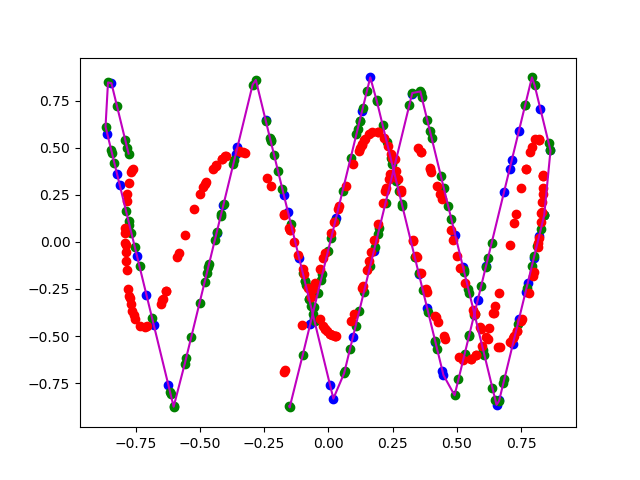}}
\subfigure[100 timestamps observed]{\includegraphics[width=0.24\linewidth, trim={2.03cm 1.3cm 1cm 1.08cm},clip]{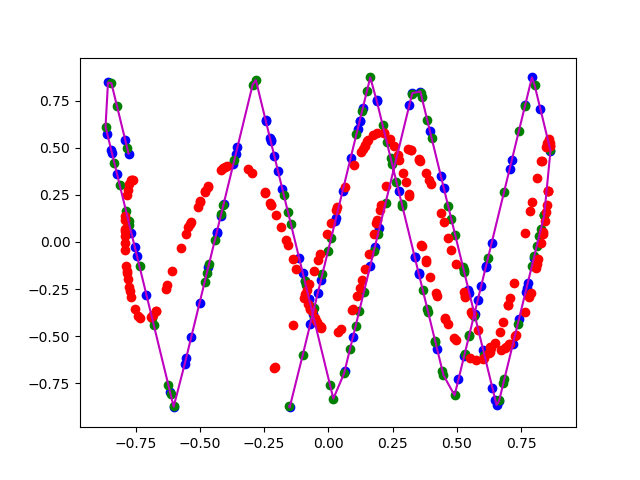}}
\subfigure[50 timestamps observed]{\includegraphics[width=0.24\linewidth, trim={2.03cm 1.3cm 1cm 1.08cm},clip]{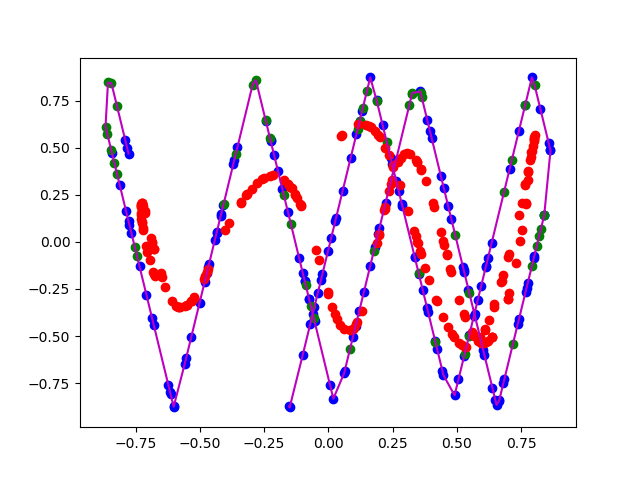}}
\subfigure[5 timestamps observed]{\includegraphics[width=0.24\linewidth, trim={2.03cm 1.3cm 1cm 1.08cm},clip]{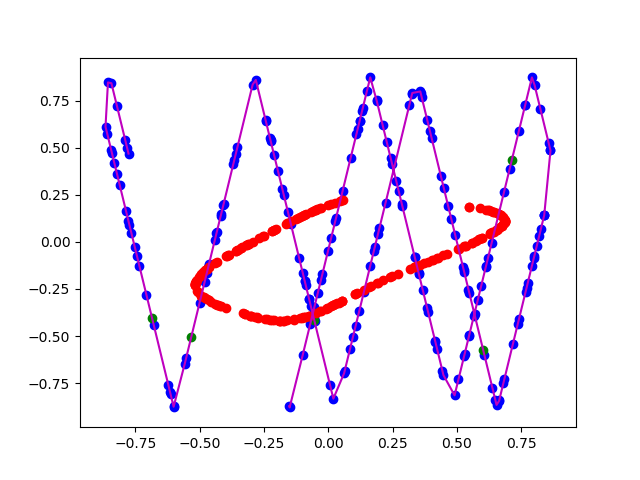}}
  \caption{Trajectories imputed by Latent ODE on the irregularly-sampled Billiards dataset. Red points are the imputed data, green points are the observed data and the blue points are the ground-truth data.}
  \label{fig:latentode}
\end{figure*}

\begin{figure*}[h]
\center
\subfigure[150 timestamps observed]{\includegraphics[width=0.24\linewidth, trim={2.03cm 1.3cm 1cm 1.08cm},clip]{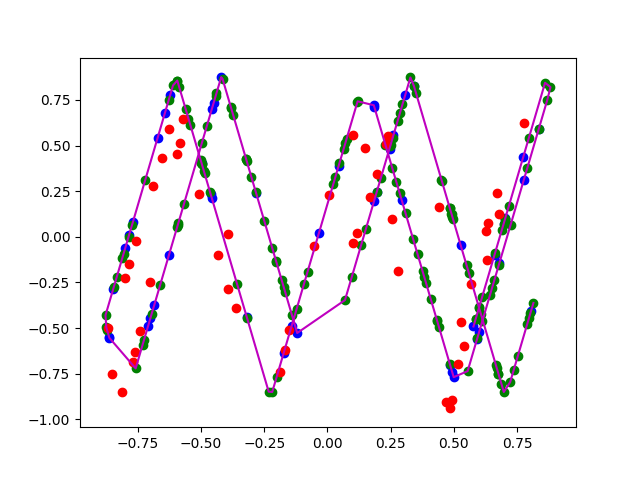}}
\subfigure[100 timestamps observed]{\includegraphics[width=0.24\linewidth, trim={2.03cm 1.3cm 1cm 1.08cm},clip]{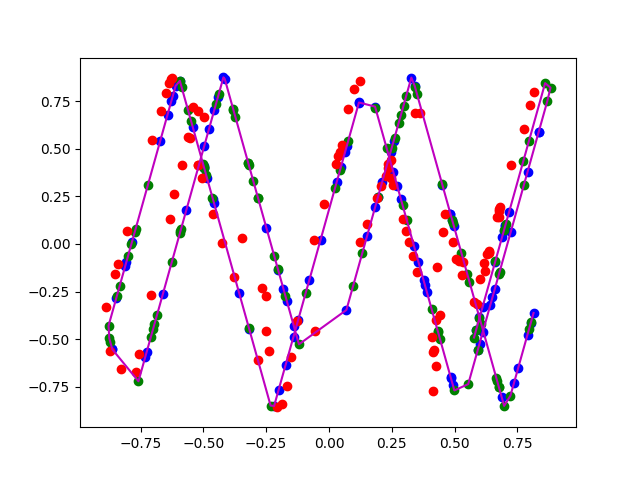}}
\subfigure[50 timestamps observed]{\includegraphics[width=0.24\linewidth, trim={2.03cm 1.3cm 1cm 1.08cm},clip]{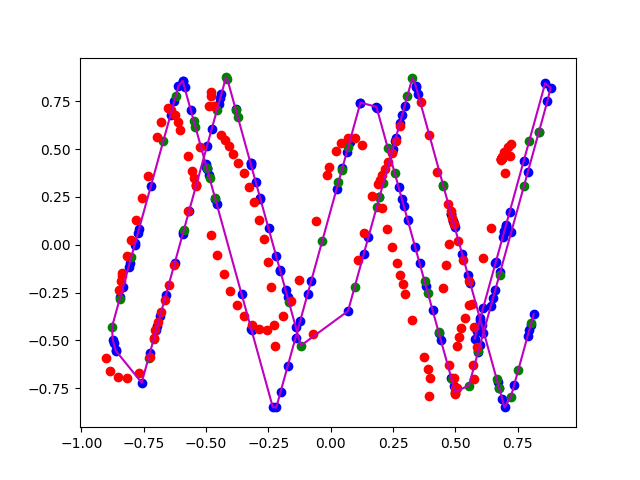}}
\subfigure[5 timestamps observed]{\includegraphics[width=0.24\linewidth, trim={2.03cm 1.3cm 1cm 1.08cm},clip]{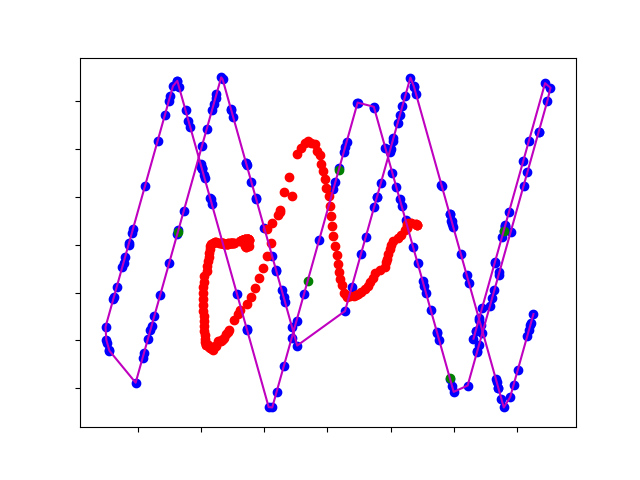}}
  \caption{Trajectories imputed by Neural CDE on the irregularly-sampled Billiards dataset. Red points are the imputed data, green points are the observed data and the blue points are the ground-truth data.}
  \label{fig:neuralcde}
\end{figure*}

\begin{figure*}[h]
\center
\subfigure[150 timestamps observed]{\includegraphics[width=0.24\linewidth, trim={2.03cm 1.3cm 1cm 1.08cm},clip]{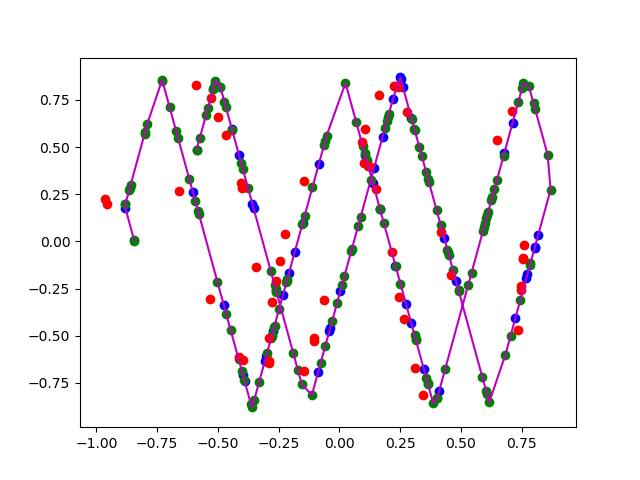}}
\subfigure[100 timestamps observed]{\includegraphics[width=0.24\linewidth, trim={2.03cm 1.3cm 1cm 1.08cm},clip]{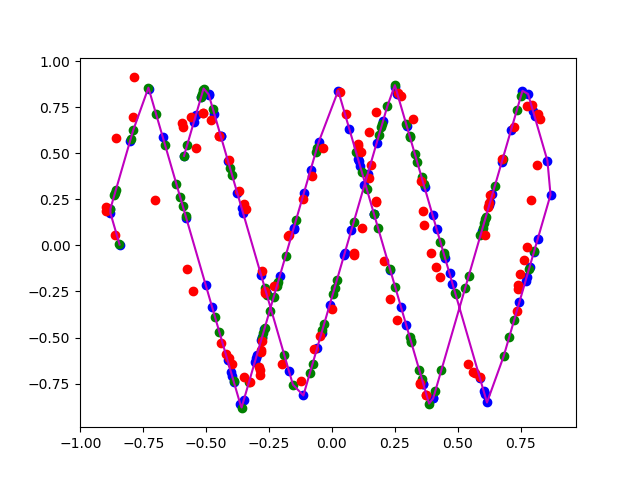}}
\subfigure[50 timestamps observed]{\includegraphics[width=0.24\linewidth, trim={2.03cm 1.3cm 1cm 1.08cm},clip]{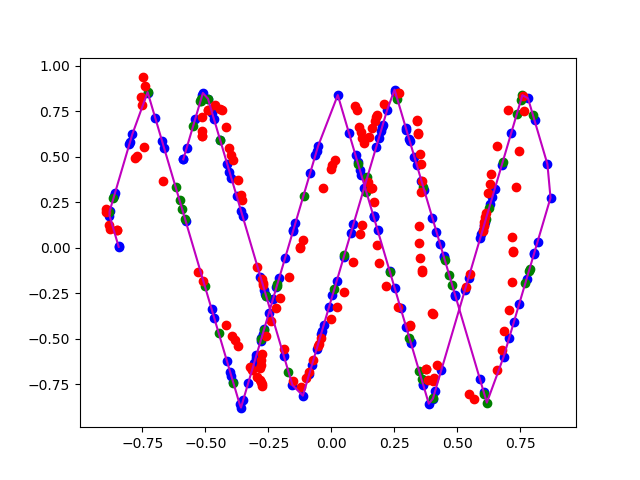}}
\subfigure[5 timestamps observed]{\includegraphics[width=0.24\linewidth, trim={2.03cm 1.3cm 1cm 1.08cm},clip]{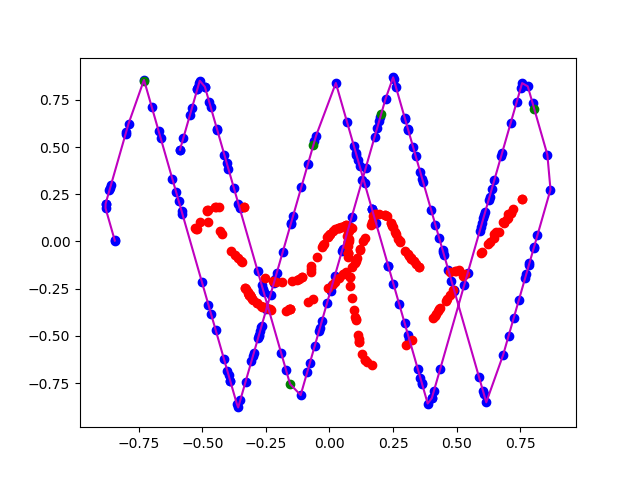}}
  \caption{Trajectories imputed by ANP on the irregularly-sampled Billiards dataset. Red points are the imputed data, green points are the observed data and the blue points are the ground-truth data.}
  \label{fig:anp}
\end{figure*}

Latent ODE \cite{rubanova2019latent} Neural CDE \cite{kidger2020neural} are among the best performing Neural ODE methods for modeling irregularly-sampled time series. ANP \cite{kim2018attentive} naturally handles irregularly-sampled data as times are directly fed into ANP as scalars. However, we find in the irregularly-sampled Billiards experiment that their performance deteriorates significantly when the observation become sparse. We visualize the billiard trajectories imputed by Latent ODE and Neural CDE and ANP with different number of observed data respectively in Figure \ref{fig:latentode},  \ref{fig:neuralcde} and \ref{fig:anp}. From Figure \ref{fig:latentode} and  \ref{fig:neuralcde}, we can see the poor performance of Latent ODE and Neural CDE is caused by the error compounding problem. For ANP, we discuss in Appendix \ref{sec:nrtsi_ablation} that removing some essential components of NRTSI negatively impacts the imputation performance, and the resulting model after the removal reduces to ANP. This explains the poor performance of ANP.

\begin{figure*}[h]
\center
\subfigure{\includegraphics[width=0.16\linewidth, trim={0.94cm 0.6cm 0.6cm 0.6cm},clip]{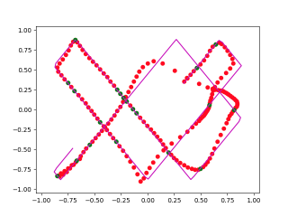}}
\subfigure{\includegraphics[width=0.16\linewidth, trim={0.94cm 0.6cm 0.6cm 0.6cm},clip]{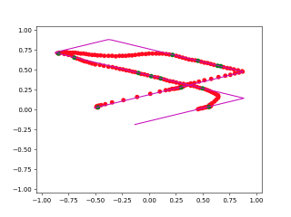}}
\subfigure{\includegraphics[width=0.16\linewidth, trim={0.94cm 0.6cm 0.6cm 0.6cm},clip]{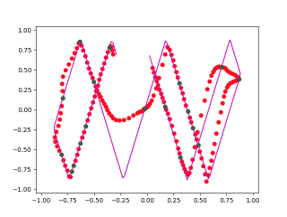}}
\subfigure{\includegraphics[width=0.16\linewidth, trim={0.94cm 0.6cm 0.6cm 0.6cm},clip]{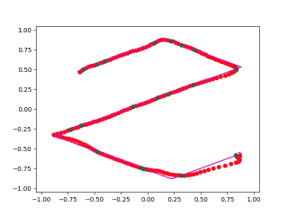}}
\subfigure{\includegraphics[width=0.16\linewidth, trim={0.94cm 0.6cm 0.6cm 0.6cm},clip]{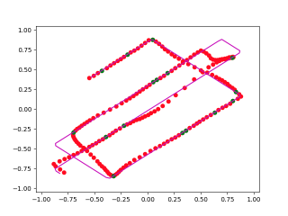}}
\subfigure{\includegraphics[width=0.16\linewidth, trim={0.94cm 0.6cm 0.6cm 0.6cm},clip]{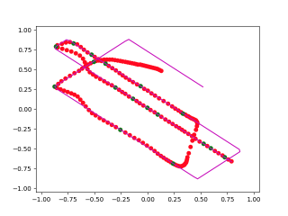}}
  \caption{Qualitative results of \texttt{NRTSI w/o Hierarchy + small} on the Billiards dataset. The purple solid lines are ground-truth trajectories, the red points are imputed data and the green points are observed data.}
  \label{fig:NRTSI-small}
\end{figure*}

\begin{figure*}[h]
\center
\subfigure{\includegraphics[width=0.16\linewidth, trim={2.05cm 1.34cm 0.8cm 0.6cm},clip]{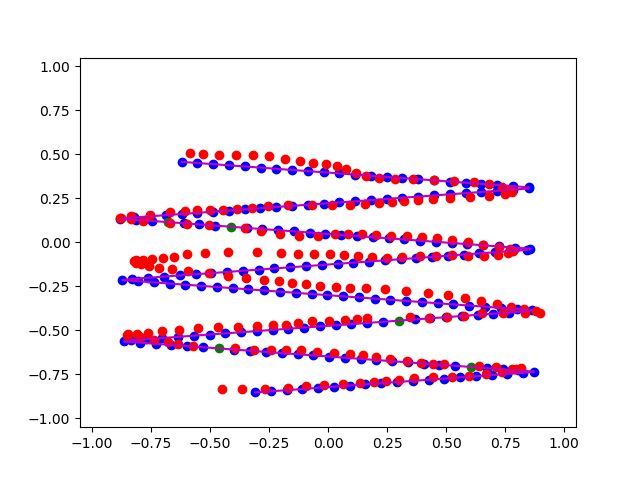}}
\subfigure{\includegraphics[width=0.16\linewidth, trim={2.05cm 1.34cm 0.8cm 0.6cm},clip]{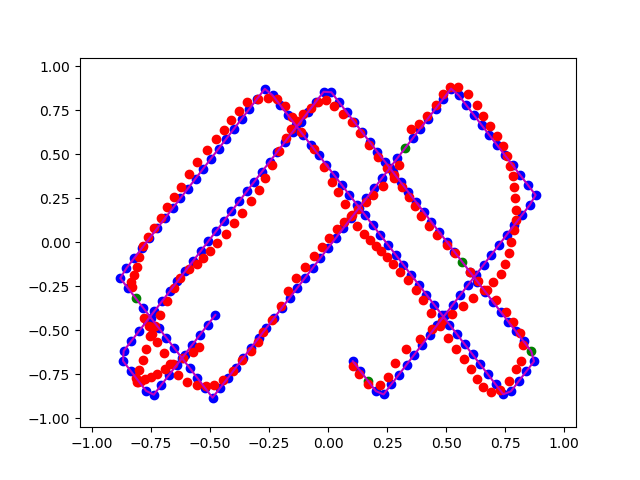}}
\subfigure{\includegraphics[width=0.16\linewidth, trim={2.05cm 1.34cm 0.8cm 0.6cm},clip]{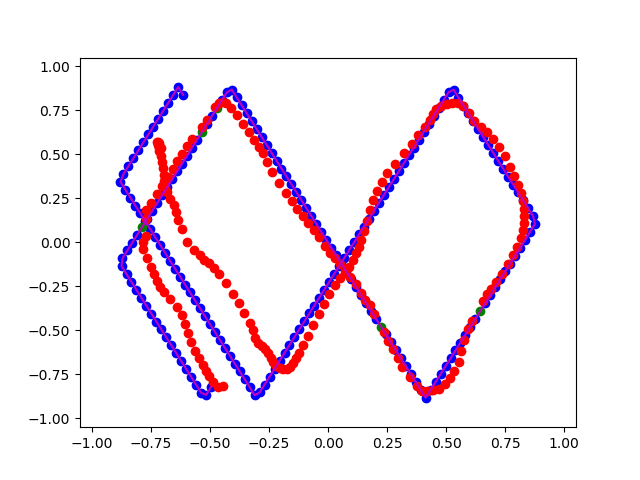}}
\subfigure{\includegraphics[width=0.16\linewidth, trim={2.05cm 1.34cm 0.8cm 0.6cm},clip]{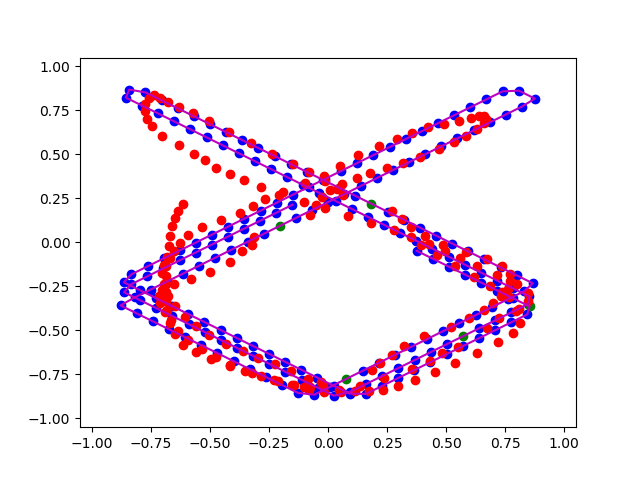}}
\subfigure{\includegraphics[width=0.16\linewidth, trim={2.05cm 1.34cm 0.8cm 0.6cm},clip]{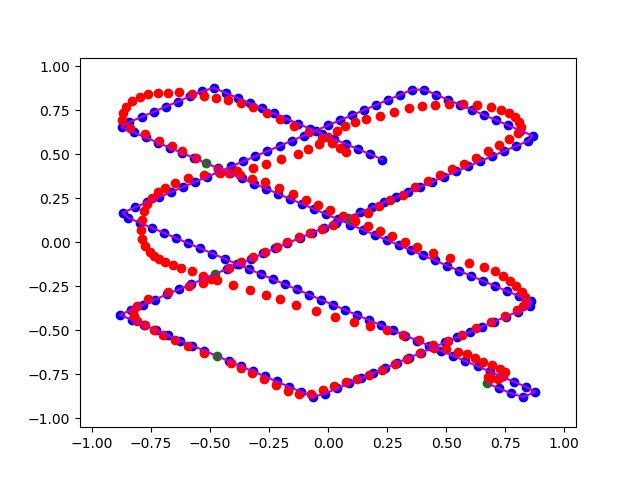}}
\subfigure{\includegraphics[width=0.16\linewidth, trim={2.05cm 1.34cm 0.8cm 0.6cm},clip]{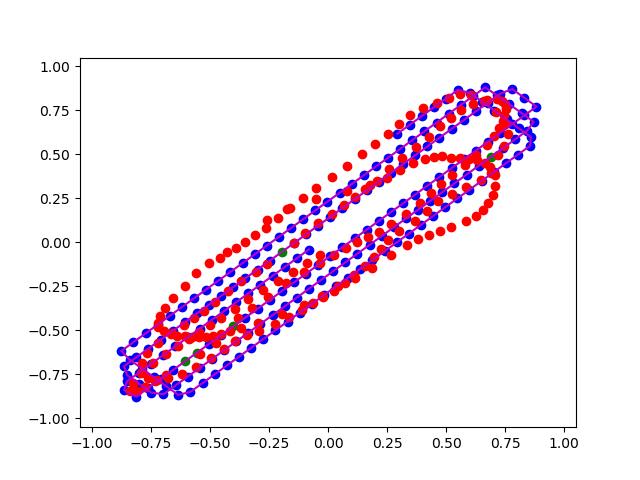}}
  \caption{Qualitative results of \texttt{NRTSI w/o Hierarchy + all} on the Billiards dataset. The purple solid lines and the blue points are ground-truth trajectories, the red points are imputed data and the green points are observed data.}
  \label{fig:NRTSI-all}
\end{figure*}

\begin{figure*}[h]
\center
\subfigure[Latent-ODE]{\includegraphics[width=0.4\linewidth, trim={2.03cm 1.3cm 1cm 1.08cm},clip]{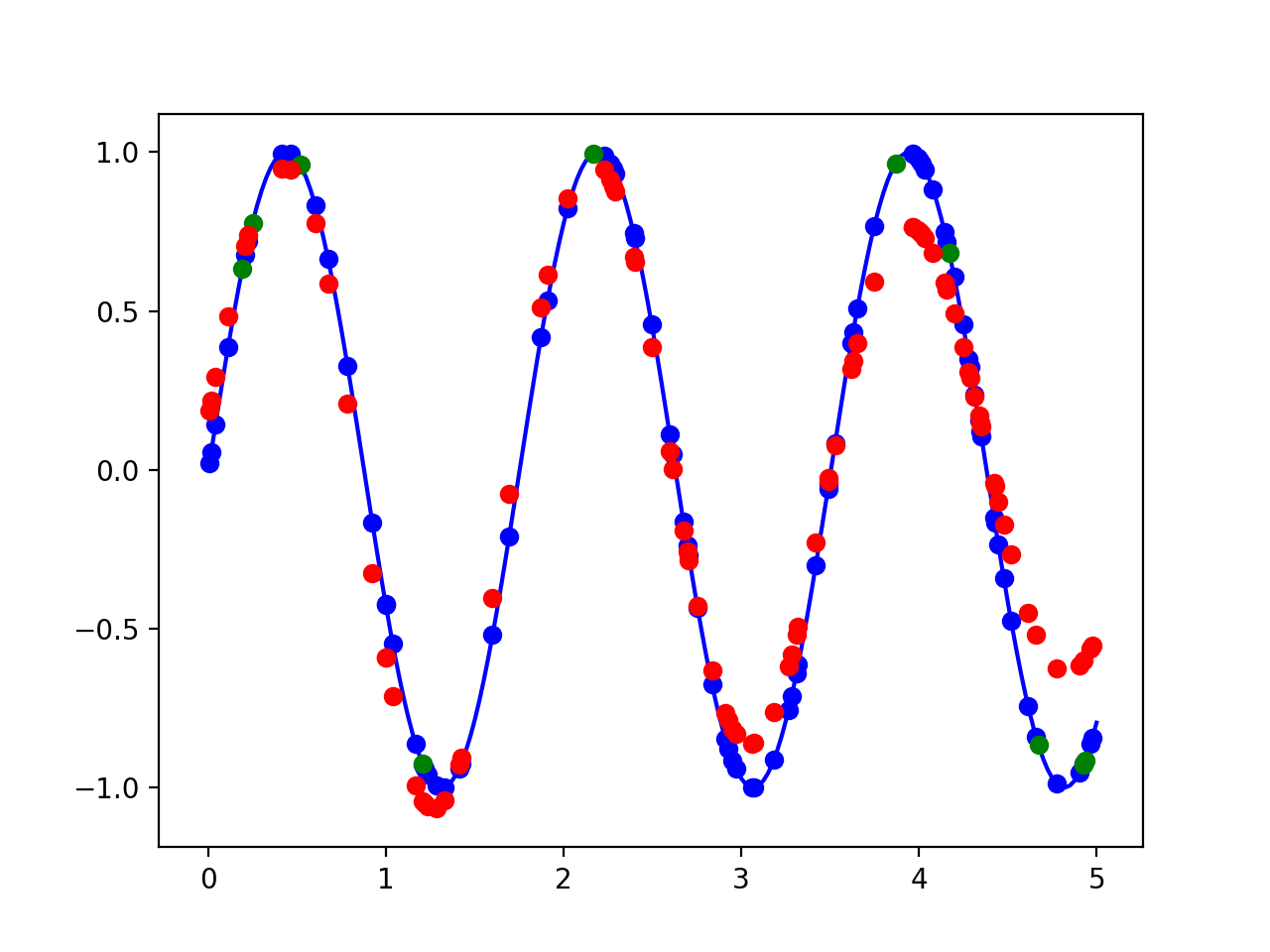}}
\subfigure[NRTSI]{\includegraphics[width=0.4\linewidth, trim={2.03cm 1.3cm 1cm 1.08cm},clip]{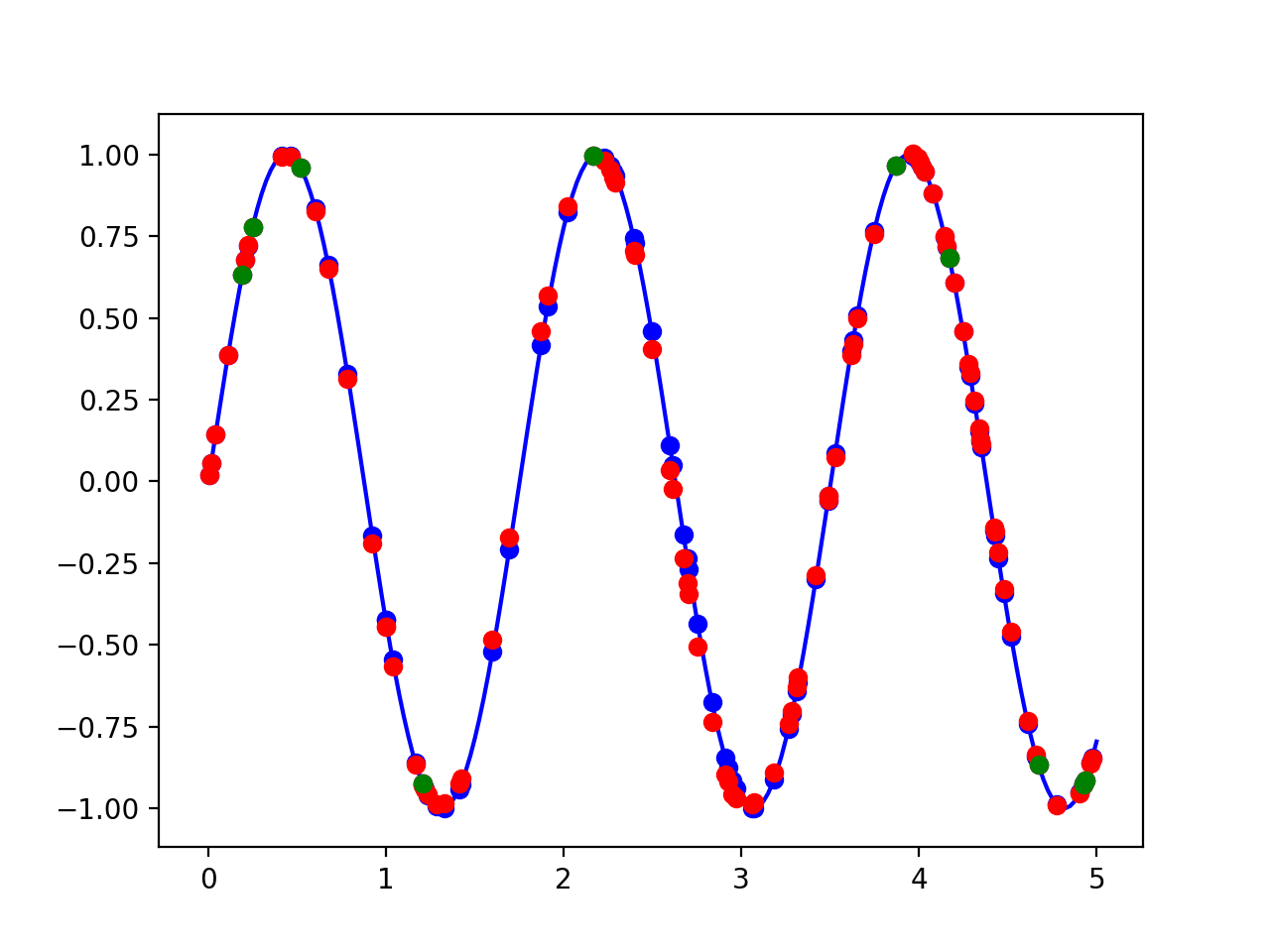}}
  \caption{Trajectories imputed by Latent ODE and NRTSI on the irregularly-sampled sinusoidal function dataset. Red points are the imputed data, green points are the observed data and the blue points are the ground-truth data.}
  \label{fig:irr_sin}
\end{figure*}

\section{Ablation Studies of NRTSI}
\subsection{Ablation Studies of the Essential Components}
\label{sec:nrtsi_ablation}
In this section, we perform several ablation studies on the regularly-sampled Billiard dataset to evaluate the contribution of each component in NRTSI. We consider four variants of NRTSI and report their L2 losses in Table \ref{table:ablation}. 

From Table \ref{table:ablation}, it can be seen that not using the proposed hierarchical imputation strategy (\texttt{NRTSI w/o Hierarchy}) deteriorates the performance most. More specifically, the variant that imputes data with the smallest missing gap first (\texttt{NRTSI w/o Hierarchy + small}) is very similar to recurrent methods that always impute data at the next nearby timestamps first. As a result, this variant suffers from the error compounding problem as shown qualitatively in Figure \ref{fig:NRTSI-small}. Another variant that imputes all the missing data at once (\texttt{NRTSI w/o Hierarchy + all}) is similar to the way that ANP \cite{kim2018attentive} imputes. We show its performance quantitatively in Table \ref{table:ablation} and qualitatively in Figure \ref{fig:NRTSI-all}. 

Another variant \texttt{NRTSI w/o Encode $\mathbf{S} \cup \mathbf{G}$} only uses the Transformer encoder to model $\mathbf {S}$ rather than $\mathbf {S} \cup \mathbf G$. Then $\mathbf{G}$ attends to the representation of $\mathbf{S}$ computed by the Transformer encoder via the cross-attention mechanism. This variant also deteriorates the performance of NRTSI because the information of what timesteps to impute (i.e. $\mathbf{G}$) is not utilized when the Transformer encoder uses self-attention to compute the representations of observed data (i.e. $\mathbf S$). Therefore, the representation might be suboptimal. This model architecture is very similar to ANP \cite{kim2018attentive} that also ignores the target input when ANP uses self-attention to compute the representations of the context input/output pairs. This shortcoming and the fact that ANP imputes all missing data at once together explain why NRTSI outperforms ANP by a large margin in Table \ref{table:irr_billiard}.

Using a single model to impute at all hierarchy levels (\texttt{NRTSI w/o Individual}) also slightly deteriorates the performance. This is because imputing at different hierarchical levels requires different types of attention patterns, and it might be hard to capture all the necessary patterns in a single model.    

\begin{table}
    \caption{Ablation studies of NRTSI. All the methods use the same Transformer encoder architecture hyperparameters according to Appendix \ref{sec:NRTSI_arch}.}
    \begin{center}
    \setlength{\tabcolsep}{4pt}
    \begin{scriptsize}
    \begin{tabular}{c|cc}
        \toprule
        Method Name  & Description & L2 loss ($\times 10^{-2}$) \\
        \midrule
        \texttt{NRTSI w/o Individual} & \makecell[c]{A variant of NRTSI that uses a single imputation model to impute \\ at all hierarchies.} & 0.16\\
        \texttt{NRTSI w/o Encode $\mathbf{S} \cup \mathbf{G}$} & \makecell[c]{A variant of NRTSI that uses the Transformer encoder to only  model $\mathbf S$. \\ Then $\mathbf G$ attends to the representation of $\mathbf S$ via cross-attention.} & 0.27 \\
        \texttt{NRTSI w/o Hierarchy + all} & \makecell[c]{A variant of NRTSI that imputes all missing data at once  without the \\ proposed hierarchical strategy.} & 1.14 \\
        \texttt{NRTSI w/o Hierarchy + small} & \makecell[c]{A variant of NRTSI that imputes data with the smallest missing gap first \\ without the proposed hierarchical strategy.} & 11.09 \\ 
        \midrule
        NRTSI & The proposed method. & \textbf{0.024}  \\
        \bottomrule
    \end{tabular}
    \end{scriptsize}
    \end{center}
    \label{table:ablation}
\end{table}

\begin{figure}[t]
\centering
\includegraphics[width=0.3\columnwidth]{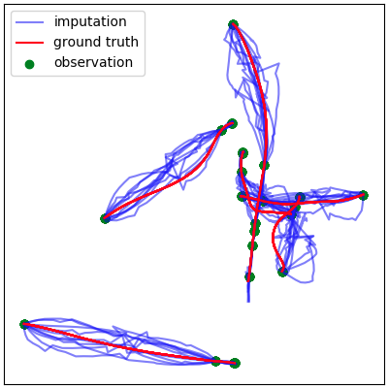} 
\caption{Imputed trajectories with the threshold set to 16.}
\label{fig:nfl_all_parallel}
\end{figure}

\subsection{Stochastic Level Threshold} According to Section 3.4 of the main text, NRTSI handles stochastic datasets by imputing data with large missing gaps one by one while imputing data with smaller missing gaps in parallel. To report the performance of NRTSI on the football player trajectory dataset in the main text, we use a threshold of 4 where data with missing gaps greater than this threshold are regarded as large-missing-gap data while data with missing gaps equal or smaller than this threshold are regarded as small-missing-gap data. 

In this section, we perform an ablation study to investigate the influence of this threshold on the imputation performance.
\begin{table}[t]
\begin{scriptsize}
\caption{Ablation studies of the stochastic level threshold on the football player trajectory dataset.}
\vspace{+0.1in}
\label{table:nfl_ablation}
\begin{center}
\begin{tabular}{c|cccc|c}
\toprule
 \multirow{2}{*}{Method} & \multicolumn{4}{c}{Threshold} & \ \multirow{2}{*}{Expert}\\
  & 2  & 4 & 8 & 16 & \\
\midrule
\textbf{step change ($\times 10^{-3}$)} &  \textbf{2.425} & 2.401 & 2.370 & 2.560 & 2.482   \\
\textbf{avg length} &  0.176 & \textbf{0.175} & 0.178 &  0.179 & 0.173 \\
\textit{min}\textbf{MSE} ($\times 10^{-3}$) &  1.956 & \textbf{1.908}& 1.915  & 1.926 & 0.000\\
\textit{avg}\textbf{MSE} / \textit{min}\textbf{MSE} & \textbf{2.14} & 2.13 & 2.09 & 2.12 &  --- \\
\textbf{Imputation Time (s)} & 0.901 & 0.512 & 0.485 & \textbf{0.436} & ---\\
\bottomrule
\end{tabular}
\end{center}
\end{scriptsize}
\end{table}
According to Table \ref{table:nfl_ablation}, the imputation speed increases as we increase the threshold, which is expected as we are imputing more data in parallel with a larger threshold. However, we find the quality of the imputed trajectories deteriorates when setting the threshold to 16, in which case our model imputes all data in parallel. We have discussed the cause of this quality deterioration problem in Section 3.4 in the main text. It is shown that imputing with a threshold of 16 yields the worst average trajectory length, as there are artifacts and abrupt changes in the imputed trajectories as shown in Figure \ref{fig:nfl_all_parallel}  due to incongruous predictions. Since using a threshold of 4 gives the overall best imputation statistics, has a fast imputation speed, and generates realistic imputation, we use a threshold of 4 to report the experiment results in the main text.

\begin{table*}[t]
\centering
\begin{small}
\begin{tabular}{@{}c|c|cccccccc@{}}
\toprule
\multirow{2}{*}{Dataset}      & \multirow{2}{*}{Method} & \multicolumn{8}{c}{missing rate}                                                                                                              \\
                              &                           & 10\%            & 20\%            & 30\%            & 40\%            & 50\%            & 60\%            & 70\%            & 80\%            \\ \midrule
\multirow{12}{*}{Air Quality} & Forward                   & .2472          & .2638          & .2707          & .2694          & .2747          & .3264          & .3888          & .4432          \\
                              & Backward                  & .1849          & .2137          & .2847          & .2832          & .2799          & .3333          & .3624          & .4445          \\
                              & MICE                      & .3405          & .4135          & .3888          & .4235          & .4309          & .5663          & -               & -               \\
                              & KNN                       & .1520           & .2323          & .2847          & .3501          & .3930           & .4327          & .4843          & .5775          \\ 
                              & GAIN                      &     .6133            &     .6761            &   .6947              &   .6915              &     .6914            &   .7256              &     .7305            &    .7503             \\
                              & GRU                       & .3137          & .3617          & .3757          & .3735          & .3834          & .4331          & .4523          & .4962          \\
                              & Bi-GRU                    & .2706          & .3136          & .3336          & .3287          & .3363          & .3902          & .4029          & .4445          \\
                              & M-RNN                      & .1583          & .2127          & .2342          & .2324          & .2417          & .2919          & .3248          & .4038          \\
                              & \texttt{Latent-ODE}                     & .2820          & .2954          & .3161          & .3291          & .3462          & .3569          & .3614          & .3762          \\
                          
                          & \texttt{NeuralCDE}                     & .2951          & .3129          & .3337          & .3524          & .3898          & .4074          & .4192          & .4865          \\
                          
                              & BRITS                     & .1659          & .2076          & .2212          & .2088          & .2141          & .2660          & .2885          & .3421          \\
                           
                              & GRU w/ RDIS                      & .1854 & .2385 & .2600 & .2557 & .2629 & .3075 & .3276 & .3826 \\
                              & Bi-GRU w/ RDIS                   & .1409 & .1807 & .2008 & .1977 & .2041 & .2528 & .2668 & .3178 \\ \cmidrule(l){2-10}  
                                      &      NRTSI-all              & \textbf{.0882} & \textbf{.0865} &  \textbf{.0911} & \textbf{.0986} & \textbf{.1064} & \textbf{.1204} & \textbf{.1398} & \textbf{.1713} \\
& NRTSI               & .1230 & .1155 &  .1189 & .1250 & .1297 & .1378 & .1542 & .1790 \\
\midrule
\multirow{12}{*}{Gas Sensor}  & Forward                   & .0834          & .0987          & .1181          & .1460          & .1845          & .2398          & .3215          & .4739          \\
                              & Backward                  & .0838          & .1016          & .1267          & .1522          & .1894          & .2449          & .3310          & .5501          \\
                              & MICE                      & .0736          & .0822          & .1090          & .1468          & .2147          & -          & -               & -               \\
                              & KNN                       & .0328           & .0356          & .0388          & .1273          & .2423           & .3771          & .4921          & .8009          \\ 
                              & GAIN                      &     .5202            &    .5266             &   .5232              &      .5276          &      .5335           &       .5386          &       .5420          &      .6021           \\ 
                              & GRU                       & .0727          & .0838          & .0940          & .1127          & .1349          & .1647          & .1998          & .3223          \\
                              & Bi-GRU                    & .0657          & .0766          & .0862          & .1024          & .1235          & .1451          & .1802          & .2903          \\
                              & M-RNN                      & .0327          & .0400          & .0378          & .0388          & .0405          & .0427          & .0435          & .1023          \\
                              
                              & \texttt{Latent-ODE}                     & .1251          & .1282          & .1278          & .1299          & .1332          & .1387          & .1487          & .1979          \\
                         
                          & \texttt{NeuralCDE}                     & .0685          & .0773          & .0821          & .1044          & .1251          & .1538          & .1754          & .3011          \\
                              & BRITS                     & .0210          & .0226          & .0233          & .0279          & .0338          & .0406          & .0518          & .1595          \\
                              & GRU w/ RDIS                      & .0239 & .0255 & .0276 & .0311 & .0373 & .0434 & .0506 & .1070 \\
                              & Bi-GRU w/ RDIS                   & .0287 & .0226 &  .0241 & .0251 & \textbf{.0277} & .0321 & \textbf{.0350} & .0837 \\
                              \cmidrule(l){2-10}
                        &      NRTSI-all              & .0384 & .0411 &  .0384 & .0414 & .0402 & .0415 & .0461 & .0524 \\
& NRTSI              & \textbf{.0165} & \textbf{.0195} &  \textbf{.0196} & \textbf{.0229} & .0286 & \textbf{.0311} & .0362 & \textbf{.0445} \\

                              \bottomrule
\end{tabular}
\end{small}
\caption{The MSE comparison of imputation methods under different missing rates on both the air quality dataset and the gas sensor dataset.}
\label{table:pod}
\end{table*}

\subsection{Partially Observed Dimensions} In the main text, we evaluate NRTSI to handle the scenario where dimensions are partially observed in Section 4.6. In this section, we conduct an ablation study to investigate a variant of NRTSI called NRTSI-all that imputes all missing data in a single forward pass without the hierarchical imputation procedure discussed in Section 3.4 of the main text. We report the MSEs of NRTSI-all, as well as all the baselines in RDIS \cite{choi2020rdis}, on the air quality dataset and the gas sensor dataset in Table \ref{table:pod}. It is shown that NRTSI-all yields inferior performance compared to our proposed hierarchical imputation strategy on the gas sensor dataset. On the air quality dataset, NRTSI-all slightly outperforms NRTSI because this dataset contains repeated daily air quality patterns in the training data, and even some simple methods like KNN, Forward, and Backward perform reasonably well.

\section{Baseline Hyperparameter Search and Number of Parameters}
\label{sec:baseline_hyper}
We perform an extensive hyperparameter search for all the baselines. Though the best configurations of these baselines contain fewer parameters compared to NRTSI (NRTSI contains 84M parameters according to Appendix \ref{sec:NRTSI_arch}), we find that further increasing their number of parameters (even comparable to NRTSI by using more layers and neurons) does not further improve their performance. Thus, the superiority of NRTSI is due to the novel architecture rather than naively using more parameters or tuning hyperparameters only for NRTSI. In fact, NRTSI shares hyperparameters across all the datasets (see Appendix \ref{sec:NRTSI_arch}).
\subsection{NAOMI and NAOMI-$\Delta_{t}$} We use the codes in the official GitHub repository \footnote{\url{https://github.com/felixykliu/NAOMI}} of NAOMI and perform extensive hyperparameter search to report the performance of NAOMI and NAOMI-$\Delta_{t}$ on the MuJoCo physics simulation dataset, the football player trajectory dataset, and the irregularly-sampled Billiards dataset. 

\textbf{Hyperparameter Searching Range}\ \ \ For each dataset, we search the RNN hidden size in \{200, 250, 300, 350, 400\}, the number of RNN layers in \{1,2,3\}, the number of resolution levels R in \{4, 5\}, the decoder dimension in \{100, 200, 300\}, and the starting learning rate in \{$1\times10^{-4}$, $5\times10^{-4}$, $1\times10^{-3}$, $5\times10^{-3}$\}. 

\textbf{Best Configurations and Number of Parameters}\ \ \ On the MuJoCo physics simulation dataset, the best model contains 1.56M parameters with RNN hidden size 250, 2 RNN layers, R equals 4, decoder dimension 200, and starting learning rate $1\times10^{-3}$. 

On the football player trajectory dataset, the best model contains 2.13M parameters with RNN hidden size 300, 2 RNN layers, R equals 4, decoder dimension 200, and starting learning rate $1\times10^{-3}$. 

On irregularly-sampled Billiards dataset, the best model contains 1.13M parameters with RNN hidden size 200, 2 RNN layers, R equals 5, decoder dimension 200, and starting learning rate $5\times10^{-4}$.  
\subsection{Latent ODE} We use the codes in the official GitHub repository of Latent ODE \footnote{\url{https://github.com/YuliaRubanova/latent_ode}} and perform extensive hyperparameter search to report the performance of Latent ODE on the irregularly-sampled Billiards dataset, the football player trajectory dataset, the air quality dataset and the gas sensor dataset. 

\textbf{Hyperparameter Searching Range}\ \ \ For each dataset, we search the latent dimension in \{8,16,32,64,128\}, the number of generation layers in \{1,2,3,4\}, the number of neurons at each generation layer in \{16,32,64,128\}, the number of encoding layers in \{1,2,3,4\}, the number of neurons at each encoding layer in \{16,32,64,128\}, the starting learning rate in \{$1\times10^{-4}$, $5\times10^{-4}$, $1\times10^{-3}$, $5\times10^{-3}$\} and the ODE solver in \{"dopri5", "euler", "midpoint", "rk4"\}. 

\textbf{Best Configurations and Number of Parameters}\ \ \ On the irregularly-sampled Billiards dataset, the best model contains 0.84M parameters with 3 encoding layers with 64 neurons, 3 decoding layers with 32 neurons, latent dimension equals to 32, starting learning rate of $5\times10^{-4}$ and ODE solver "dopri5". 

On the football player trajectory dataset, the best model contains 1.22M parameters with 3 encoding layers with 64 neurons, 3 decoding layers with 64 neurons, latent dimension equals to 64, starting learning rate of $1\times10^{-4}$ and ODE solver "dopri5".

On the air quality dataset, the best model contains 0.68M parameters with 2 encoding layers with 64 neurons, 2 decoding layers with 32 neurons, latent dimension equals to 32, starting learning rate of $1\times10^{-4}$ and ODE solver "dopri5". 

On the gas sensor dataset, the best model contains 1.17M parameters with 3 encoding layers with 64 neurons, 3 decoding layers with 64 neurons, latent dimension equals to 32, starting learning rate of $1\times10^{-4}$ and ODE solver "dopri5".

On the irregularly-sampled sinusoidal function dataset, the best model contains 72.8K parameters with 1 encoding layers with 32 neurons, 1 decoding layers with 32 neurons, latent dimension equals to 16, starting learning rate of $1\times10^{-4}$ and ODE solver "dopri5".

\subsection{Neural CDE} We use the codes in the official GitHub repository of Neural CDE \footnote{\url{https://github.com/patrick-kidger/torchcde}} and perform extensive hyperparameter search to report the performance of Neural CDE on the irregularly-sampled Billiards dataset, MuJoCo dataset, the air quality dataset and the gas sensor dataset. 

\textbf{Hyperparameter Searching Range}\ \ \ We search the latent dimension in the number of linear layers in \{1,2,3,4,5\}, the number of neurons at each layer in \{32,64,128,256\}, the starting learning rate in \{$1\times10^{-4}$, $5\times10^{-4}$, $1\times10^{-3}$, $5\times10^{-3}$\}, the interpolation schemes in \{"natural cubic splines", "linear interpolation", "rectilinear interpolation"\} and the ODE solver in \{"dopri5", "euler", "midpoint", "rk4"\}. 

\textbf{Best Configurations and Number of Parameters}\ \ \ On the irregularly-sampled Billiards dataset, the best model contains 0.49M parameters with 4 linear layers with 128 neurons, starting learning rate of $5\times10^{-4}$, natural cubic splines interpolation and ODE solver "dopri5". 

On the MuJoCo dataset, the best model contains 0.54M parameters with 4 linear layers with 128 neurons, starting learning rate of $1\times10^{-4}$, natural cubic splines interpolation and ODE solver "dopri5". 

On the air quality, the best model contains 0.37M parameters with 3 linear layers with 128 neurons, starting learning rate of $1\times10^{-4}$, natural cubic splines interpolation and ODE solver "dopri5".

On the gas sensor quality, the best model contains 0.53M parameters with 4 linear layers with 128 neurons, starting learning rate of $5\times10^{-4}$, natural cubic splines interpolation and ODE solver "dopri5".

\subsection{Attentive Neural Process} We use the codes in the official GitHub repository \footnote{\url{https://github.com/deepmind/neural-processes}} of Attentive Neural Process (ANP) and perform extensive hyperparameter search to report the performance of ANP on the irregularly-sampled Billiards dataset. 

\textbf{Hyperparameter Searching Range}\ \ \ There are a deterministic path and a latent path in ANP where the latent path introduces stochasticity in the prediction. Since the trajectory of the irregularly-sampled dataset is fully deterministic, we remove the latent path of ANP and train ANP with a MSE loss. We search the number of layers in the MLPs of the deterministic path in \{1,2,3,4\}, the number of multi-head self-attention blocks in \{1,2,3,4,5,6\}, the number of attention heads in \{2,4,8,10,12\}, the latent dimension in \{64, 128, 256, 512, 1024\} and the starting learning rate in \{$1\times10^{-3}$, $5\times10^{-4}$, $1\times10^{-4}$, $5\times10^{-5}$\}. 

\textbf{Best Configurations and Number of Parameters}\ \ \ The best model contains 6.30M parameters with 2 layers in MLPs, 3 multi-head self-attention blocks with 8 heads, a latent dimension of 512, and a starting learning rate of $1\times10^{-4}$.

\section{Why the Absolute Time Encoding Works}
\label{sec:absolute}
According to \eqref{eq:time} in the main text, we are using the absolute time information, which seems less meaningful than the relative time information for imputation. However, as pointed out in \cite{vaswani2017attention}, the sinusoid functions allow the model to easily learn to attend by relative positions since for any fixed offset $\Delta t$, $\phi(t+\Delta t)$ can be represented as a linear function of $\phi(t)$.
\paragraph{Theorem 1} For every sine-cosine pair $[\sin(\omega t), \cos(\omega t)]$ with frequency $\omega$, there is a linear transformation $A \in \R^{2\times2}$ that is independent of $t$ and satisfies the following equation,
\begin{align}
\label{eq:1}
A\begin{bmatrix}
	    \sin(\omega t) \\
	    \cos(\omega t)
	\end{bmatrix} = \begin{bmatrix}
	    \sin(\omega (t + \Delta t)) \\
	    \cos(\omega (t + \Delta t))
	\end{bmatrix}.
\end{align}

\begin{proof}[Proof]
Suppose 
\begin{align}
\label{eq:2}
    A = \begin{bmatrix}
	    u_1 & v_1 \\
	    u_2 & v_2
	\end{bmatrix},
\end{align}
then
\begin{align}
\label{eq:3}
A\begin{bmatrix}
	    \sin(\omega t) \\
	    \cos(\omega t)
	\end{bmatrix} = \begin{bmatrix}
	    u_1 \sin(\omega t) + v_1 \cos(\omega t)\\
	    u_2 \sin(\omega t) + v_2 \cos(\omega t)
	\end{bmatrix}.
\end{align}

By applying the trigonometric formulas, we can expand the right hand side of Eq. \eqref{eq:1} as 
\begin{align}
\label{eq:4}
\begin{bmatrix}
	    \sin(\omega (t + \Delta t)) \\
	    \cos(\omega (t + \Delta t))
	\end{bmatrix} = \begin{bmatrix}
	    \sin(\omega t)\cos(\omega \Delta t) + \cos(\omega t)\sin(\omega \Delta t) \\
	    \cos(\omega t)\cos(\omega \Delta t) - \sin(\omega t)\sin(\omega \Delta t)
	\end{bmatrix}.
\end{align}

Substituting the left hand side of Eq. \eqref{eq:1} with Eq. \eqref{eq:3} and the right hand side of Eq. \eqref{eq:1} with Eq. \eqref{eq:4}, we have  
\begin{align}
\label{eq:5}
    A = \begin{bmatrix}
	    \cos(\omega \Delta t) & \sin(\omega \Delta t) \\
	    -\sin(\omega \Delta t) & \cos(\omega \Delta t)
	\end{bmatrix}.
\end{align}

From Eq. \eqref{eq:5}, we can see that there exists a linear transformation $A$ that is independent of $t$ and satisfies Eq. \eqref{eq:1}.

\end{proof}

\section{Potential Negative Societal Impacts} 
\label{sec:impact}
Same with any other imputation models, the imputation results of NRTSI depend on the training data. Therefore, the results may be biased if the training data is biased.
\end{document}